\RequirePackage{fix-cm}
\documentclass[smallextended]{svjour3}       
\smartqed  
\usepackage{graphicx}
\usepackage{amsmath}
\usepackage{amsfonts}
\usepackage{xcolor}
\usepackage{enumerate}
\usepackage{soul}

\usepackage{caption}
\usepackage{subcaption}
\usepackage{multirow}

\begin{document}
\newcommand{\bff}{\mathbf{f}}

\title{
Variational Geometry-aware Neural Network based Method for Solving High-dimensional Diffeomorphic Mapping Problems
}


\author{Zhiwen LI$^\dagger$ \and Cheuk Hin HO$^\dagger$ \and Lok Ming LUI$^*$ } 



\institute{
 \at
 {$^\dagger$Equal contributions\\
 $^*$Corresponding author\\
 Funding: Lok Ming Lui is supported by HKRGC GRF (Project ID: 14306721).
 }
 \\
Zhiwen LI\at
Department of Mathematics, The Chinese University of Hong Kong, Shatin, Hong Kong \\
\email{zwli@math.cuhk.edu.hk}           
\and
Cheuk Hin HO \at
Department of Mathematics, University of British Columbia, Vancouver, V6T1Z2, BC, Canada \\
\email{jerryho528@math.ubc.ca}
\and
Lok Ming LUI \at
Department of Mathematics, The Chinese University of Hong Kong, Shatin, N.T., Hong Kong \\
\email{lmlui@math.cuhk.edu.hk}
}

\date{Received: date / Accepted: date}

\maketitle

\begin{abstract}
Traditional methods for high-dimensional diffeomorphic mapping often struggle with the curse of dimensionality. We propose a mesh-free learning framework designed for $n$-dimensional mapping problems, seamlessly combining variational principles with quasi-conformal theory. Our approach ensures accurate, bijective mappings by regulating conformality distortion and volume distortion, enabling robust control over deformation quality. The framework is inherently compatible with gradient-based optimization and neural network architectures, making it highly flexible and scalable to higher-dimensional settings. Numerical experiments on both synthetic and real-world medical image data validate the accuracy, robustness, and effectiveness of the proposed method in complex registration scenarios.

\end{abstract}

\newpage

\section{Introduction}
Many problems in imaging science and computational geometry can be conceptualized as a mapping problem, which involves optimizing a mapping between two related domains that satisfies specific conditions. For instance, image registration can be framed as seeking the optimal mapping between two images by aligning their intensity while adhering to other specific constraints such as bijectivity \cite{chen2019image,LamKaChun2014Lair,LamKaChun2015QHMI,LeeYinTat2016LTwL,zhang2019multimodality,ZhangDaoping2022AUFf}. Another notable example is image segmentation, which targets a deformation map on a predefined template shape to extract important objects from a target image \cite{ChanHei-Long2018TISb,SiuyChunYin2020Iswp,ZhangDaoping2021Taci,ZhangDaoping2021T3IS}. Deformation study of patterns and parameterization of surfaces could also be, respectively, formulated as a deformation mapping problem \cite{ChoiGaryP.T.2020Savi,LuiLokMing2014SAoP} and a spatial map problem that seeks a map to assign points on a surface to a separate coordinate system \cite{Choi2016Fsqp,Choi2015FFla,Choi2015FDCP,gu2003global}. 

Various nonrigid mapping methods have been developed in the past few decades for solving the mapping problem. A family of intensity-based methods such as the Demons algorithm \cite{THIRION1998243,VERCAUTEREN2009S61}, the elastic method \cite{HeJianchun2003LDIC} and the large deformation diffeomorphic metric mapping \cite{BegM.Faisal2005CLDM,ChristensenG.E.1996Dtul,DUPUISPAUL1998VPOF}, has been proposed. These methods are commonly applied together with given intensity/signal information on the source and target domains to determine the optimal mapping. Landmark-based and hybrid methods, which aid the mapping searching algorithm by matching the given information, are also  presented in previous works \cite{RohrK.2001Leru,BooksteinF.L.1989Pwts,GlaunèsJoan2008LDDM,JoshiS.C.2000Lmvl,LuiLokMing2010OCSR,LUI2007847}.

Most of the above mentioned applications, practically, seek for a diffeomorphic mapping. This particular class of mapping problems is referred to as \emph{Diffeomorphism Optimization Problems} (DOP). The largest challenge of DOP arises from the difficulty of ensuring the diffeomorphicity of the solution. To impose diffeomorphicity, Christensen et al. \cite{ChristensenG.E.1996Dtul} introduced a regridding algorithm to confine the transformation of image deformation, which ensures a globally positive definite Jacobian, while Leow et al. \cite{LeowA.D.2007SPoJ} conducted a statistical analysis of Jacobian maps and developed an unbiased deformation field construction framework. More recently, Modat et al. \cite{Modat2010LungNiftyReg} proposed a variational model incorporating joint bending energy and squared Jacobian determinant penalty terms to obtain a transformation for lung registration. 

Among various proposed methods for maintaining the diffeomorphicity of the resulting mapping,  the Quasi-Conformal (QC) theory has gained popularity in DOP due to its ability to measure the diffeomorphicity of the target map. The bijectivity and local geometric distortion of the map can be easily controlled by the corresponding Beltrami coefficient $\mu$. This includes a wide range of works in nonrigid image registration \cite{LamKaChun2014Lair,Tu2020Diff,YUNG2018561,zeng2011registration}, surface mapping methods \cite{chien2016bounded,choi2020parallelizable,Choi2018Alff,Choi2015FDCP,LamKaChun2017Oqpw,LipmanYaron2012Bdms,meng2016tempo,WeberOfir2012CEQM,WongTszWai2014Coqs,WongTszWai2015Csuu,YangYi-Jun2020Qrm,zeng2011registration} with applications to geometry processing \cite{choi2017scrvtem,Choi2016Scpo,Choi2016Fsqp,lyu2024spherical}, biological shape analysis \cite{Choi2020Tmuq,ChoiGaryP.T.2020Savi} and medical visualization \cite{Choi2017Cmoc,Choi2015FFla,TaDuyan2022Qcot,zeng2014colon,lyu2023bijective}. Still, most of the aforementioned methods were designed only for 2D Euclidean space. A few frameworks have been developed to compute high-dimensional quasi-conformal mappings based on conformality distortion and landmark mismatching \cite{LeeYinTat2016LTwL,NaitsatAlexander2018Gatd,NaitsatAlexander2021OIMa,NaitsatAlexander2018Gdmf,ZhangDaoping2022AUFf}. Zhang et al. \cite{ZhangDaoping2022AUFf} proposed a unifying framework that considers a variational model that integrates landmark constraints, intensity constraints and volumetric information for computing $n$-D quasi-conformal mappings. Still, discretization over the spatial domain suffers from an exponential scaling in both memory and computational cost, which remains intractable in practice.

To overcome a similar challenge that arises in numerical PDEs, machine learning techniques have been employed, exhibiting notable scalability with respect to the dimensionality of the spatial domain. Raissi et al. \cite{PINN} proposed frameworks on both data-driven solution and data-driven discovery of partial differential equations highlighting that a range of classical problems in diverse fields such as fluid dynamics and quantum mechanics can be effectively tackled using shallow neural networks. More importantly, Yu \cite{DeepRitz} introduced the Deep Ritz method and demonstrated that simple neural networks can be used to solve variational problems that arise from traditional elliptic PDEs.

In this work, we present a framework that bridges the gap between variational formulations for high-dimensional diffeomorphism optimization problems and machine learning techniques for PDEs, primarily aiming to address the curse of dimensionality (i.e. scalability) stemming from domain discretization in conventional approaches. In short, the main contributions in this paper are threefold:
\begin{enumerate}
\item The number of parameters in our mesh-free model does not directly depend on any input information, such as image size, giving rise to better scalability, flexibility and efficiency compared to traditional methods that rely on domain discretization. 
\item By parametrizing the solution mapping smoothly using neural networks, our learning-based framework ensures the mapping is smooth with respect to the input. This smooth ansatz, combined with our proposed bijectivity loss and the conformality distortion metric in high-dimensional Quasi-conformal geometry, readily adapts to gradient-based optimization and effortlessly regulates the diffeomorphic property of the learned transformation.
\item Our proposed machine learning architecture guarantees the satisfaction of the prevalent Dirichlet boundary conditions in imaging problem. Strictly imposing these conditions during optimization allows for stable convergence and more efficient optimization.
\end{enumerate}

The rest of the paper is organized as follows. We first review the mathematical background of QC maps in 2D, its generalization to $n$-dimension and the famous deep Ritz method~\cite{DeepRitz} in Section~\ref{sec:background}. In Section~\ref{sec:proposed_method}, we propose a mesh-free learning framework that is parameterized smoothly with explicitly built-in boundary conditions for mapping problems defined on hyper-cube. Numerical experiments are presented in Section~\ref{sec:experiments} to demonstrate the accuracy and scalability of our proposed framework on a wide range of analytic examples as well as real-world problems in medical imaging.

\section{Background} 
\label{sec:background}

To introduce notation and review involved concepts, we first start with a brief review on quasi-conformal maps (cf. Section~\ref{sec:2dqc}). A volume-preserving prior, which is common in imaging tasks, is then introduced in (cf. Section~\ref{sec:vol_preserving}). Finally, we outline the formalism of the deep Ritz method, which is the major tool to be used in this work. 

\subsection{Quasi-conformal Mapppings} \label{sec:2dqc}

As a generalization of conformal maps, the \emph{quasi-conformal maps} are orientation-preserving homeomorphisms between Riemann surfaces with bounded conformality distortion \cite{gardiner2000quasiconformal}. Mathematically, $f : \mathbb{C} \rightarrow \mathbb{C}$ is a quasi-conformal if there exists a Lebesgue measurable $\mu: \mathbb{C} \rightarrow \mathbb{C}$ with $\|\mu\|_{\infty} < 1$ such that it satisfies the Beltrami Equation
\begin{equation}
\label{eqn:BeltramiEq}
    \frac{\partial f}{\partial \bar{z}} = \mu(z) \frac{\partial f}{\partial z}.
\end{equation}
The \emph{Beltrami coefficient} $\mu$ measures the conformality of $f$. Conversely, given a Beltrami coefficient $\mu:\mathbb C\rightarrow \mathbb C$ with $\|\mu\|_\infty < 1$, there always exists a quasi-conformal mapping from $\mathbb C$ onto itself, satisfying the \eqref{eqn:BeltramiEq} in distribution sense \cite{gardiner2000quasiconformal}. 

\begin{figure}
\centering
\begin{subfigure}{.5\textwidth}
  \centering
  \includegraphics[width=\linewidth]{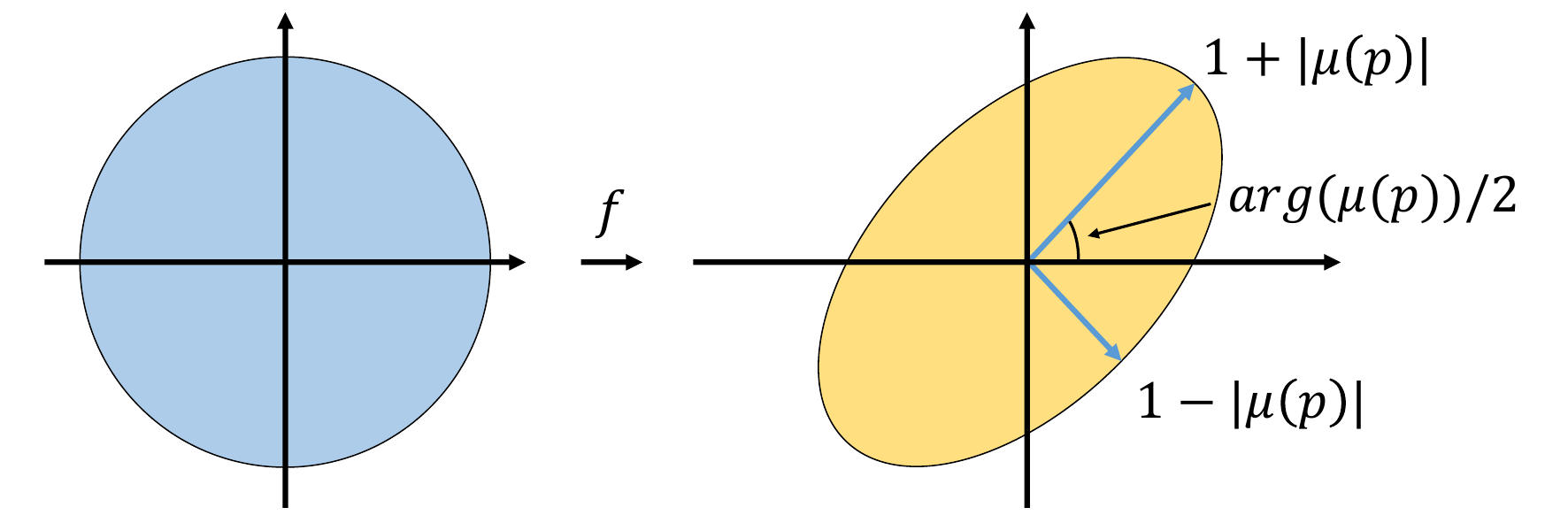}
  \caption{}
  \label{fig:2d_qcmap}
\end{subfigure}%
\begin{subfigure}{.5\textwidth}
  \centering
  \includegraphics[width=\linewidth]{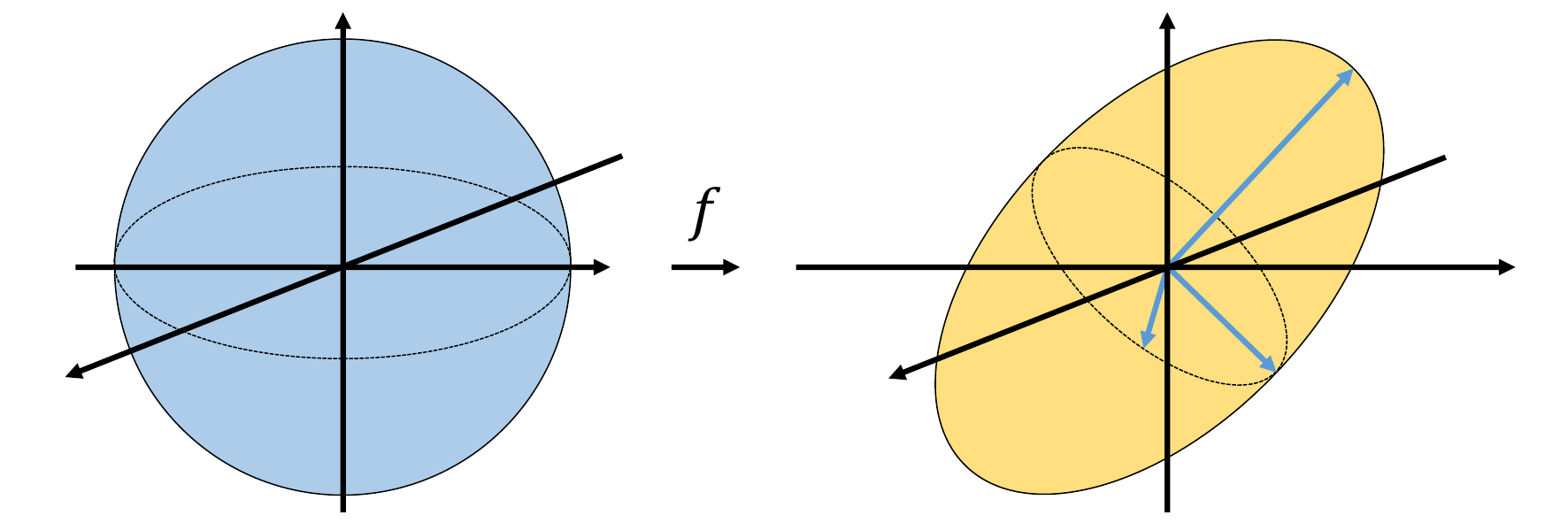}
  \caption{}
  \label{fig:3d_qcmap}
\end{subfigure}
\caption{An illustration for the quasi-conformal maps. (a) Under a $2$D quasi-conformal map, infinitesimal circles are mapped to infinitesimal ellipses. (b) Under an $n$-D quasi-conformal map, infinitesimal spheres of dimension $n-1$ are mapped to infinitesimal ellipsoids of dimension $n-1$.}
\label{fig:qc_map}
\end{figure}

From that end, one can define the notion of conformality distortion of $f$ at a point $p \in \mathbb{C}$ in relation to the Beltrami coefficient $\mu = \mu(f)$. Indeed, the magnitudes and angles of the maximal magnification and shrinking can be determined from $\mu$. Specifically, the angle of maximal magnification is $\arg(\mu(p))/2$ with magnifying factor $1 + |\mu(p)|$, whereas the angle of maximal shrinking is the orthogonal angle $(\arg(\mu(p)) - \pi)/2$ with shrinking factor $1 - |\mu(p)|$. This implies that $\mu(p)$ contains all the information about local conformality distortion of $f$ near $p$ (cf. Fig. \ref{fig:2d_qcmap}). 

The above discussion motivates the definition of dilation (or distortion) of $f$ at a point $p$
\begin{equation}
    K_d(f)(p)=\frac{1+|\mu(p)|}{1-|\mu(p)|},
\end{equation}
which is indeed the ratio between the largest and the smallest singular value of the Jacobian matrix of $f$. $K_d(f)$ can then be regarded as a quantitative measurement of non-conformality of $f$. More recently, \cite{LeeYinTat2016LTwL} extended the concept of quasi-conformal theories to $n-$D spaces by introducing a metric for the conformality distortion for diffeomorphisms of $n-$D Euclidean space 
\begin{equation} 
\label{eqn:nd-dilation}
K(f)(p) = \left\{
\begin{aligned}
\frac{1}{n} \frac{\|\nabla f(p)\|_F^2}{(\det \nabla f(p))^{2/n}}, & \quad  \text{if } \det\nabla f(p) > 0, \\
+\infty \qquad \quad, & \quad \text{if otherwise.}
\end{aligned}\right.
\end{equation}
As the value of $K(f(p))$ increases, the conformality distortion of the mapping $f$ at point $p$ becomes larger. This characteristic enables the extension of quasi-conformal theory to $n$ dimension.


\subsection{Volumetric Prior}
\label{sec:vol_preserving}
In many mapping problems, it is often desirable for the mappings to satisfy certain prescribed geometric constraints and possess low distortion in conformality or volume. Indeed, the geometric meaning of the Jacobian determinant represents the ratio of volume change under a transformation. More precisely, for a mapping  $T$ that transforms a region $R$ to $T(R)$, the Jacobian determinant $\det(\nabla T(x))$ at a point $x$ quantifies the local scaling of infinitesimal volume elements. A positive determinant indicates preservation of orientation, while a negative value signals inversion. This property is crucial in applications such as imaging and shape registration, where mappings must preserve geometric constraints like volume or conformality. Given any mapping problem with target solution $f$, by imposing a variation loss, or equivalently a regularizer, on the Jacobian determinant, one can impose a prior on volume change of the target transformation mapping. More precisely, the volumetric prior can be defined by
\begin{equation}
    \int_{\Omega_T} | \det \nabla f - \bar{V} | 
\end{equation}
where $\bar{V}: \Omega_T \rightarrow \mathbb{R}^{+}$. In this work, we focus on the case $\bar{V}=1$ which imposes a volumetric preserving prior.

\subsection{Deep Ritz Method}
The Deep Ritz method stands as one of the very first proposed PINNs. Despite its simplicity, this approach has proven highly effective in addressing PDEs that can be cast as variational losses, including eigenvalue problems. Specifically, the Deep Ritz method aims to minimize variational problems formulated as
\begin{equation}
\label{eqn:general_variational_loss}
    \min_{\Phi \in \mathcal{I}} \mathcal{L}(\Phi)
\end{equation}
where $\mathcal{I}$ denotes the space of admissible functions. Comprising three fundamental components—a parameterized neural network serving as an ansatz, a quadrature rule for the functional, and an optimization algorithm for solving the final problem—the Deep Ritz method has demonstrated robustness and efficacy across a broad spectrum of variational problems. Notably, previous works have highlighted that even relatively simple ansatz can suffice for solving diverse variational problems, offering robustness against exponential scaling inherent to the number of parameters with respect to the input domain's dimensionality.

\section{Proposed Method} \label{sec:proposed_method}

\subsection{Problem Setting}
\label{sec:nDmapingproblem}
In this work, we focus on the $n$-D \emph{diffeomorphism optimization problem} \cite{ashburner2007fast,VERCAUTEREN2009S61,JoshiS.C.2000Lmvl}, a sub-class of the general mapping problem \eqref{eqn:mapping_prob}, with application to registration tasks. To gain a mathematical insight, we first define the notion of mapping problems. Given two domains $\Omega_S,\Omega_T \subset \mathbb{R}^n$, a mapping problem can be formulated as
\begin{equation}
\label{eqn:mapping_prob}
f = \mathop{\arg\min}_{g \in \mathcal{C}}\mathcal{L}(g),
\end{equation}
where $\mathcal{L}$ is an energy functional and $\mathcal{C} \subset \{g \,|\, g: \Omega_T \rightarrow \Omega_S \}$ is the constrained solution space with desired properties. For example, in the context of image registration, the energy functional $\mathcal{L}$ typically comprises a fidelity term that measures the intensity mismatching and a smoothness regularization.

In the context of DOP, we consider a constrained space $\mathcal{C} := \mathcal{C}_{\mathbf{d}} \cap \mathcal{C}_{\rm diff}$, where $\mathcal{C}_{\rm diff}$ is the space of all diffeomorphic functions from $\Omega_T$ to $\Omega_S$, and $\mathcal{C}_{\mathbf{d}}$ is defined by
\begin{equation}
    \mathcal{C}_{\mathbf{d}} := \{g \,|\, S \circ g = T \,;\, g(q_i) = p_i, i = 1, \dots, N_{\rm lm} \},
\end{equation}
here $\mathbf{d}$ is the input data information
\begin{equation}
\label{eqn:data_d}
\mathbf{d} := (S,T) \cup \{(p_i,q_i) \,|\, p_i\in\Omega_S, q_i\in\Omega_T\}_{i=1}^{N_{\rm lm}}, 
\end{equation}
where $S:\Omega_S\rightarrow\mathbb{R}^c$ and $T:\Omega_T\rightarrow\mathbb{R}^c$ are the source and target images of $c$ channels, which are called the \emph{intensity information}. The set of landmark correspondences $\{(p_i,q_i) \,|\, p_i\in\Omega_S, q_i\in\Omega_T\}_{i=1}^{N_{\rm lm}}$ is said to be the \emph{landmark information}. 

From that end, the optimization problem \eqref{eqn:mapping_prob} to look for a diffeomorphic mapping can be formulated as
\begin{equation}
\label{eqn:d_mapping_prod}
    \min_{f\in\mathcal{C}}\, \widetilde{\mathcal{R}}(f), \\
\end{equation}
where $\widetilde{\mathcal{R}}(\cdot)$ is the regularization term to enforce the desired properties, e.g. minimizing conformality distortion \eqref{eqn:nd-dilation}, on the transformation $f$ with $\mathcal{C} = \mathcal{C}_{\mathbf{d}} \cap \mathcal{C}_{\rm diff}$. However, solution to \eqref{eqn:d_mapping_prod} may not exist. Practically, one employs a relaxation on the space $\mathcal{C}$ and reformulates \eqref{eqn:d_mapping_prod} as
\begin{equation}
\label{eqn:general_reg_loss_d}
    \min_{f : \Omega_T \rightarrow \Omega_S}\, \mathcal{R}(f) + \mathcal{L}_{\mathbf{d}}(f),
\end{equation}
where $\mathcal{L}_{\mathbf{d}}$ corresponds to imposing a soft constraint (i.e. can be inexact) with respect to the observation data $\mathbf{d}$, and $\mathcal{R}(f)$ is a regularization that takes care of, in addition to properties considered by $\mathcal{R}$, the diffeomorphicity (cf. Section~\ref{sec:three_formulations}).

\subsection{Variational Formulations of $3$D Diffeomorphic Registration Problem} \label{sec:Variational_models}

Inspired from \cite{ZhangDaoping2022AUFf,LeeYinTat2016LTwL}, we propose a learning framework that combines the traditional variational models with the machine learning framework for $n$-D diffeomorphic registration. Without loss of generality, we specifically consider the case $n=3$ while the framework is readily generalizable to higher dimension. 

\subsubsection{Diffeomorphic Loss}
\label{sec:diff_loss}

To make \eqref{eqn:general_reg_loss_d} specific, it suffices to define $\mathcal{R}(f)$ and $\mathcal{L}_{\mathbf{d}}$. As above mentioned, the regularization loss $\mathcal{R}$ takes into consideration the diffeomorphicity, and any additional property of the target mapping $f$. To ensure bijectivity and smoothness, we introduce the following regularization functional $\mathcal{R}$ for \eqref{eqn:general_reg_loss_d} defined by 
\begin{align}
\label{eqn:lee_model}
     \mathcal{R}(f; \alpha_1) & := \frac{\alpha_1}{2}\int_{\Omega_T}\|\Delta f\|^2 \,dq\, ,\\
     \label{eqn:bijectivity_const} \textrm{s.t. } & \det \nabla f > 0.
\end{align}
where $\alpha_1$ is a hyperparameter that balances the strength between $\mathcal{R}$ and $\mathcal{L}_{\mathbf{d}}$. Here, we incorporate a \emph{bijectivity} constraint \eqref{eqn:bijectivity_const} to ensure that $f$ is bijective. Intuitively, minimizing \eqref{eqn:lee_model} yields a diffeomorphic transformation, which, combined with the data loss, produces a diffeomorphic mapping that satisfies the given information.

Still, it remains unclear how to implement the inequality constraint \eqref{eqn:bijectivity_const}. Traditionally, an auxiliary variable is introduced to form a Lagrangian, which is then solved using the alternating direction method of multipliers (ADMM). However, this approach increases computational overhead and often yields unsatisfactory results due to the difficulty of computing the gradient of the loss associated with the bijectivity constraint with respect to the discretized solution. We eliminate this by directly incorporating a bijectivity loss variationally into \eqref{eqn:lee_model} through redefinition:
\begin{align}
\label{eqn:bij_varia_regularizer}
    \begin{split}
     \mathcal{R}(f; \alpha_1, \alpha_2) := \frac{\alpha_1}{2}\int_{\Omega_T}\|\Delta f\|^2 \,dq\, + \frac{\alpha_2}{2} \int_{\Omega_T \backslash \Omega^{+}_T} | \det \nabla f |^2 \,dq\,
    \end{split}
\end{align}
where $\Omega_T^+ := \{q\in\Omega_T: \det \nabla f(q) > 0\}$. Notably \eqref{eqn:bij_varia_regularizer} is fully \textit{differentiable} with respect to arbitrary parameterization for $f$. 



\subsubsection{Conformality Loss}
As discussed in Section~\ref{sec:2dqc}, in many context we are interested in minimizing the conformal distortion of the target mapping. To that end, we can define 
\begin{align}
\label{eqn:bij_varia_regularizer_conformal}
    \begin{split}
     \mathcal{R}(f; \alpha_1, \alpha_2, \alpha_3) := \frac{\alpha_1}{2}\int_{\Omega_T}\|\Delta f\|^2 \,dq\, + \frac{\alpha_2}{2}  \int_{\Omega_T \backslash \Omega^{+}_T} | \det \nabla f |^2 \,dq\, + \alpha_3 \int_{\Omega_T } & K(f)\,dq\ .
    \end{split}
\end{align}
Which, as motivated in \eqref{eqn:nd-dilation}, generalizes the concept of quasi-conformal mapping to arbitrary dimensions and allows our framework to be applied in arbitrary dimensions.

\subsubsection{Volumetric-prior Loss}
\label{sec:volume_loss}
In registration-based segmentation problems, one would expect the volume-preserving property, or impose a prior on the mapping to be either diminishing or expanding. 
\begin{align}
\label{eqn:bij_varia_regularizer_conformal_volume}
    \begin{split}
     \mathcal{R}(f; \alpha_1, \alpha_2, \alpha_3, \alpha_4) := &  \frac{\alpha_1}{2}\int_{\Omega_T}\|\Delta f\|^2 \,dq\, + \frac{\alpha_2}{2}  \int_{\Omega_T \backslash \Omega^{+}_T} | \det \nabla f |^2 \,dq\, + \\
     & \alpha_3 \int_{\Omega_T } K(f)\,dq\ +  \frac{\alpha_4}{2} \int_{\Omega_T} | \det \nabla f - \bar{V} |^2 \,dq\ 
    \end{split}
\end{align}
In this work, we are mostly interested in the case where $\bar{V}=1$, equivalent to a volume-preserving prior.

\subsubsection{Data Loss}
\label{sec:three_formulations}
With a regularizer targeting diffeomorphicity, we are at the place of encoding the given information $\mathbf{d}$ into our variational model. We first consider the simplest form of $\mathbf{d}$ that contains only the landmark information. We propose a \textit{soft} (i.e. non-exact) landmark constraint that adapts canonically to our proposed framework and similarly applied in non-rigid landmark registrations \cite{crum2004non}. More precisely, let $\boldsymbol{\alpha} = (\alpha_1, \alpha_2, \alpha_3, \alpha_4)$, the \emph{landmark matching formulation} reads
\begin{equation}\label{eqn:landmark_formulation}
\begin{aligned}
\min_{f: \Omega_{T} \rightarrow \Omega_{S}} \mathcal{R}(f; \boldsymbol{\alpha}) & + \frac{\alpha_5}{2 N_{\rm lm}}\sum^{N_{\rm lm}}_{i = 1} |f(q_i) - p_i|^2. \\
\end{aligned}
\end{equation}
In the context where only the intensity information is given, one can substitute the landmark-matching term in $\eqref{eqn:landmark_formulation}$ by the intensity-matching term to obtain the \emph{intensity matching formulation}:
\begin{equation}\label{eqn:intensity_formulation}
\begin{aligned}
\min_{f: \Omega_{T} \rightarrow \Omega_{S}} \mathcal{R}(f; \boldsymbol{\alpha}) & + \frac{\alpha_6}{2} \int_{\Omega_T}(S\circ f - T)^2 \,dq.\\
\end{aligned}
\end{equation}

Similarly, when incorporating both the landmark-matching term and the intensity-matching term, the \emph{hybrid matching formulation} reads
\begin{equation}
\begin{aligned}
\min_{f: \Omega_{T} \rightarrow \Omega_{S}} \mathcal{R}(f; \boldsymbol{\alpha}) + \frac{\alpha_5}{2 N_{\rm lm}} \sum^{N_{\rm lm}}_{i = 1} |f(q_i) & - p_i|^2 + \frac{\alpha_6}{2} \int_{\Omega_T}(S\circ f - T)^2 \,dq. \\
\end{aligned}
\label{eqn:hybrid_formulation}
\end{equation}
In practice, the above variational formation \eqref{eqn:hybrid_formulation} can be estimated by discretizing the following equivalent formulation in expectation notation
\begin{equation}
\label{eqn:hybird_formulation_expectation}
\min_{f: \Omega_{T} \rightarrow \Omega_{S}} \mathbb{E}_{x \sim \mathbb{S}_1}[\mathcal{R}(f(x); \boldsymbol{\alpha})] + \frac{ \alpha_5}{2 N_{\rm lm}} \sum^{N_{\rm lm}}_{i = 1} |f(q_i) - p_i|^2 + \frac{\alpha_6}{2} \mathbb{E}_{x \sim \mathbb{S}_2}[(S\circ f - T)^2].
\end{equation}
where we considered different sampling distributions $\mathbb{S}_1$ and $\mathbb{S}_2$ for the regularization term and image matching terms. It is important to note that $\mathbb{S}_2$ should be designated based on different image resolutions to balance the efficiency and accuracy when minimizing the variational loss for the resulting mapping $f$.

To conclude this section, we demonstrate that the variational model \eqref{eqn:hybrid_formulation} admits a minimizer. While we assume $\Omega_S=\Omega_T = \Omega$, the proof can be readily extended to any $\Omega_T$ and $\Omega_S$, provided that both of them are simply connected and bounded.
\begin{theorem}\label{thm:existence}
    Suppose $\Omega$ is bounded and simply connected, $S$, $T$ are continuous functions from $\Omega \subset \mathbb{R}^n \rightarrow \mathbb{R}$, and $\alpha_i > 0$ for $i = 1, \cdots, 6$. Let
    \begin{equation*}
        \mathcal{A} := \{f \in \mathcal{C}^2(\Omega) : \| f \|_\infty \leq c_1, \| \nabla f \|_{\infty} \leq c_2, \| \nabla^2 f\|_{\infty} \leq c_3, \, f \text{ satisfy \eqref{eqn:boundary_cond}} \}
    \end{equation*}
    for some $c_j > 0$ for $j = 1, \cdots, 3$. Then the proposed model \eqref{eqn:hybrid_formulation} admits a minimizer in $\mathcal{A}$.
\end{theorem}
The proof is given in Appendix~\ref{append:thm_existence}.

\subsubsection{Boundary Conditions: Hard Constraint v.s. Soft Constraint}
The mapping registration problem is typically addressed within an $n$-D hypercube \cite{ZhangDaoping2022AUFf}. For simple illustration, we focus on $\Omega_S=\Omega_T=[0,1]^3$. Denote the mapping $f: \Omega_T \rightarrow \Omega_S$ by $f(x, y, z)=(u, v, w)$, then the general Dirichlet boundary conditions for mapping problems are given by
\begin{equation}\label{eqn:boundary_cond}
\begin{split}
    f(a, y, z) = (a, v, w); \quad f(x, b, z) = (u, b, w); \quad f(x, y, c) = (u, v, c);
\end{split}
\end{equation}
where $a, b, c \in \{0,1\}$ are fixed and $x,y,z \in [0,1]$. These conditions are commonly imposed so that the points located on an edge are allowed to move exclusively along that edge, while points situated on a boundary plane are constrained to move exclusively within that particular plane.

\emph{Soft constraint.} Traditional approaches for enforcing boundary conditions on the solution map typically impose a soft boundary constraint by incorporating a boundary loss term into the augmented Lagrangian function. Note that the boundary of a three-dimensional unit cube consists of 6 planes $\{P_i\}_{i=1}^6$ and 12 edges $\{E_j\}_{j=1}^{12}$. Let $n_i$ be the normal vector of the plane $P_i$ and let $\{ n_j^1, n_j^2 \}$ be the two normal vectors of the two planes that intersect at edge $E_j$. Then the boundary conditions \eqref{eqn:boundary_cond} can be enforced by
\begin{equation}\label{eqn:loss_soft_bdy}
\begin{split}
    \mathcal{L}_{boundary} = & \sum_{i=1}^6 \int_{P_i} |\left(f(q)-q\right) \cdot n_i|^2 dq    \\
    + & \sum_{j=1}^{12} \int_{E_j} |\left(f(q)-q\right) \cdot n_j^1|^2 + |\left(f(q)-q\right) \cdot n_j^2|^2 dq,
\end{split}
\end{equation}
where the integrals are, similar to \eqref{eqn:hybird_formulation_expectation}, implemented by the expectation of the integrand by sampling on the boundary, 
\begin{equation}\label{eqn:loss_soft_bdy_expectation}
\begin{split}
    \mathcal{L}_{boundary} = & \sum_{i=1}^6 \mathbb{E}_{q \sim \mathbb{P}_i} \left[|\left(f(q)-q\right) \cdot n_i|^2\right]    \\
    + & \sum_{j=1}^{12} \mathbb{E}_{q \sim \mathbb{E}_j} \left[ |\left(f(q)-q\right) \cdot n_j^1|^2 + |\left(f(q)-q\right) \cdot n_j^2|^2 \right]
\end{split}
\end{equation}
where $\cdot$ is the dot product, $\mathbb{P}_i$ is the sample distribution on $P_i$, and $\mathbb{E}_j$ is the sample distribution on $E_j$. The boundary loss term $\mathcal{L}_{boundary}$ \eqref{eqn:loss_soft_bdy_expectation}, multiplied by a weight $\alpha_7$, is added to the total loss \eqref{eqn:hybird_formulation_expectation} for updating the network parameters $\theta$. This gives the optimization problem 
\begin{equation}
\label{eqn:hybird_formulation_softbdy_expectation}
\begin{split}
\min_{f: \Omega_{T} \rightarrow \Omega_{S}} & \mathbb{E}_{x \sim \mathbb{S}_1}[\mathcal{R}(f(x); \boldsymbol{\alpha})] + \frac{2 \alpha_5}{N_{\rm lm}} \sum^{N_{\rm lm}}_{i = 1} |f(q_i) - p_i|^2 + \frac{\alpha_6}{2} \mathbb{E}_{x \sim \mathbb{S}_2}[(S\circ f - T)^2] \\
& +  \frac{\alpha_7}{2} \left\{ \sum_{i=1}^6 \mathbb{E}_{q \sim \mathbb{P}_i} \left[|\left(f(q)-q\right) \cdot n_i|^2\right] + \sum_{j=1}^{12} \mathbb{E}_{q \sim \mathbb{E}_j} \left[ \sum_{k=1}^2 |\left(f(q)-q\right) \cdot n_j^k|^2\right] \right\}
\end{split}
\end{equation}

\emph{Hard constraint.} Though the soft constraint approach for boundary conditions is widely used, we observe that such relaxation results in convergence issues, as we would demonstrate in Section~\ref{sec:ablation_study_boundary}. Motivated by these issues, we advocate for a hard boundary constraint approach. By explicitly designing the model structure to adhere to the boundary conditions (cf. Section~\ref{sec:net_architecture}), we effectively improve the robustness of the optimization process. Additionally, we do not have to sample on the boundary, which again speeds up the computation.

\subsection{Network Architecture}
\label{sec:net_architecture}

Our network architecture, which would be discussed later in this section, is flexible and general that any parameterized model $\widetilde{f}_\theta : \Omega_T \rightarrow \mathbb{R}^3$ can be used as a drop-in replacement in the neural network ansatz for the variational formulations \eqref{eqn:landmark_formulation}, \eqref{eqn:intensity_formulation}, and \eqref{eqn:hybrid_formulation}. To guarantee the smoothness of $f_\theta$ with respect to the input, all the activation functions and trainable layers are smooth, as demonstrated in \cite{DeepRitz}. Specifically, we implemented the deep Ritz network architecture, modifying the activation function to utilize the arctan function.

A key novelty of the architecture is the incorporation of the hard constraint for boundary conditions. Specifically, we ensure the adherence to the widely-used boundary conditions \eqref{eqn:boundary_cond}. Suppose $\mathbf{x} \in \Omega_T$. We define the full network ansatz $f_\theta$ by manipulating the output from $\widetilde{f}_{\theta}$
\begin{equation}
\label{eqn:hardboundary}
f_{\theta}(\mathbf{x}) := \widetilde{f}_\theta(\mathbf{x}) \odot \mathbf{x} \odot (1 - \mathbf{x}) + \mathbf{x},
\end{equation}
where $\odot$ denotes the Hadamard (point-wise) product. It is easy to see that $f_\theta$ satisfies the boundary conditions \eqref{eqn:boundary_cond} regardless of $\theta$. See Section~\ref{sec:ablation_study_boundary} for numerical experiments demonstrating the robustness of such a modification.

\begin{remark}
We remark here that \eqref{eqn:hardboundary} is designed for the domain $\Omega=[0,1]^3$ and the boundary conditions \eqref{eqn:boundary_cond} which are commonly used in imaging science. For a general domain $\Omega$ and general boundary conditions on $\partial \Omega$, ideally our framework can be generalized as
\begin{equation} \label{eqn:general_hardboundary}
f_{\theta}(\mathbf{x}) := \widetilde{f}_\theta(\mathbf{x}) \odot D(\mathbf{x}) + G(\mathbf{x}),
\end{equation}
where $D$ is nonzero in the interior of $\Omega$, $D(\mathbf{x}) \rightarrow 0$ smoothly as $\mathbf{x} \rightarrow \partial \Omega$, and $G(\mathbf{x})$ is a smooth extension of any given boundary condition. However, $G$ and $D$ are commonly intractable. Berg et al. \cite{berg2018unified} suggested pretraining of $G$ and $D$ with tiny neural networks. Still, the pretraining step still suffers from training error and difficult to be exact at corners. Also, it requires a two step training which imposes further computational cost, which our specifically designed architecture does not have both the above issues.
\end{remark}

\section{Numerical Experiments}
\label{sec:experiments}

\subsection{Implementation Details}
In this section, we briefly discuss the implementation details of the gradient evaluation (cf. Section~\ref{sec:gradient_evaluation}) and the sampling scheme (cf. Section~\ref{sec:sampling_scheme}) used to evaluate the variational formulation \eqref{eqn:hybird_formulation_expectation}. Common hyperparameters used in all experiments are described in Section~\ref{sec:common_hyperparameters}.

\subsubsection{Gradient and Laplacian Computation}
\label{sec:gradient_evaluation}
The proposed variational model requires both the gradient $\nabla_{\mathbf{x}}f_{\theta}$ and the Laplacian $\Delta_{\mathbf{x}}f_{\theta}$ for the optimization problem \eqref{eqn:hybrid_formulation}. Two common approaches, namely automatic differentiation and traditional methods such as finite differences, for evaluating $\nabla_{\mathbf{x}}f_\theta$ are widely adopted in the training of physics-informed neural networks \cite{DeepRitz,lim2022physics}. The approach used in this work is automatic differentiation, implemented via backward-mode differentiation in {\tt PyTorch} \cite{NEURIPS2019_9015}. Some recent studies suggest that combining numerical differentiation (e.g., finite differences) with automatic differentiation can improve both computational efficiency and accuracy; however, this direction lies beyond the scope of the current work. We refer interested readers to \cite{chiu2022can,xiang2022hybrid} for further discussion. For evaluating the Laplacian $\Delta_{\mathbf{x}}f_{\theta}$, we compute the trace of the Hessian using {\tt autograd} in {\tt PyTorch}. While more efficient and memory-optimized approaches for Laplacian evaluation exist, they are also beyond the scope of this study.

\subsubsection{Sampling Strategy for Variational Objectives}
\label{sec:sampling_scheme}
In this section, we detail the sampling scheme used to evaluate the variational formulation \eqref{eqn:hybird_formulation_expectation}. Following \cite{DeepRitz}, we begin by sampling a fixed set of interior points $\{x_m\}^{N_{\rm int}}_{m=1}$ for estimating the expectation over $\mathbb{S}_1$ in \eqref{eqn:hybird_formulation_expectation}. For the expectation over $\mathbb{S}_2$, a grid that align with the resolution  of the input image is defined and random subsampling is performed for each mini-batch. Regarding the distribution of the soft constraint in \eqref{eqn:loss_soft_bdy_expectation}, we randomly sample 400 points on each of the six boundary planes and 20 points on each of the twelve boundary edges, followed by integration using a Monte Carlo method for all examples where the constraint is imposed.

Notably, we observe that resampling the interior points $\{x_m\}^{N_{\rm int}}_{m=1}$ at every epoch hinders convergence and incurs additional computational overhead. Moreover, the proposed hard boundary constraint \eqref{eqn:hardboundary} offers a practical advantage in that no explicit sampling of boundary points is required.

\subsubsection{Experimental Setup and Hyperparameters} 
\label{sec:common_hyperparameters}
In this section, we outline several key hyperparameter choices used in our experiments. Throughout, we assume $\Omega_S = \Omega_T = [0, 1]^3$, though all experiments are readily generalizable to any hypercube domain. Unless otherwise stated, the coefficients $\alpha_1$, $\alpha_2$, $\alpha_3$, $\alpha_4$, $\alpha_5$ and $\alpha_6$ in \eqref{eqn:hybrid_formulation} are set to $0.01$, $50$, $1$, $0$, $500$ and $500$, respectively. In the intensity- or landmark-based formulations, we set $\alpha_4 = 0$ or $\alpha_5 = 0$ accordingly. All experiments are conducted on an NVIDIA RTX A6000 GPU.

The models are implemented in {\tt PyTorch} \cite{NEURIPS2019_9015}. For both synthetic and real-world datasets, training is performed for up to 8000 epochs using the ADAM optimizer \cite{DBLP:journals/corr/KingmaB14} with a learning rate of 0.001. In our experiments, we find that sampling 10,000 random points within $[0,1]^3$ is sufficient to guarantee the bijectivity of $f_\theta$. Notably, training loss typically converges within the first few thousand epochs.

To assess the effectiveness of the proposed method, we conduct experiments on both synthetic data (see Section~\ref{sec:3D_syn_data}) and real medical data (see Section~\ref{sec:4DCT_exp}).

\subsection{$3$D Diffeomorphic Registration on Synthetic Data}
\label{sec:3D_syn_data}
In this section, we evaluate the effectiveness and efficiency of our proposed framework in Section~\ref{sec:three_formulations} through experiments on several synthetic $3$D examples. We first present the results of the landmark matching formulation, where the deformation indicated by landmark information is large and asymmetric to demonstrate the efficacy of our method. Subsequently, the three formulations introduced in Section \ref{sec:three_formulations} are applied on matching intensity and landmark information generated by a synthetic $3$D mappings of large distortion. The codes and further references for better interactive visualization can be found in {\tt https://github.com/CheukHinHoJerry/QCPINN} \cite{QCPINN_sw}.

\subsubsection{Twisted Landmark Pairs}
\label{sec:lm_8}
\begin{figure}[!htbp] 
    \centering
    \subfloat[\label{fig:lm_8_dem}]{\includegraphics[width=.32\textwidth]
    {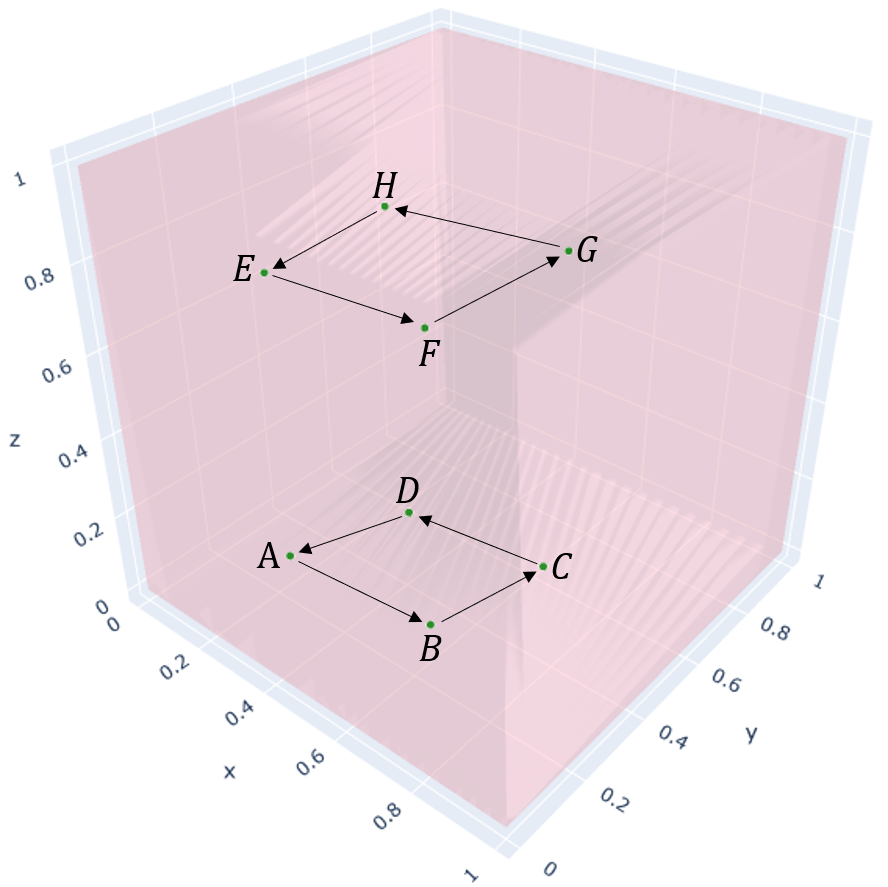}}\hspace{.1em}
    \subfloat[]{\includegraphics[width=.35\textwidth]{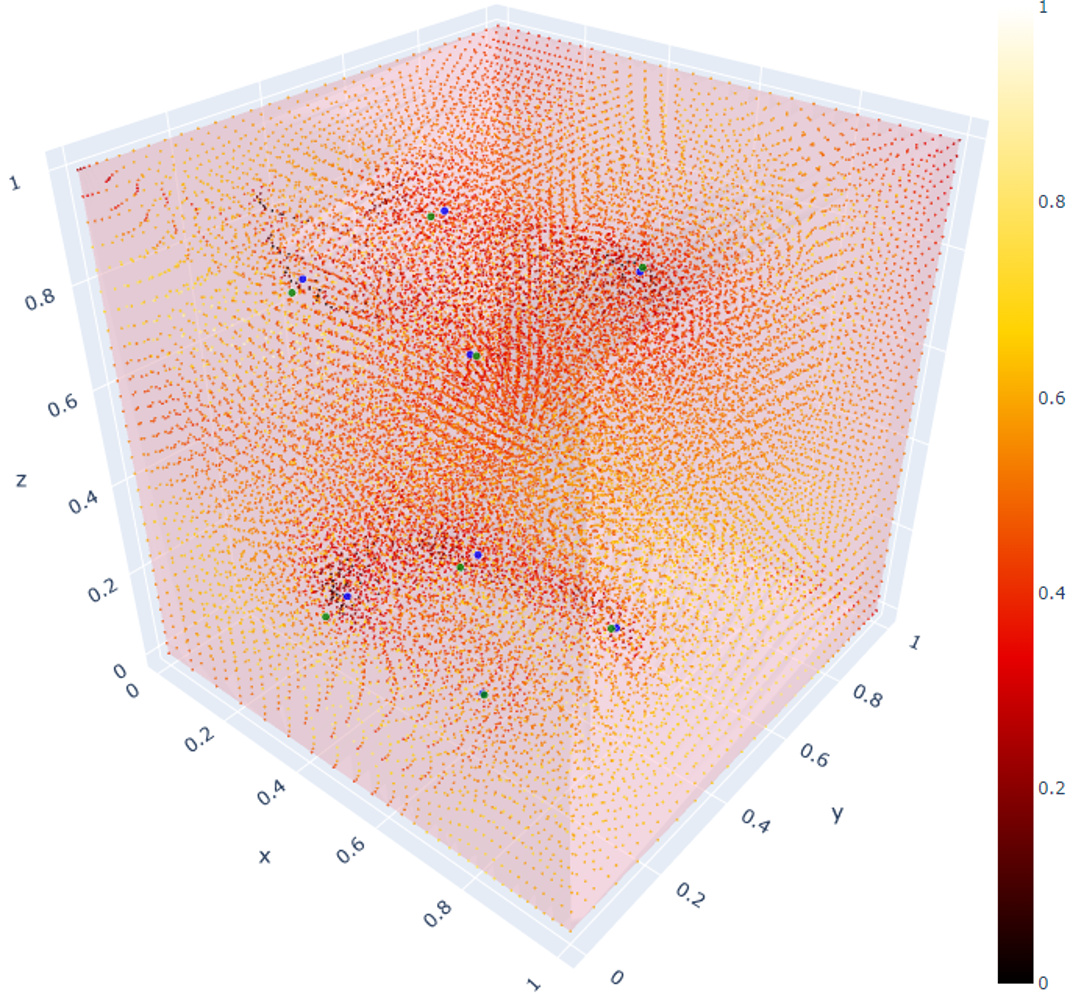}}\hspace{.1em}
    \subfloat[]{\includegraphics[width=.28\textwidth]{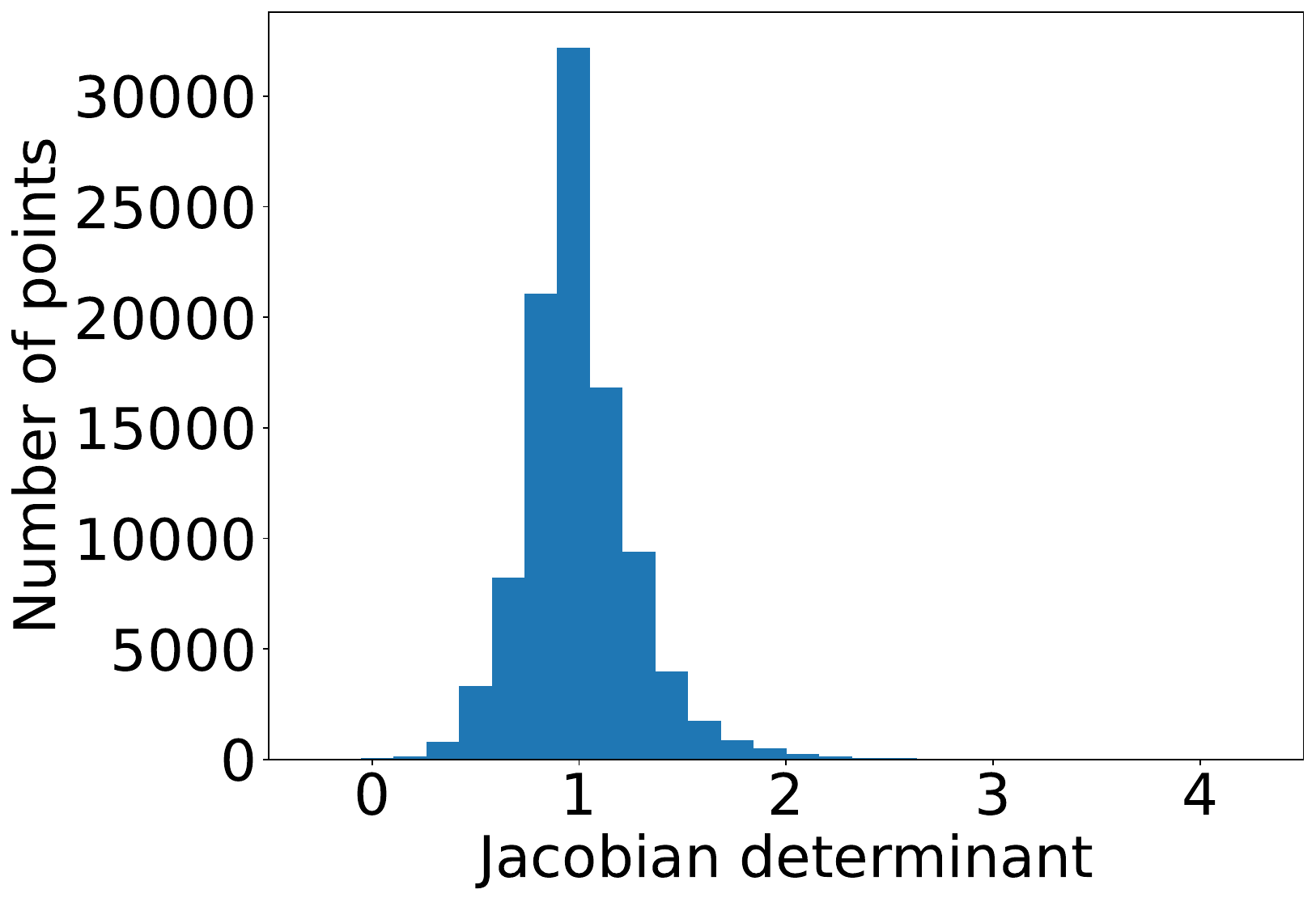}}\hspace{.1em}

    \subfloat[]{\includegraphics[width=.32\textwidth]{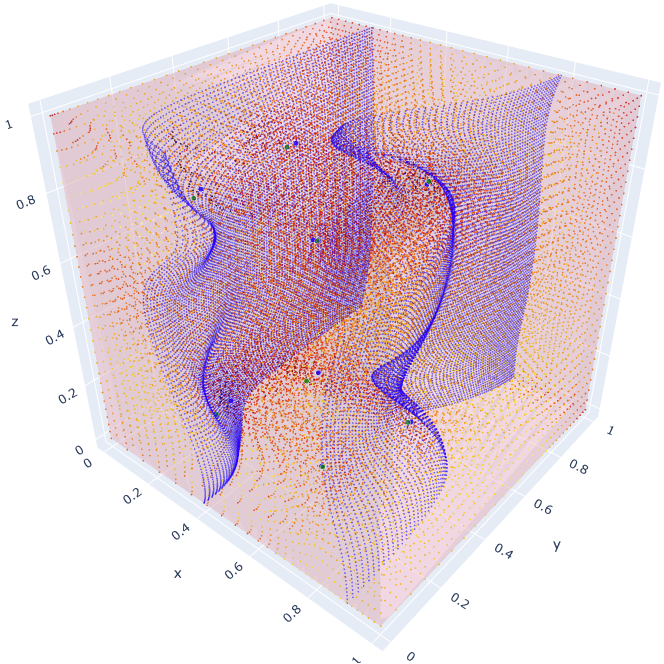}}\hspace{.1em}
    \subfloat[]{\includegraphics[width=.32\textwidth]{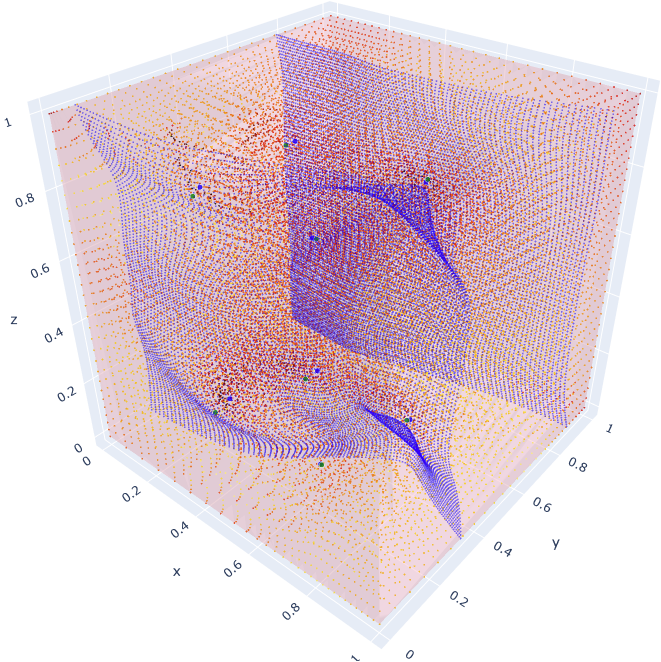}}\hspace{.1em}
    \subfloat[]{\includegraphics[width=.32\textwidth]{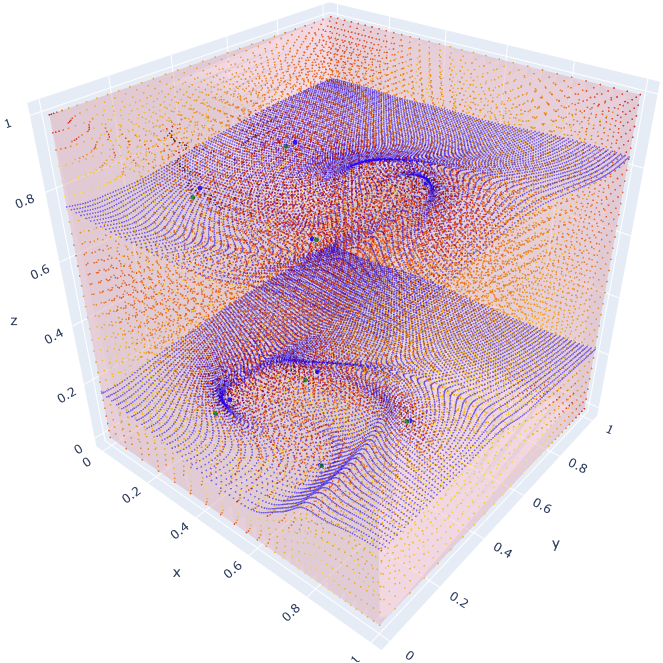}}\hspace{.1em}
    
    \caption{Results of landmark registration \eqref{eqn:landmark_formulation} on the Twisted Landmark Pairs. (a) The eight landmarks with their target positions indicated by the arrows. (b) The $3$D transformation $f_\theta$ obtained by the trained network. The points are colored by $\tanh(\frac{\ln 3}{2} \cdot \det \nabla f_\theta)$ so that the color$(p)=0.5$ when $\det \nabla f_\theta(p) = 1$. (c) Histogram of $\det\nabla f_\theta(p)$ with $100000$ uniformly random sampled points. (d) Two cross-sectional views $f_\theta(x=0.2)$ and $f_\theta(x=0.8)$. (e) Two cross-sectional views $f_\theta(y=0.2)$ and $f_\theta(y=0.8)$. (f) Two cross-sectional views $f_\theta(z=0.2)$ and $f_\theta(z=0.8)$.} 
    \label{fig:lm_8}
\end{figure}

We start with simple landmark matching tasks to demonstrate the feasibility of our framework. Consider two groups of points in $\mathbb{R}^3$ denoted by $(A,B,C,D)$ and $(E,F,G,H)$ as illustrated in Fig.~\ref{fig:lm_8_dem}. For each group of points, we assign each point to its neighbour, resulting in two large-scale twists in an anti-clockwise manner at different angles.

As shown in Fig.~\ref{fig:lm_8}, the resulting solution map $f_\theta$ is a smooth $3$D mapping that satisfies the landmark correspondence and the boundary conditions. In Fig.~\ref{fig:lm_8}(c), the histogram represents the values of $\det\nabla f_\theta$ at 100,000 randomly sampled points. This histogram provides two important insights: (i) The inclusion of the bijectivity loss guarantees that $f_\theta$ is a bijective mapping over the entire domain $\Omega_T$. (ii) The conformality loss efficiently guides the mapping $f_\theta$ such that the determinant of the Jacobian matrix, $\det \nabla f_\theta$, remains close to unity for the majority of points. Therefore, our model is capable of producing a landmark matching diffeomorphic $3$D mapping that perfectly satisfies the boundary conditions. 

The landmark loss, conformality loss and smoothness loss of the trained model are listed in Table~\ref{tab:lm_8_ball_rotate}. It is noteworthy that the prescribed landmarks are located near the boundary, which potentially counteracts with the boundary condition during optimization. Still, our model applies the hard constraint approach \eqref{eqn:hardboundary} and will not suffer from this issue. Further ablation study demonstrate such an advantage in will be presented in Section~\ref{sec:ablation_study_boundary} by comparing against models with more commonly used soft boundary conditions.

\subsubsection{Axis-Rotated Sphere}
\label{sec:rotate}
\begin{figure}[!htbp]
    \centering
    
    \subfloat[\label{fig:rot_sphere_dem}]{\includegraphics[width=.32\textwidth]{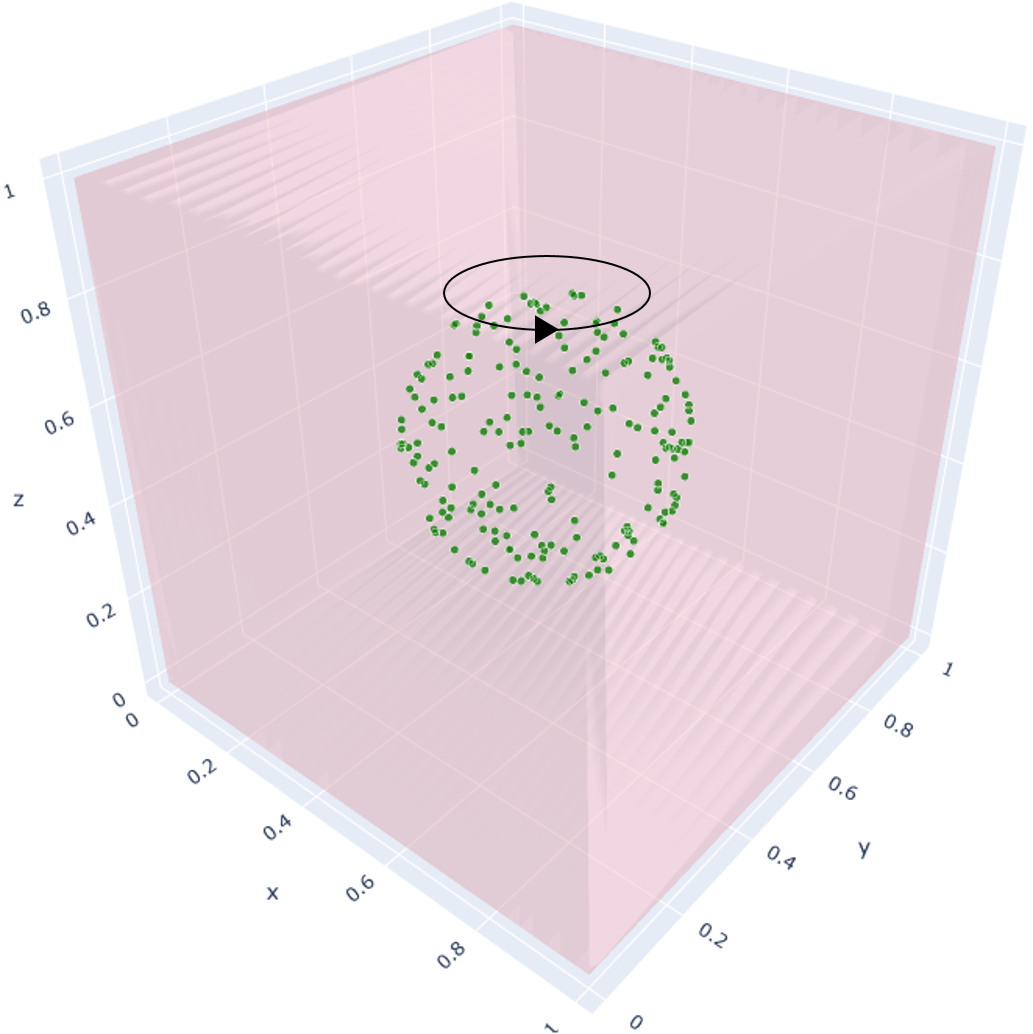}}\hspace{.1em}
    \subfloat[]{\includegraphics[width=.35\textwidth]{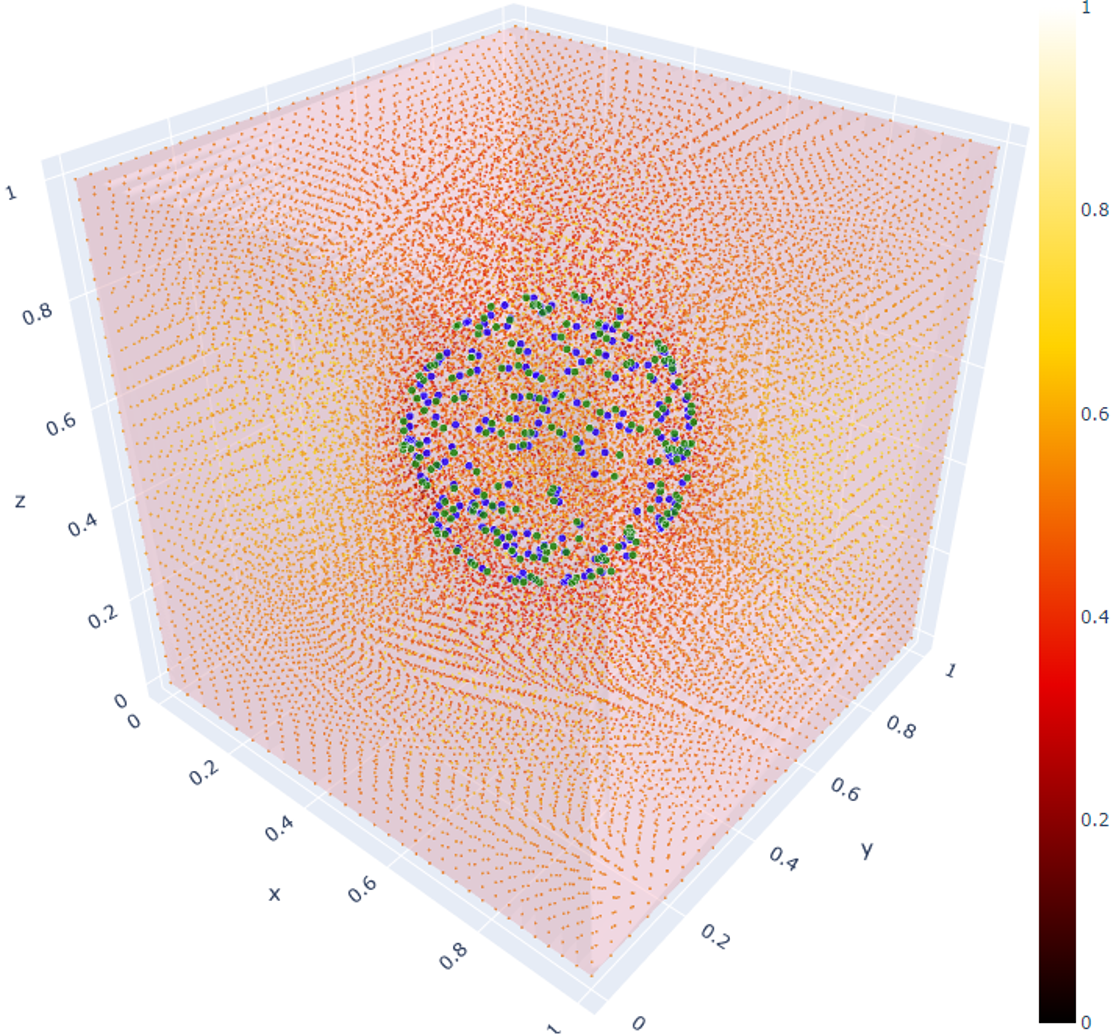}}\hspace{.1em}
    \subfloat[]{\includegraphics[width=.28\textwidth]{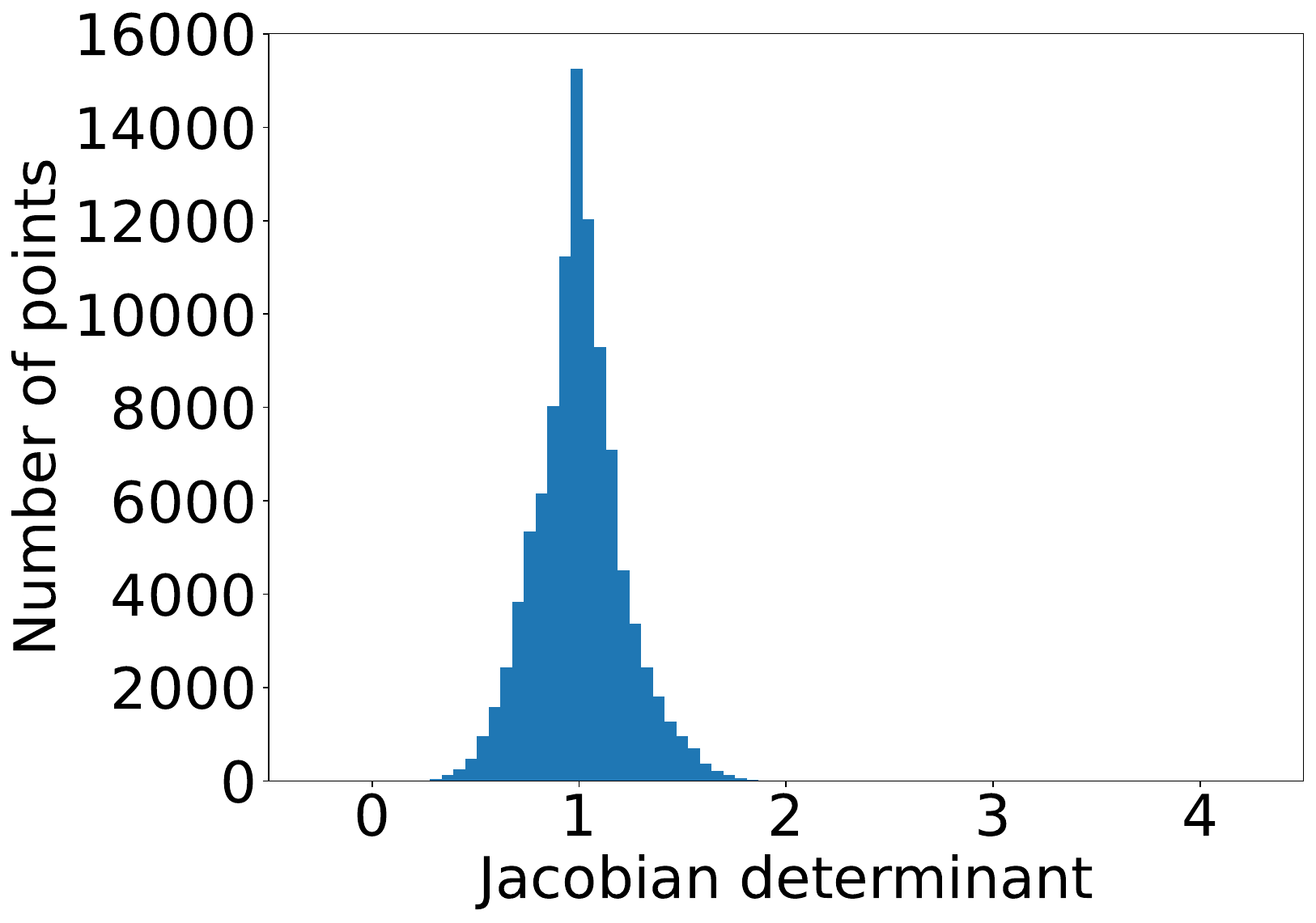}}\hspace{.1em}

    \subfloat[]{\includegraphics[width=.32\textwidth]{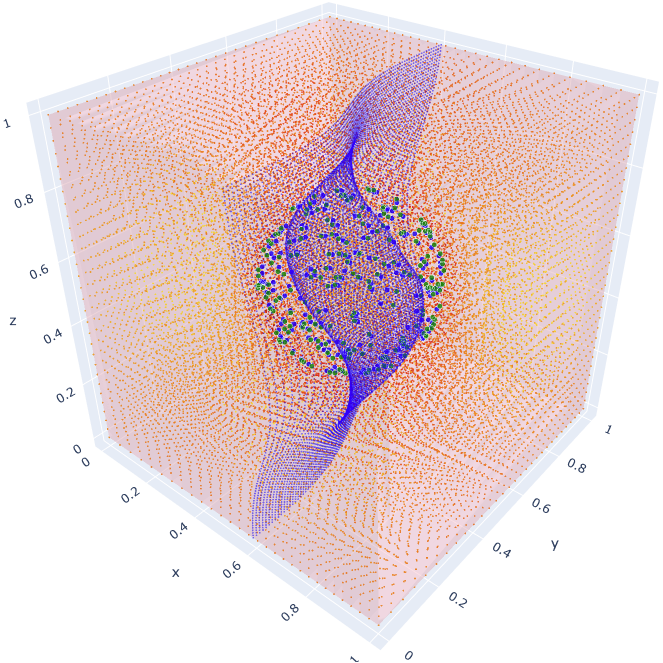}}\hspace{.1em}
    \subfloat[]{\includegraphics[width=.32\textwidth]{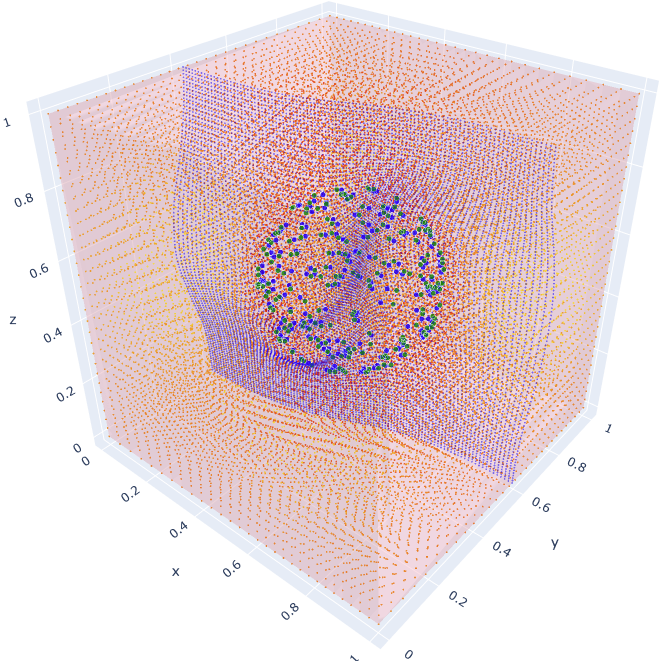}}\hspace{.1em}
    \subfloat[]{\includegraphics[width=.32\textwidth]{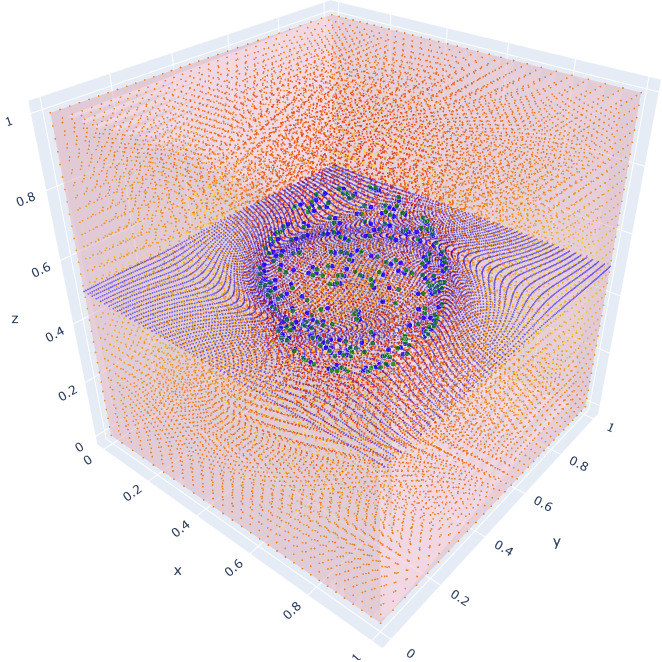}}\hspace{.1em}
    
    \caption{Results of our proposed framework on the rotated sphere example in Section~\ref{sec:rotate}. (a) Visualization of the rotated sphere. (b) The $3$D transformation $f_\theta$ obtained by the trained network. 
    The color at each point is defined as in Fig.~\ref{fig:lm_8}. (c) Histogram of $\det\nabla f_\theta(p)$. (d) Cross-sectional view $f_\theta(x=0.5)$. (e) Cross-sectional view $f_\theta(y=0.5)$. (f) Cross-sectional view $f_\theta(z=0.5)$.} 
    \label{fig:ball_rotate}
\end{figure}

To stress test our proposed framework and demonstrate its scalability to more complex information, a larger number of landmarks are chosen and landmark matching is performed under identical network architecture as in Section~\ref{sec:lm_8}. To define the landmarks, we first consider a ball of radius $0.25$ units positioned at the center of the domain $\Omega_S = \Omega_T = [0, 1]^3$. The sphere of the ball is rotated $90^{\circ}$ anticlockwise with respect to the line $(x,y) = (0.5,0.5)$ (cf. Fig.~\ref{fig:rot_sphere_dem}). The landmarks information is then generated by matching the initial and rotated positions of $200$ randomly sampled points on the sphere.

As shown in Fig. \ref{fig:ball_rotate}, our model can handle extremely large distortion while satisfying the landmark constraint and boundary conditions. This suggests that our proposed method has great scalability between number of parameters and complexity of the given information. The histogram presented in Fig. \ref{fig:ball_rotate}(c), depicting the values of $\det \nabla f_\theta$, serves as evidence for the effectiveness of the conformality loss and the bijectivity loss in ensuring that the mapping $f_\theta$ is both bijective and minimally distorted in terms of conformality. The training loss of the Twisted Landmark Pairs case and the Axis-Rotated Sphere case can be found in Table~\ref{tab:lm_8_ball_rotate}.

\begin{table}[!htbp]
    \centering
    \begin{tabular}{lcc}
        \hline
        & Twisted Landmark Pairs & Axis-Rotated Sphere \\
        \hline
        Landmark loss & 2.8815e-4 & 9.2231e-5  \\ 
        Conformality loss & 1.3166e0 & 1.3307e0  \\ 
        Smoothness loss & 3.2233e1 & 2.5543e1  \\
        \hline
    \end{tabular}
    \caption{The landmark loss, conformality loss and smoothness loss of the Twisted Landmark Pairs example and the Axis-Rotated Sphere example \ref{sec:lm_8}. Losses are extracted from the last epoch of training.}
    \label{tab:lm_8_ball_rotate}
\end{table}

\subsubsection{Large Distortion Mapping} \label{sec:low_freq}
In this example, we test the three formulations introduced in Section~\ref{sec:three_formulations} with larger distortion mapping, this aims at demonstrating the interplay between three losses. Consider a $3$D mapping $g: \Omega_T \rightarrow \Omega_S$ that is of large distortion. We generate the landmark data and image data using $g$. For detailed construction of the $N_{\rm lm}=512$ landmark pairs and the image pairs $S:[0,1]^3 \rightarrow \mathbb{R}$ and $T := S \circ g$ of the same size $128^3$, please refer to Appendix~\ref{appendix:synthetic_map}.

Note that the synthetic mapping $g$ is not exactly diffeomorphic by its definition. So the resulting map $f_\theta$ trained by landmarks and/or images generated from $g$ is not supposed to be identical to non-diffeomorphic $g$, at least in the region where $\det\nabla g \leq 0$.



\begin{figure}[!htbp]
    \centering
    
    \subfloat[$f_\theta$: Landmark-matching]{\includegraphics[width=.32\textwidth]{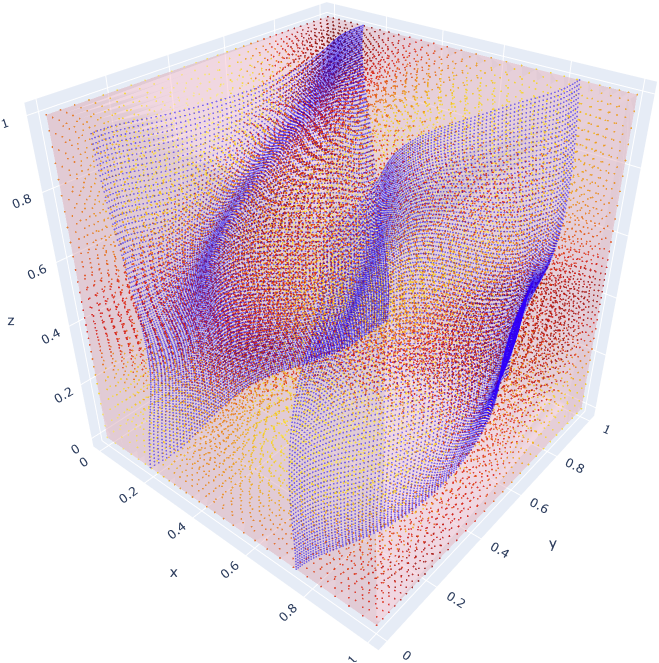}}\hspace{.1em}
    \subfloat[$f_\theta$: Intensity-matching]{\includegraphics[width=.32\textwidth]{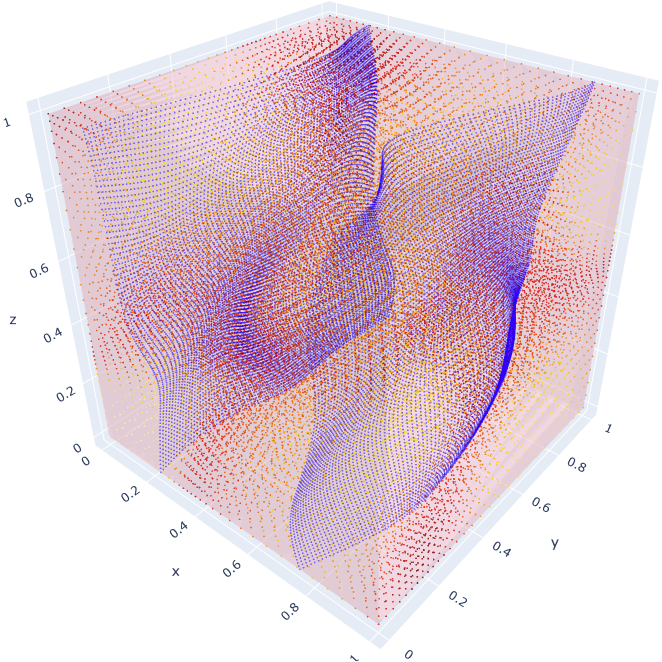}}\hspace{.1em}
    \subfloat[$f_\theta$: Hybrid-matching]{\includegraphics[width=.32\textwidth]{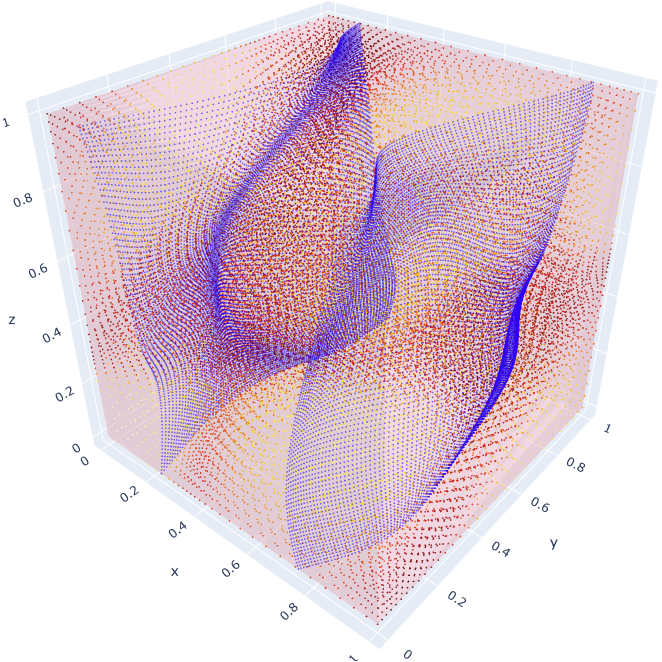}}\hspace{.1em}
    
    \subfloat[$\det \nabla f_\theta$: Landmark-matching]{\includegraphics[width=.32\textwidth]{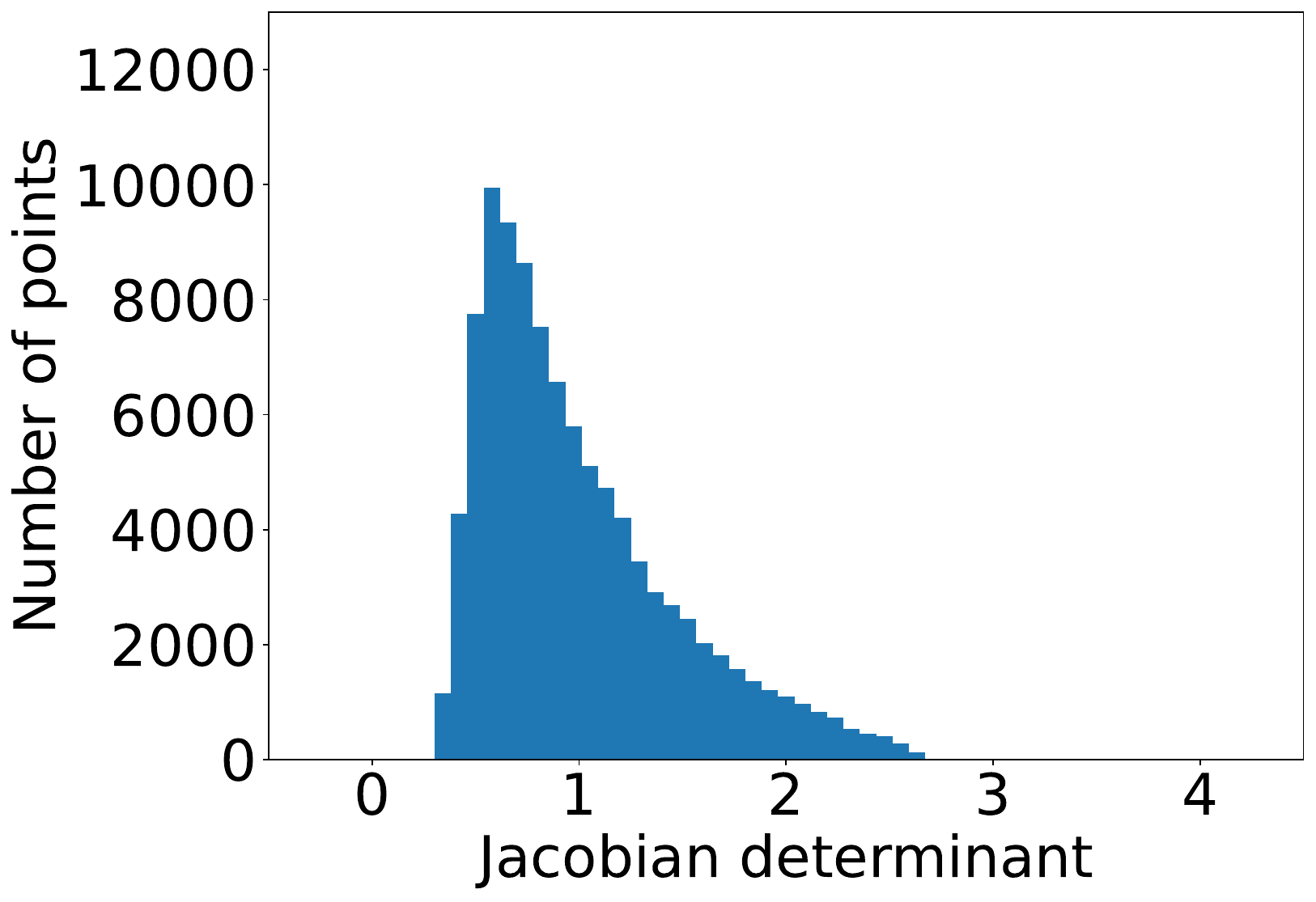}}\hspace{.1em}
    \subfloat[$\det \nabla f_\theta$: Intensity-matching]{\includegraphics[width=.32\textwidth]{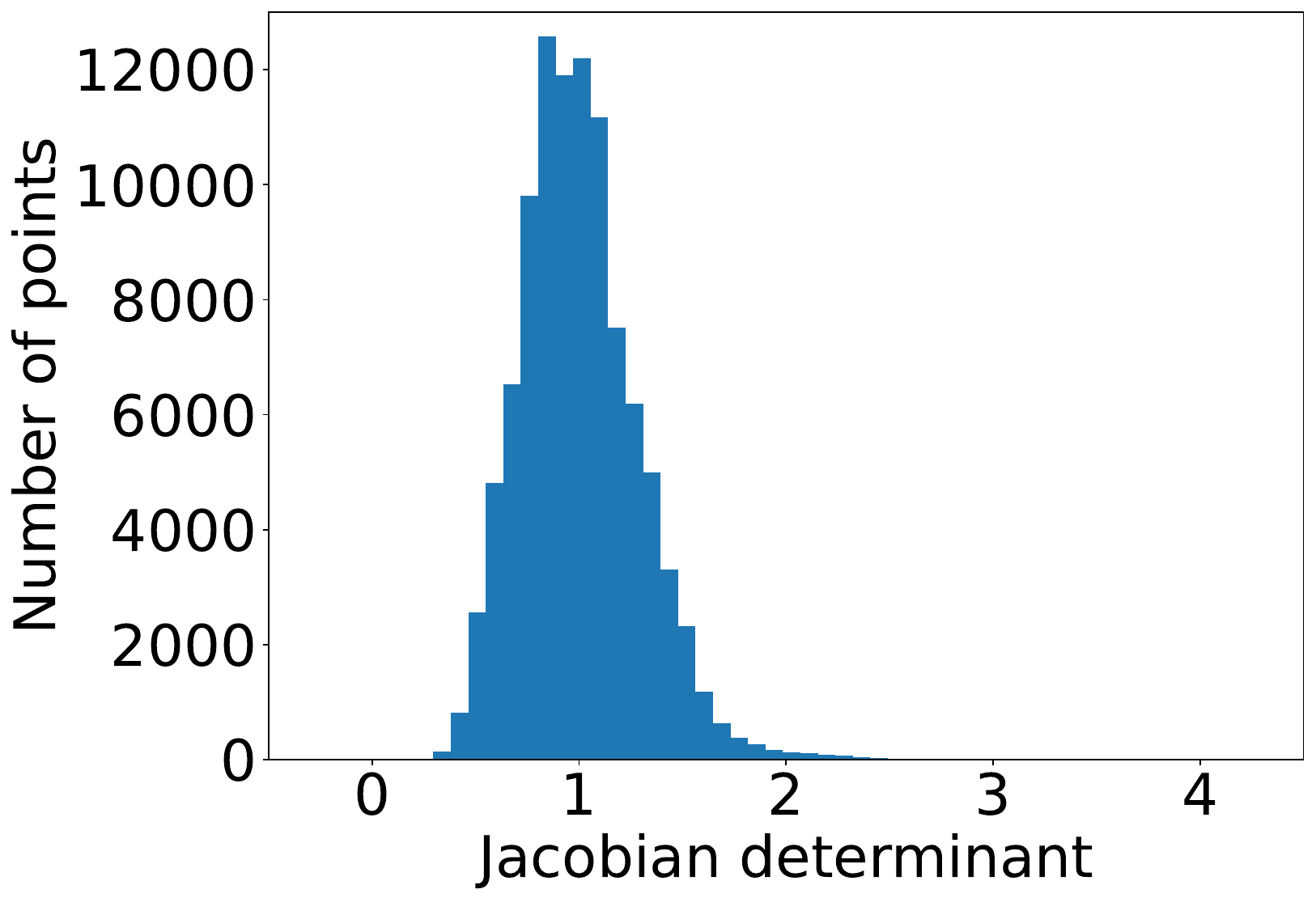}}\hspace{.1em}
    \subfloat[$\det \nabla f_\theta$: Hybrid-matching]{\includegraphics[width=.32\textwidth]{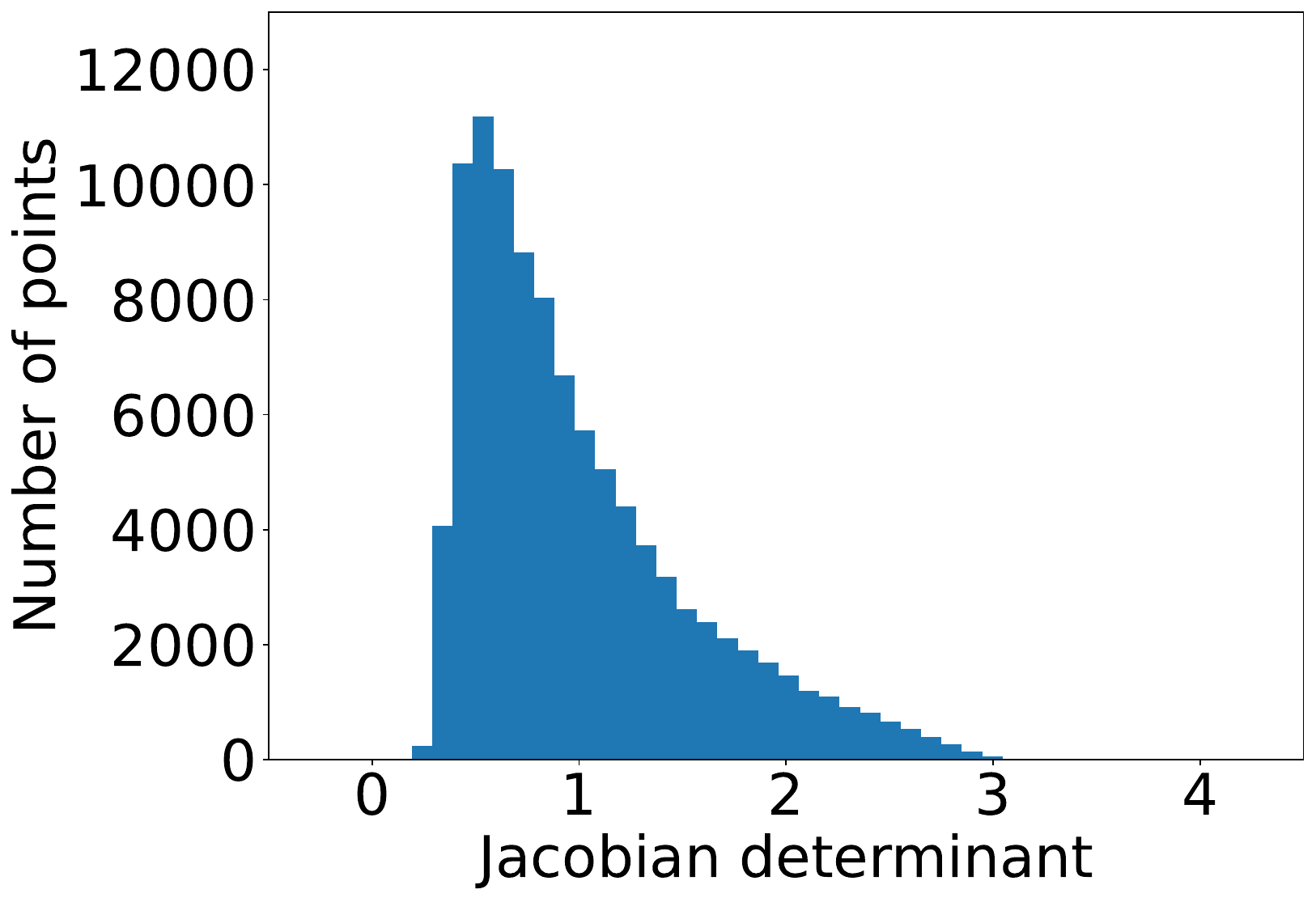}}\hspace{.1em}
    
    \subfloat[Conformality loss]{\includegraphics[width=.32\textwidth]{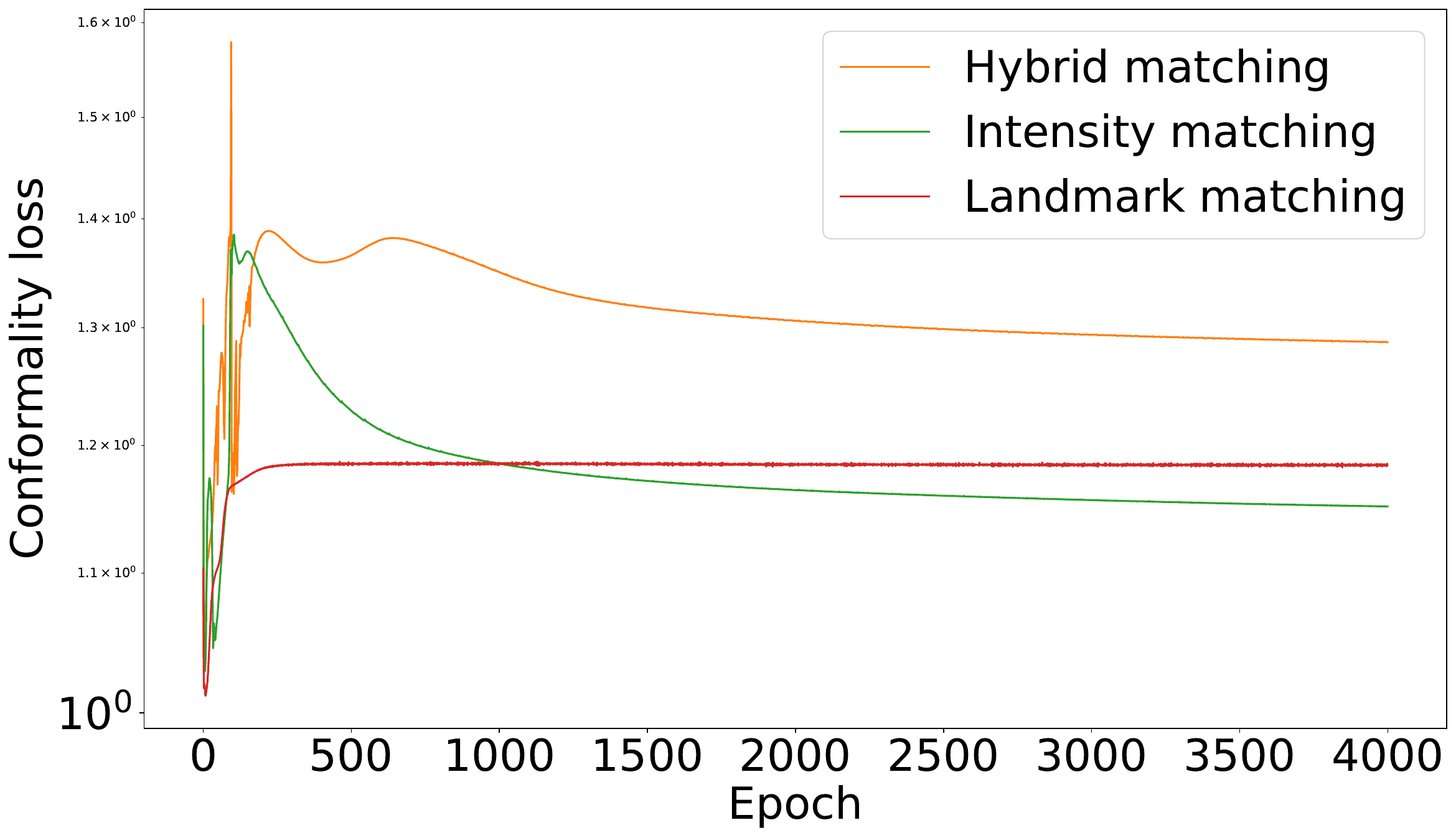}}\hspace{.1em}
    \subfloat[Intensity loss]{\includegraphics[width=.32\textwidth]{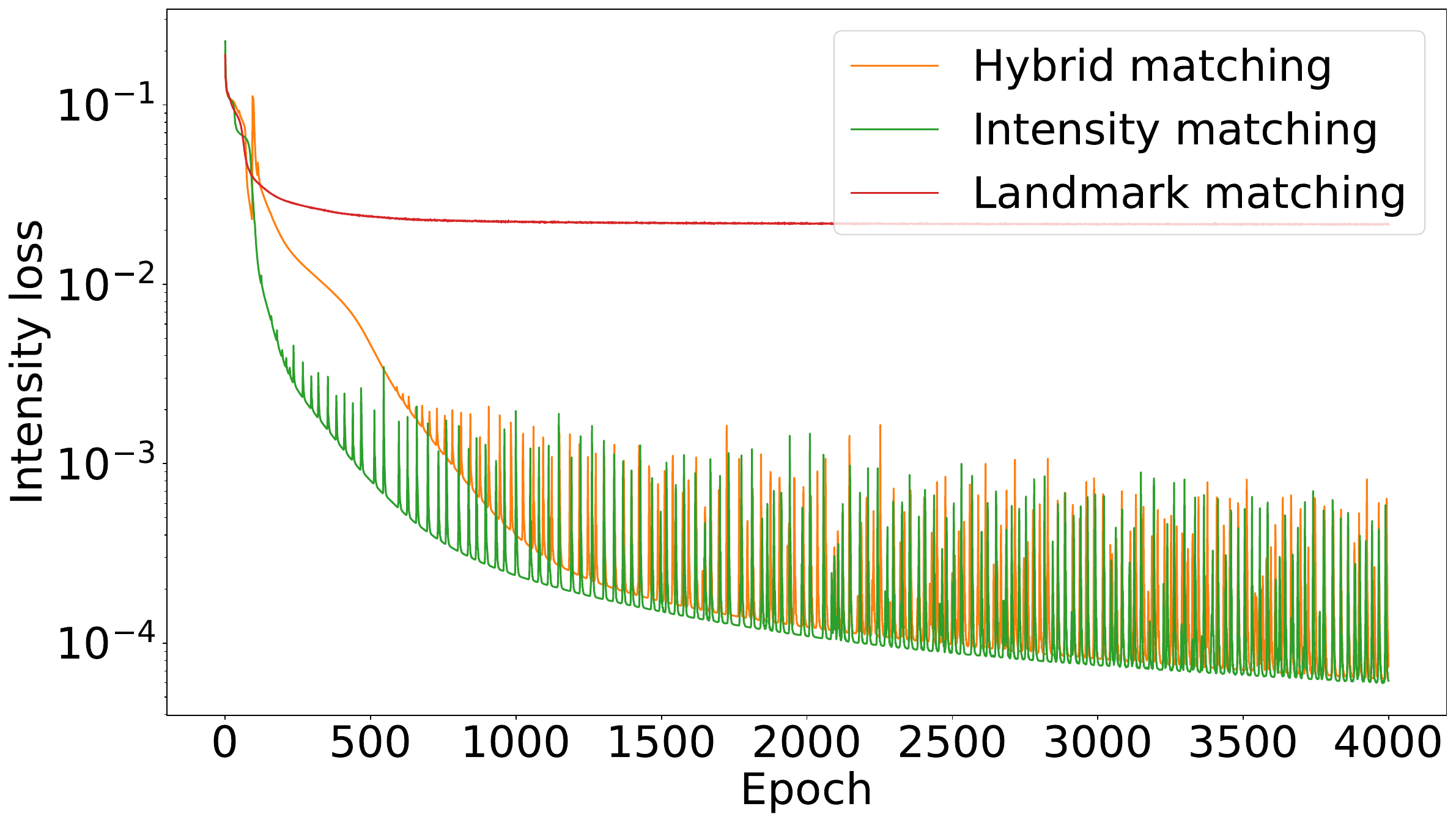}}\hspace{.1em}
    \subfloat[Landmark loss]{\includegraphics[width=.32\textwidth]{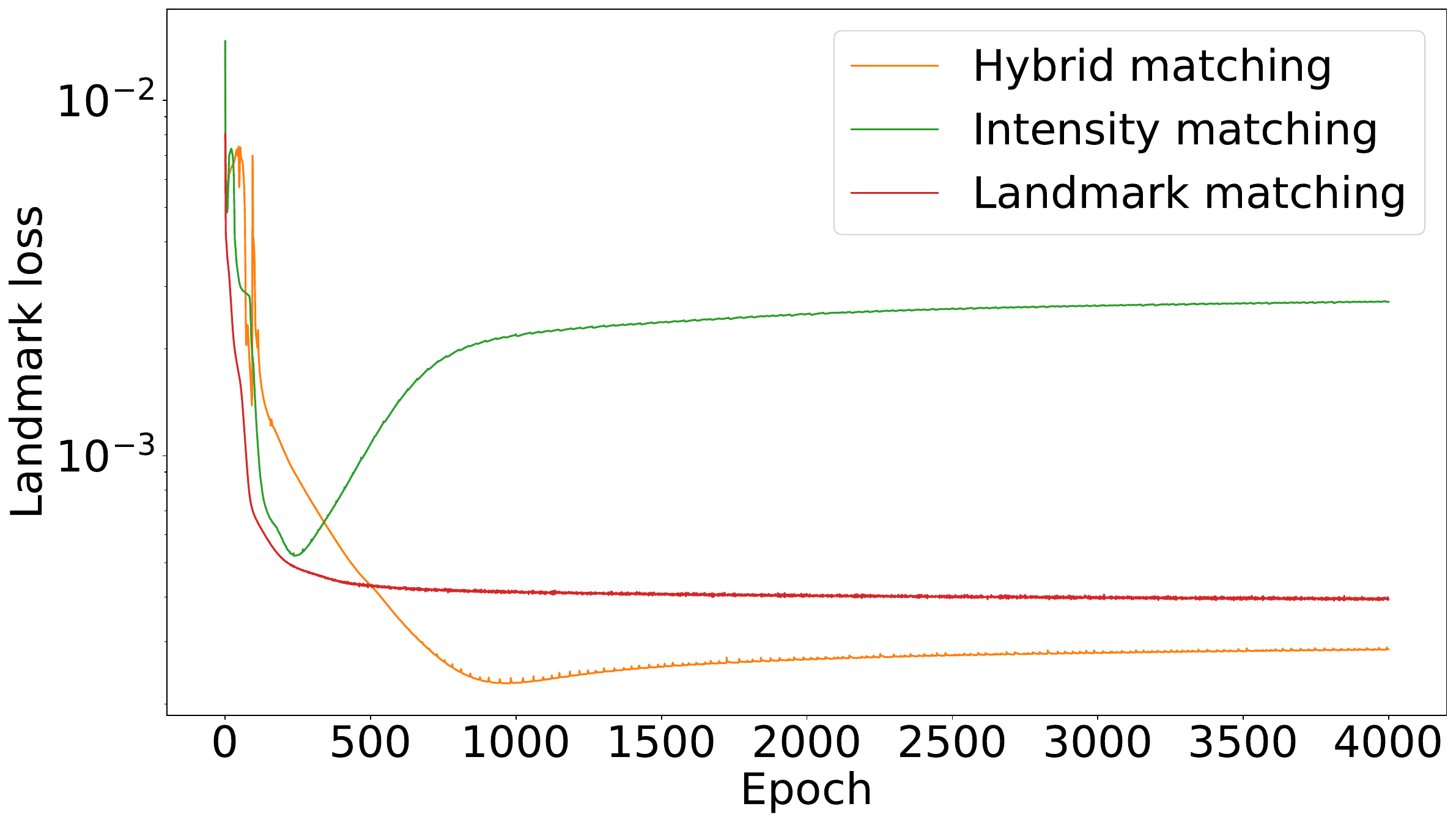}}\hspace{.1em}
    
    \caption{Results of the Large Distortion Mapping example. (a)-(c) visualize the resulting mappings $f_\theta$, with two cross-sectional views $f_\theta(x=0.2),\, f_\theta(x=0.8)$ based on the three formulations in Section~\ref{sec:three_formulations}. The color at each point is defined the same as in Fig.~\ref{fig:lm_8}. (d)-(f) show the histograms of $\det \nabla f_\theta$ obtained from the three formulations. (g)-(i) display the conformality loss, intensity loss and landmark loss during training of the three formulations, respectively.} 
    \label{fig:low_freq_maps}
\end{figure}

\begin{figure}[!htbp]
    \centering
    
    \subfloat[Source image $S$ ]{\includegraphics[width=.49\textwidth]{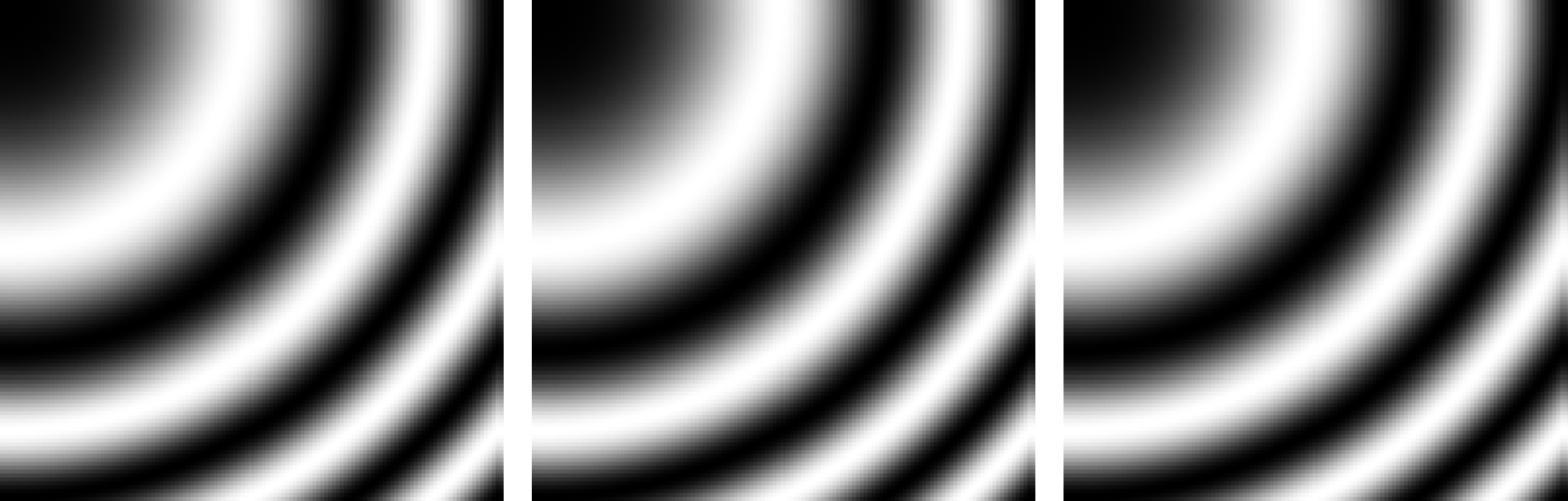}}\hspace{.1em}
    
    \subfloat[Target image $T$ ]{\includegraphics[width=.49\textwidth]{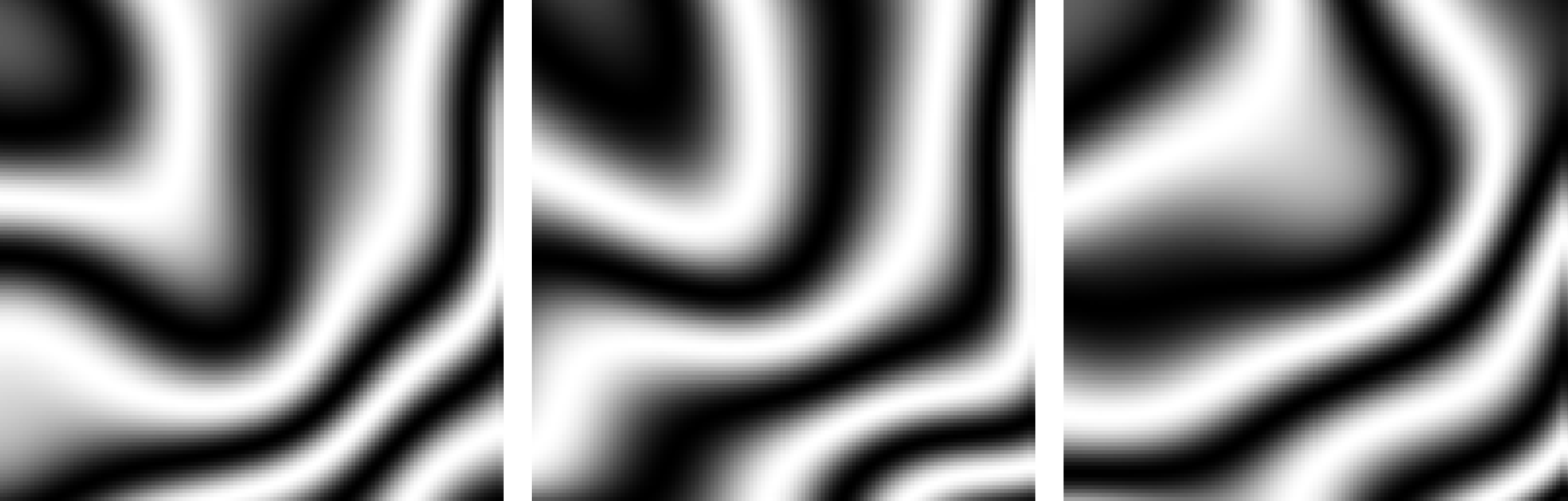}}\hspace{.1em}
    \subfloat[$|S-T|$]{\includegraphics[width=.49\textwidth]{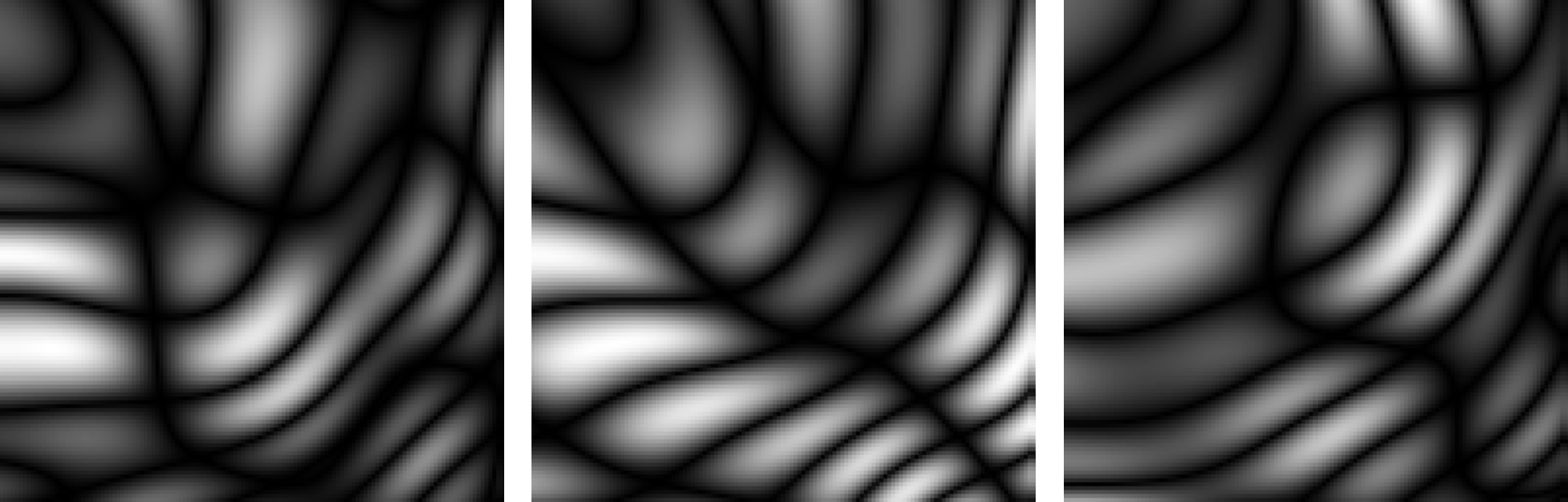}}\hspace{.1em}
    
    \subfloat[$S\circ f_\theta$: Landmark-matching ]{\includegraphics[width=.49\textwidth]{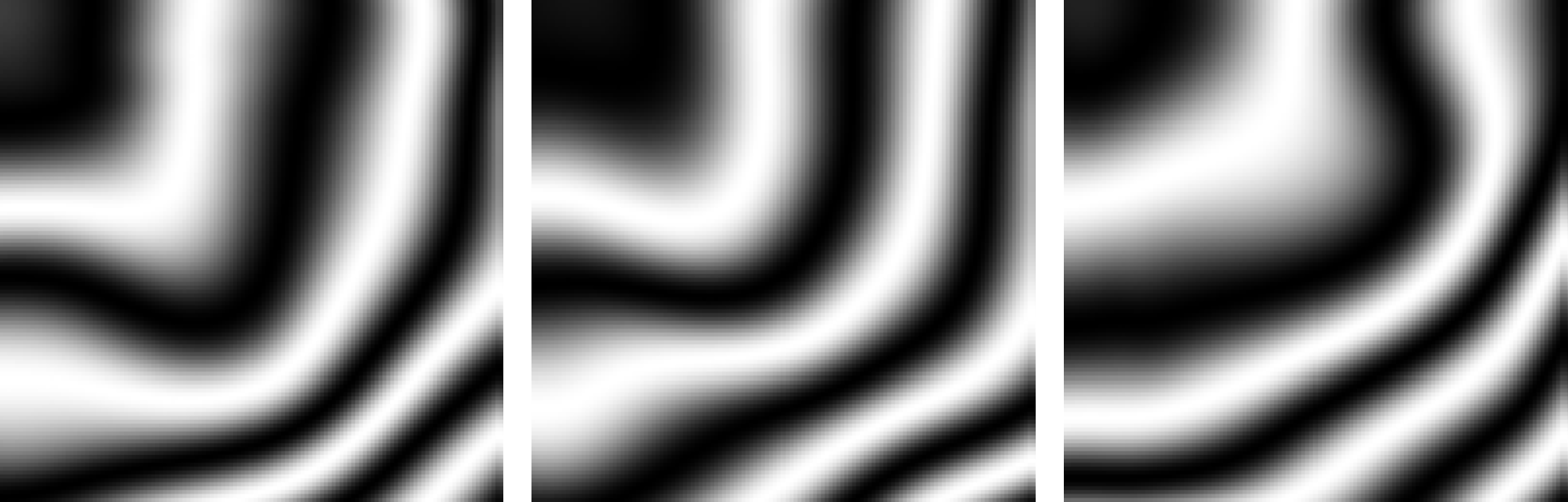}}\hspace{.1em}
    \subfloat[$|S\circ f_\theta - T|$ : Landmark-matching ]{\includegraphics[width=.49\textwidth]{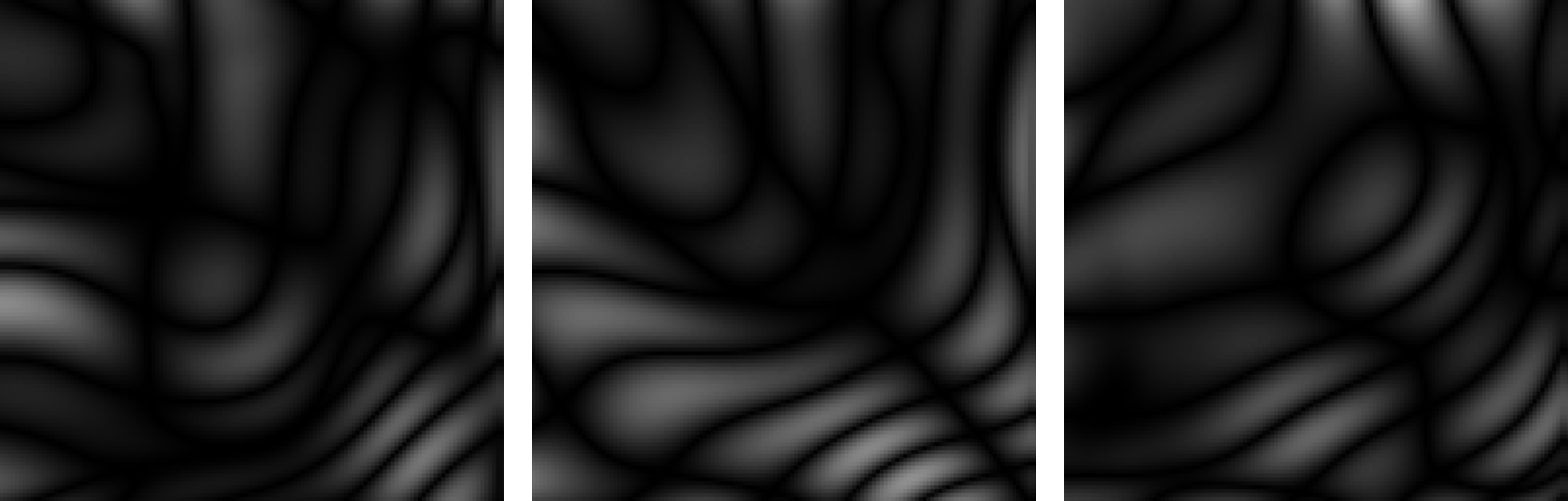}}\hspace{.1em}

    \subfloat[$S\circ f_\theta$: Intensity-matching ]{\includegraphics[width=.49\textwidth]{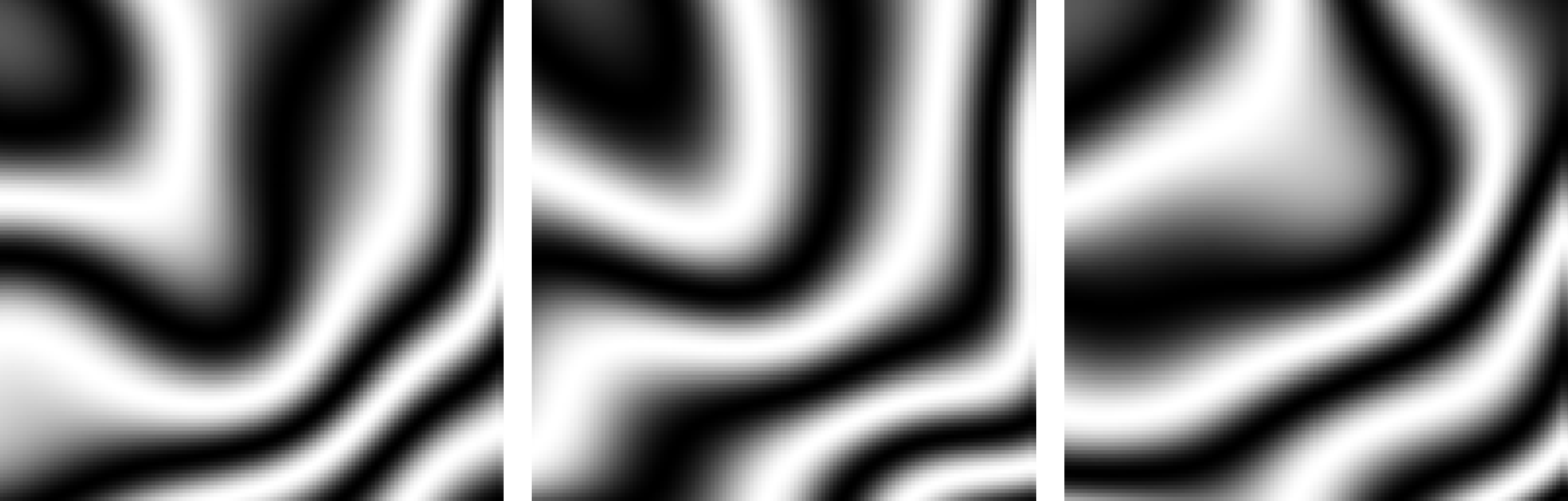}}\hspace{.1em}
    \subfloat[$|S\circ f_\theta - T|$: Intensity-matching]{\includegraphics[width=.49\textwidth]{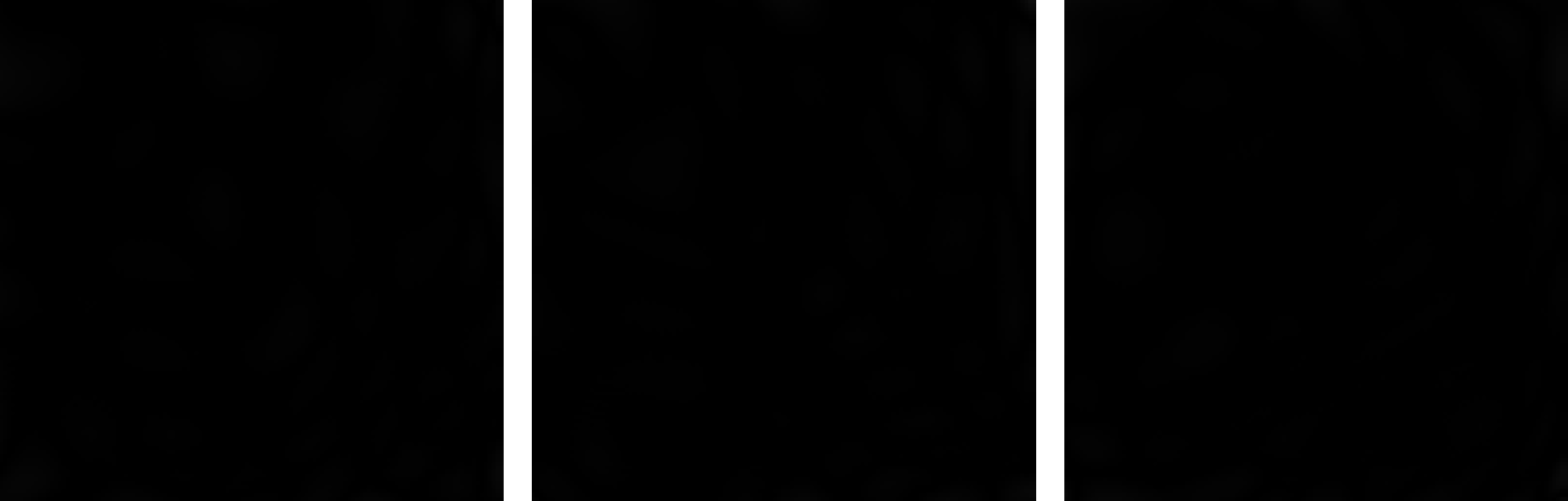}}\hspace{.1em}
    
    \subfloat[$S\circ f_\theta$: Hybrid-matching ]{\includegraphics[width=.49\textwidth]{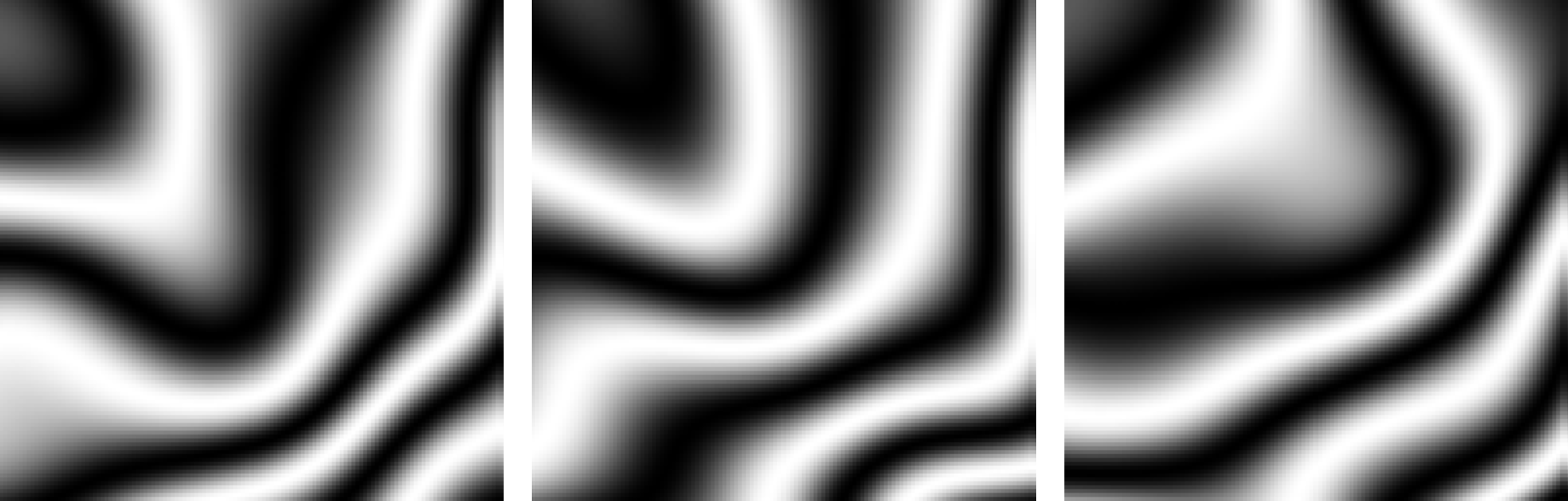}}\hspace{.1em}
    \subfloat[$|S\circ f_\theta - T|$: Hybrid-matching ]{\includegraphics[width=.49\textwidth]{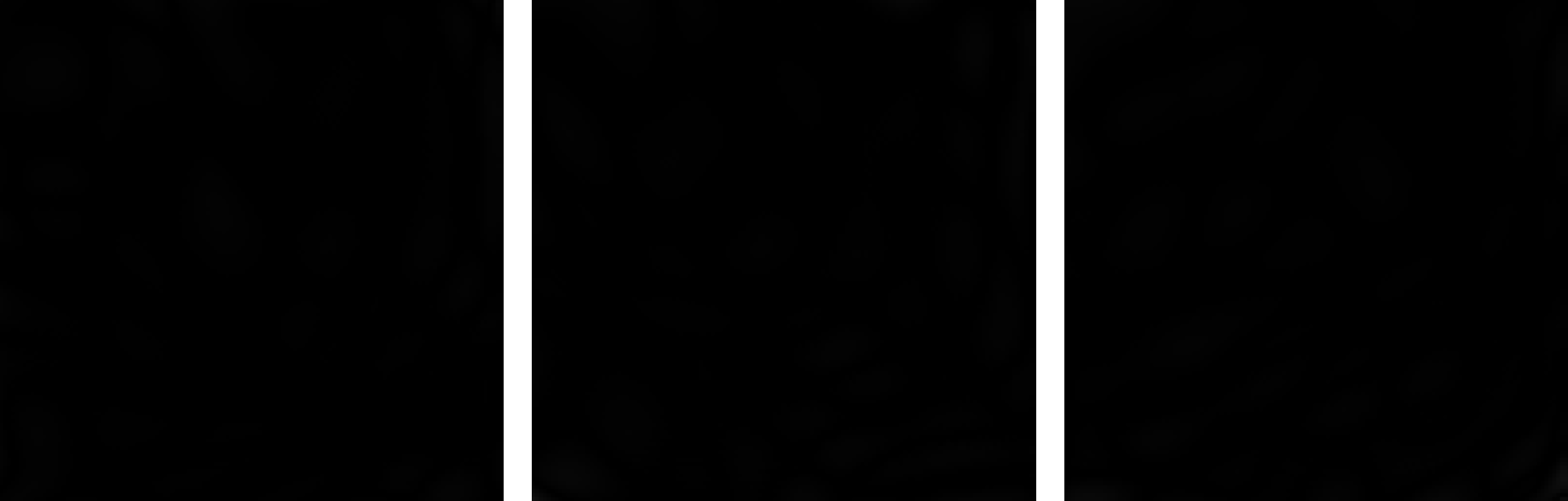}}\hspace{.1em}

    \caption{The Large Distortion Mapping example. Visualization of the registration results via three views, i.e. $x=0.5,\,y=0.5,\,z=0.5$. (a)-(b) Three views (slices) of the source image $S$ and target images $T$ respectively. (c) Three views of the absolute difference between $S$ and $T$. (d)-(e) The landmark matching registration results. (f)-(g) The intensity matching registration results. (h)-(i) The hybrid matching registration results.}
    \label{fig:low_freq_slices}
\end{figure}

As shown in Fig. \ref{fig:low_freq_maps} and Fig. \ref{fig:low_freq_slices}, the intensity matching formulation is comparable to the hybrid matching formulation, both of which perform better than the landmark matching formulation, in a sense that the hybrid matching formulation achieves a smaller landmark error than the landmark matching formulation and an intensity error that is comparable to the intensity matching formulation (cf. Fig. \ref{fig:low_freq_maps}(h)(i)) on top of the fact that the intensity information is \textit{consistent} with landmark information (cf. Table \ref{tab:low_freq}). The hybrid and intensity matching formulation is also qualitatively more similar to $g$ (cf. Fig.~\ref{fig:synthetic_maps}). Additionally, Fig.~\ref{fig:low_freq_slices} shows the slices from three views of the source image $S$, target image $T$ and the warped image $S\circ f_\theta$ under the three formulations. 

It is observed that both the hybrid and intensity matching formulations are more sensitive to the intensity loss than to the landmark loss, giving rise to the fluctuation of intensity loss during training. However, this fluctuation eventually stabilize within the range $[10^{-4}, 10^{-3}]$, which is much smaller than the landmark matching formulation, indicating that the intensity and hybrid matching formulations exhibit excellent performance in matching the intensity, while the landmark matching formulation performs worse due to insufficient information contained in the landmark correspondence.


\begin{table}[!htbp]
    \begin{tabular}{lccc}
        \hline
         & Landmark matching & Intensity matching & Hybrid matching \\
         \hline
         Landmark loss & 3.8660e-4 & 2.8165e-3 & 2.9553e-4 \\ 
         Intensity loss & 2.1936e-2 & 3.5523e-5 & 3.5238e-5 \\ 
         Conformality loss & 1.1830e0 & 1.1447e0 & 1.2774e0 \\ 
         Smoothness loss & 1.5059e1 & 1.3695e1 & 2.5872e1 \\
         \hline
    \end{tabular}
    \caption{The landmark loss, intensity loss, conformality loss and smoothness loss of three formulations for the large distortion example \ref{sec:low_freq}. Losses are extracted from the last epoch of training.}
    \label{tab:low_freq}
\end{table}

\subsubsection{Volume vs. Angle Preservation for Translating Disk}

\begin{figure}[htbp]
    \centering
    \begin{subfigure}[b]{0.32\textwidth}
        \includegraphics[width=\textwidth]{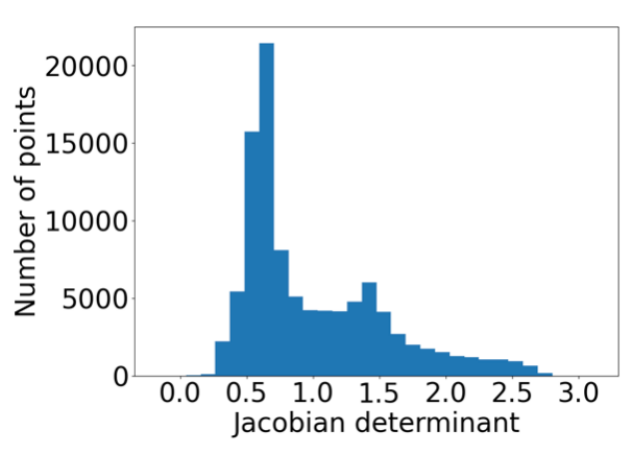}
        \caption{}
        \label{fig:jac_1}
    \end{subfigure}
    \hfill
    \begin{subfigure}[b]{0.32\textwidth}
        \includegraphics[width=\textwidth]{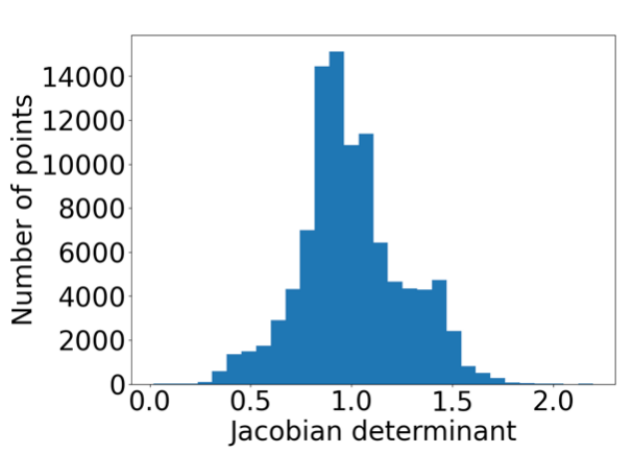}
        \caption{}
        \label{fig:jac_2}
    \end{subfigure}
    \hfill
    \begin{subfigure}[b]{0.32\textwidth}
        \includegraphics[width=\textwidth]{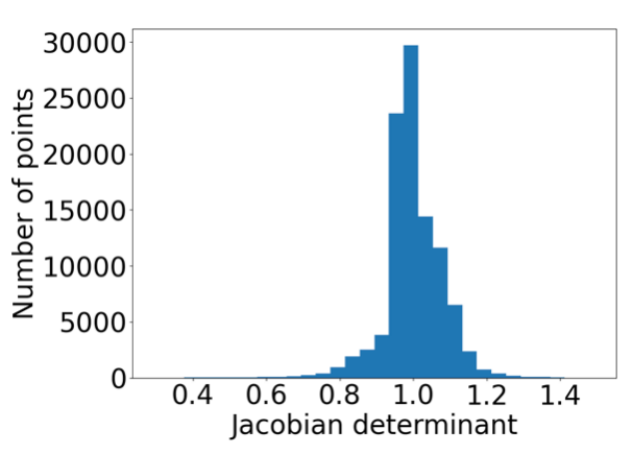}
        \caption{}
        \label{fig:jac_3}
    \end{subfigure}
    
    \vspace{1em}
    
    \begin{subfigure}[b]{0.32\textwidth}
        \includegraphics[width=\textwidth]{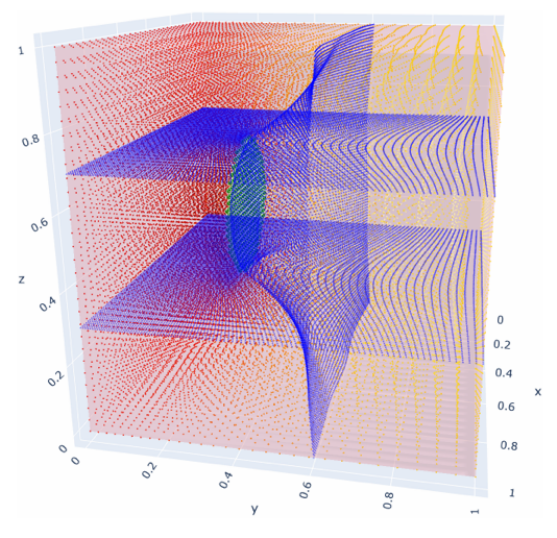}
        \caption{}
        \label{fig:box_1}
    \end{subfigure}
    \hfill
    \begin{subfigure}[b]{0.32\textwidth}
        \includegraphics[width=\textwidth]{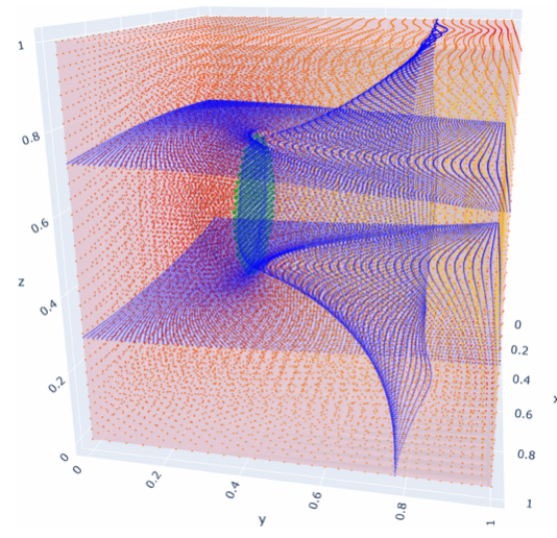}
        \caption{}
        \label{fig:box_2}
    \end{subfigure}
    \hfill
    \begin{subfigure}[b]{0.32\textwidth}
        \includegraphics[width=\textwidth]{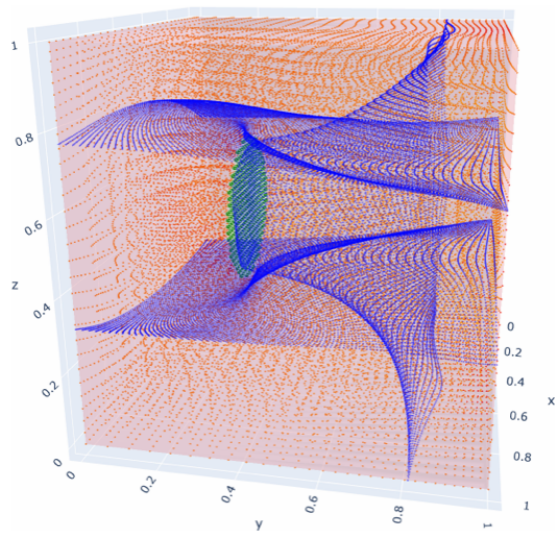}
        \caption{}
        \label{fig:box_3}
    \end{subfigure}
    
    \caption{Comparison of Jacobian distributions represented as bar charts (top row) and deformation maps with $3$ cross-sectional view $f_\theta(z=0.2),f_\theta(z=0.8),f_\theta(y=0.7)$ (bottom row). Note that the parameter $\alpha_3$ denotes the weight of the conformality loss, while $\alpha_4$ corresponds to the weight of the volumetric-prior loss. The left column shows results for $\alpha_3 = 1, \alpha_4 = 0$, the middle column for $\alpha_3 = 0, \alpha_4 = 1$, and the right column for $\alpha_3 = 0, \alpha_4 = 10$.}
    \label{fig:jacobian_boxplots}
\end{figure}

In this section, we demonstrate the capability of the proposed model to incorporate a volumetric preserving prior. Specifically, we consider a disk of radius $0.25$ centered at $(0.7, 0.5, 0.5)$, initially positioned on the plane $y = 0.7$, which is translated vertically to $y = 0.3$. Figure~\ref{fig:jacobian_boxplots} presents results from applying our method to this moving disk, labeled with 400 landmarks, under varying weights of the conformal and volumetric prior losses.

When only the conformal loss is applied and the volumetric prior is omitted, the resulting deformation preserves a stable mapping along planes orthogonal to the $y$-axis. This is due to the uniform displacement vector $v = (0, -0.4, 0)$ applied to all landmarks, which causes each point to follow the same directional shift, preserving planar structure.

However, when the conformal loss is excluded and the volumetric prior weight is increased, the model resists local compression and expansion. As a result, hyperplanes such as $f(z = 0.2)$ exhibit non-uniform deformation. This behavior is expected: regions with $y < 0.3$ undergo compression due to the disk's movement, but the volume-preserving constraint prevents such compression. Consequently, these points are redistributed toward spatially sparse regions, leading to instability in the mapping of certain hyperplanes (e.g. $f(z = 0.2)$ or $f(z = 0.8)$).

In Figure~\ref{fig:jacobian_boxplots}, the color of each point is determined by the Jacobian determinant, which serves as a visualization of density. The colormap ranges from yellow to red, corresponding to regions of sparser to denser density, respectively. In the conformal case, the warped point density is highly uneven, with significant crowding in the region $y < 0.3$. In contrast, the volume-preserving setting yields a more uniform distribution, especially when higher weights are assigned to the volumetric prior.

Finally, the range of the $y$-coordinate on the surface $f(y = 0.7)$ increases with the volume-preserving weight. This expansion, and the above observations, reflects the redistribution of points driven by the model's effort to enforce uniform density through volumetric regularization.

\subsection{Application on Medical Images Registration}
\label{sec:4DCT_exp}
In this section, we show the capability of the proposed three formulations in Section~\ref{sec:three_formulations} in real data. The 4DCT dataset, consisting of two lung CT scans featuring $300$ pairs of landmarks, is chosen for demonstration. This dataset can be accessed through the Deformable Image Registration Laboratory (www.dir-lab.com). 

\begin{figure}[!htbp]
    \centering
    \subfloat[$f_\theta$: Landmark-matching]{\includegraphics[width=.32\textwidth]{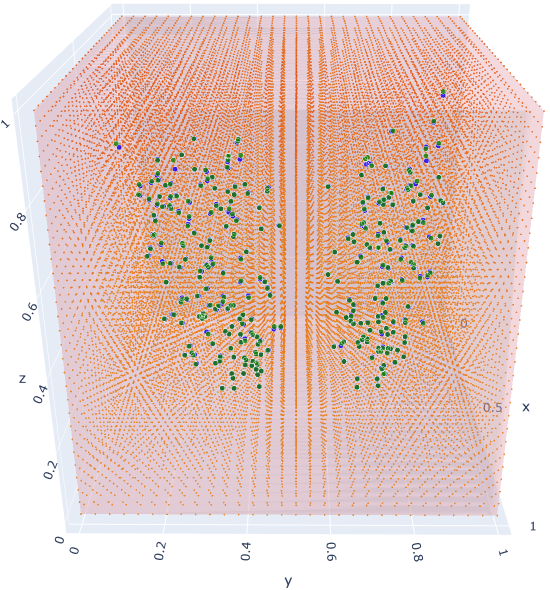}}\hspace{.1em}
    \subfloat[$f_\theta$: Intensity-matching]{\includegraphics[width=.32\textwidth]{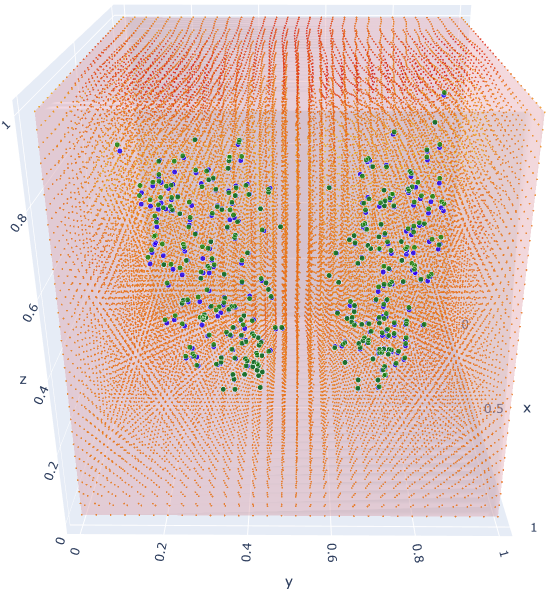}}\hspace{.1em}
    \subfloat[$f_\theta$: Hybrid-matching]{\includegraphics[width=.32\textwidth]{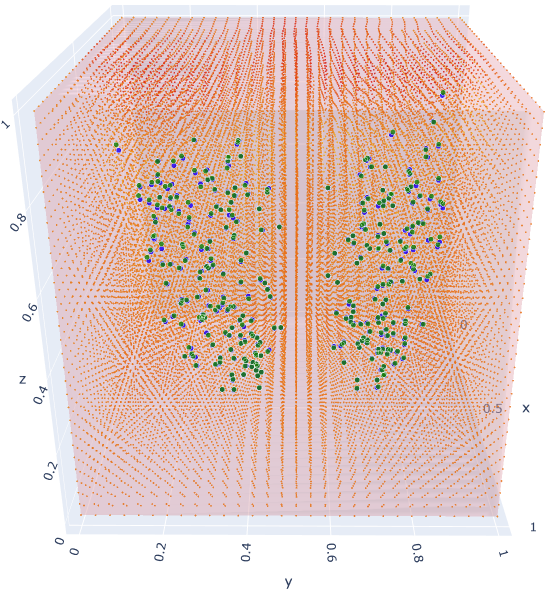}}\hspace{.1em}
    
    \subfloat[$\det{\nabla f_\theta}$:~Landmark-matching]{\includegraphics[width=.32\textwidth]{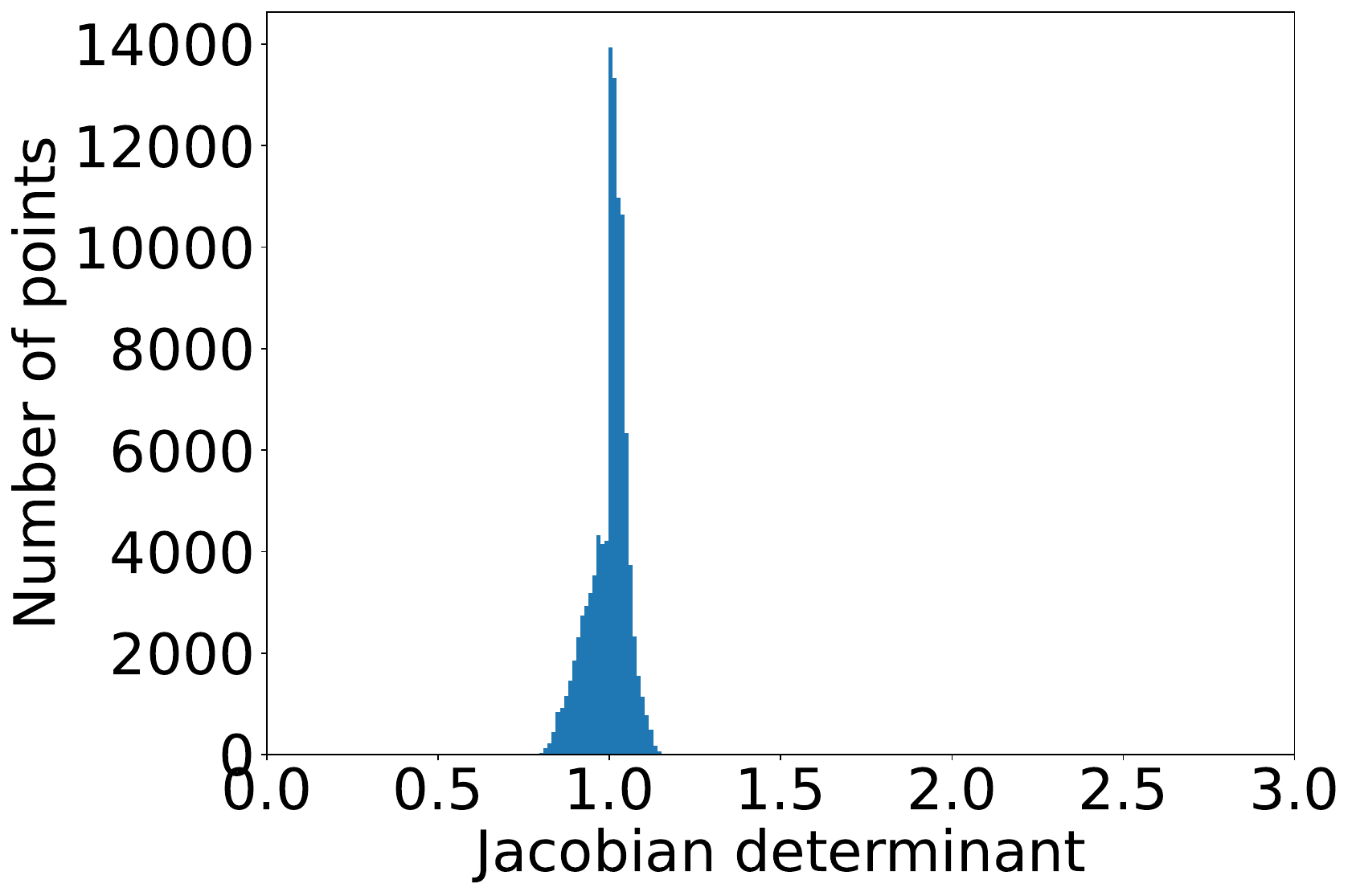}}\hspace{.1em}
    \subfloat[$\det{\nabla f_\theta}$:~Intensity-matching]{\includegraphics[width=.32\textwidth]{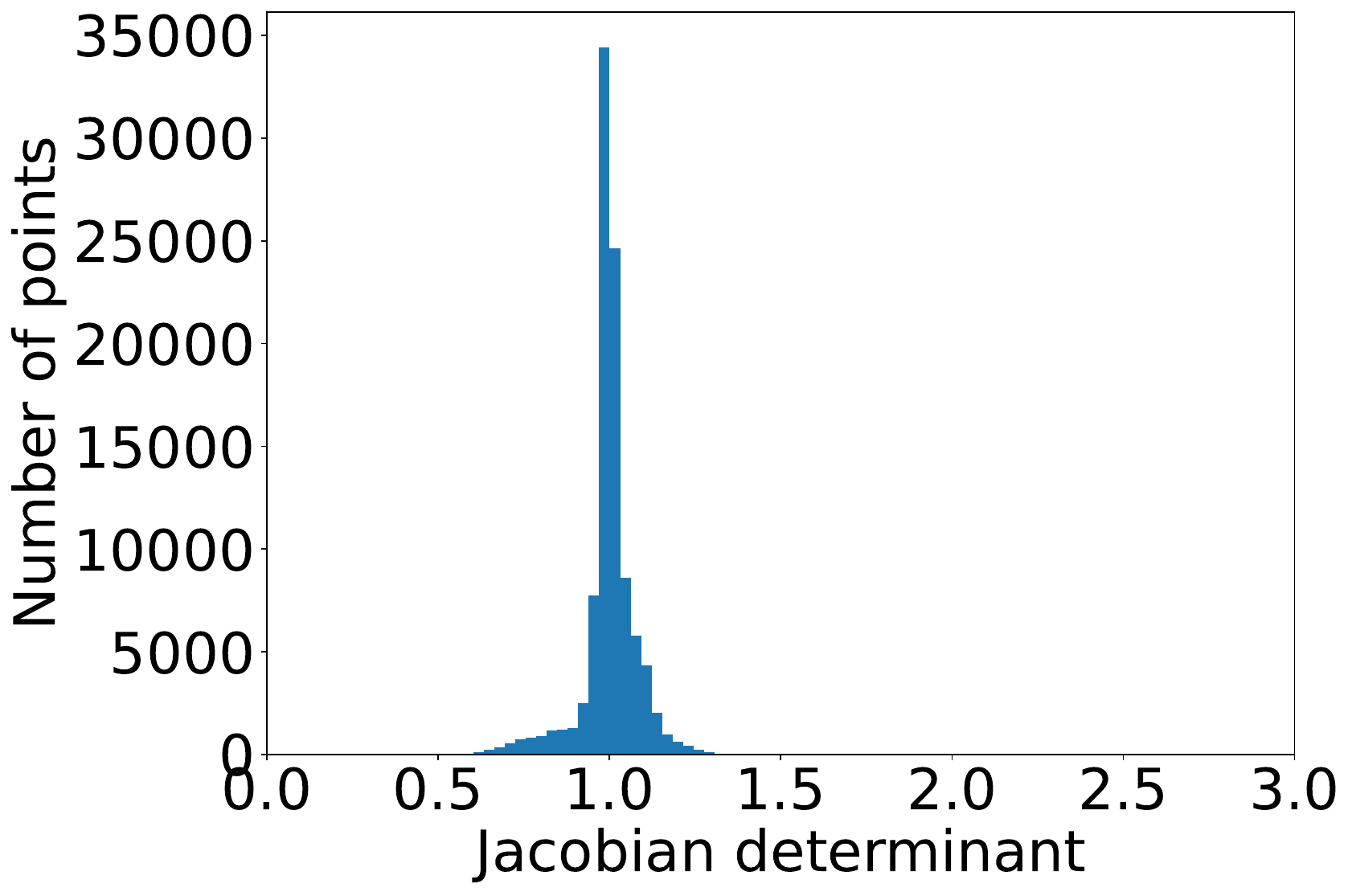}}\hspace{.1em}
    \subfloat[$\det{\nabla f_\theta}$:~Hybrid-matching]{\includegraphics[width=.32\textwidth]{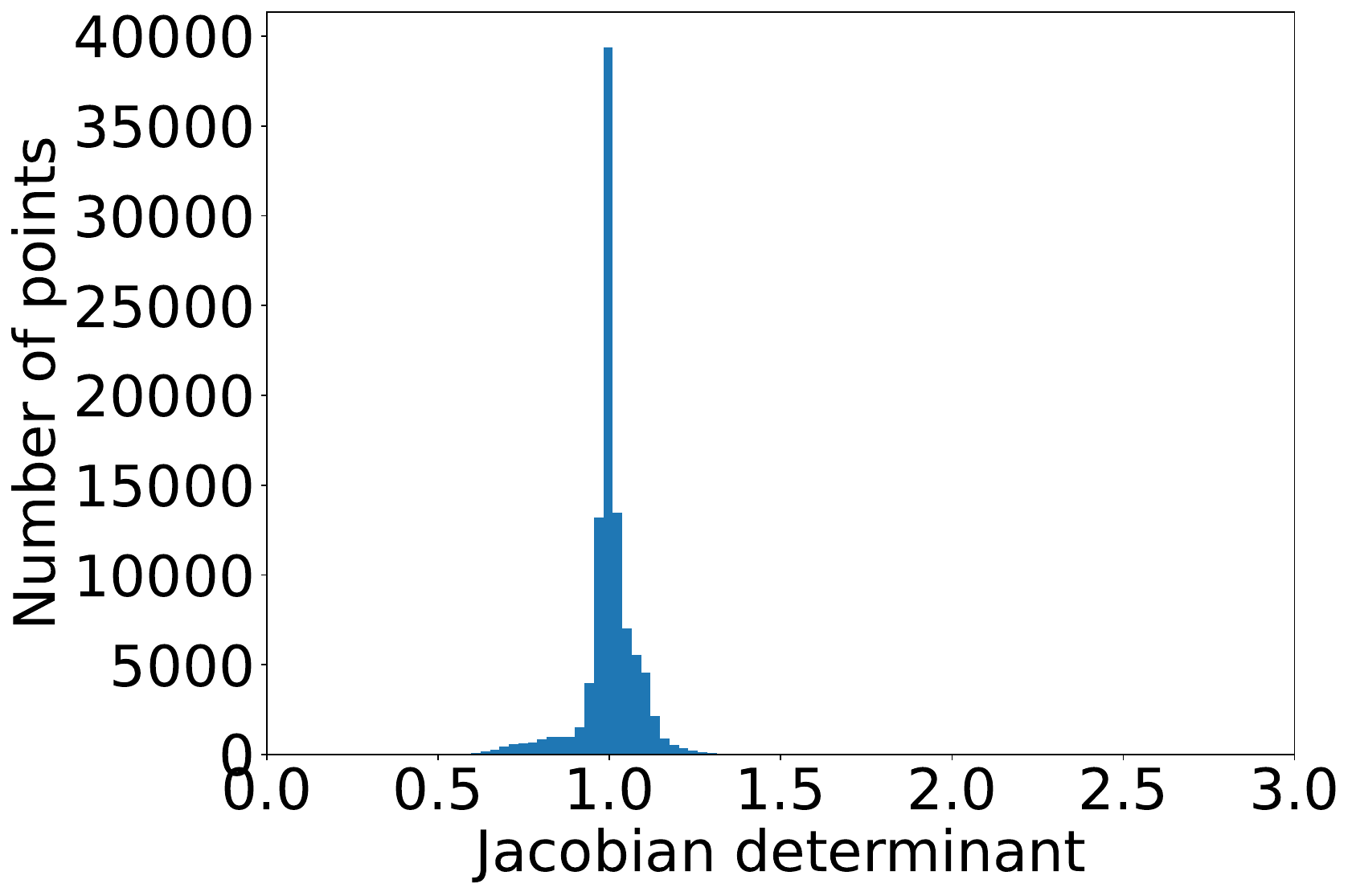}}\hspace{.1em}
    
    \subfloat[Conformality loss \label{fig:conformalLoss_4DCT}]{\includegraphics[width=.32\textwidth]{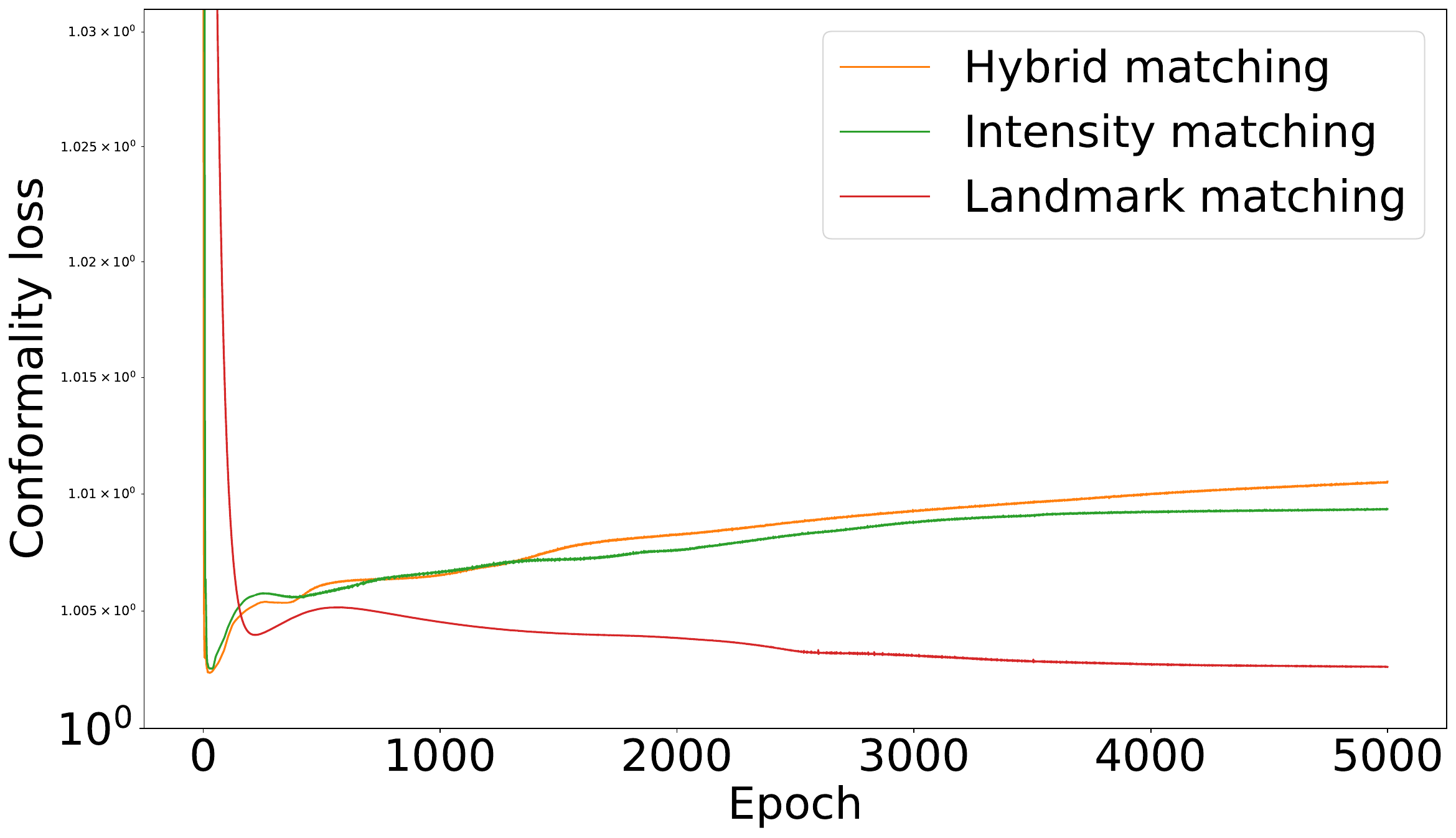}}\hspace{.1em}
    \subfloat[Intensity loss \label{fig:intensityLoss_4DCT}]{\includegraphics[width=.32\textwidth]{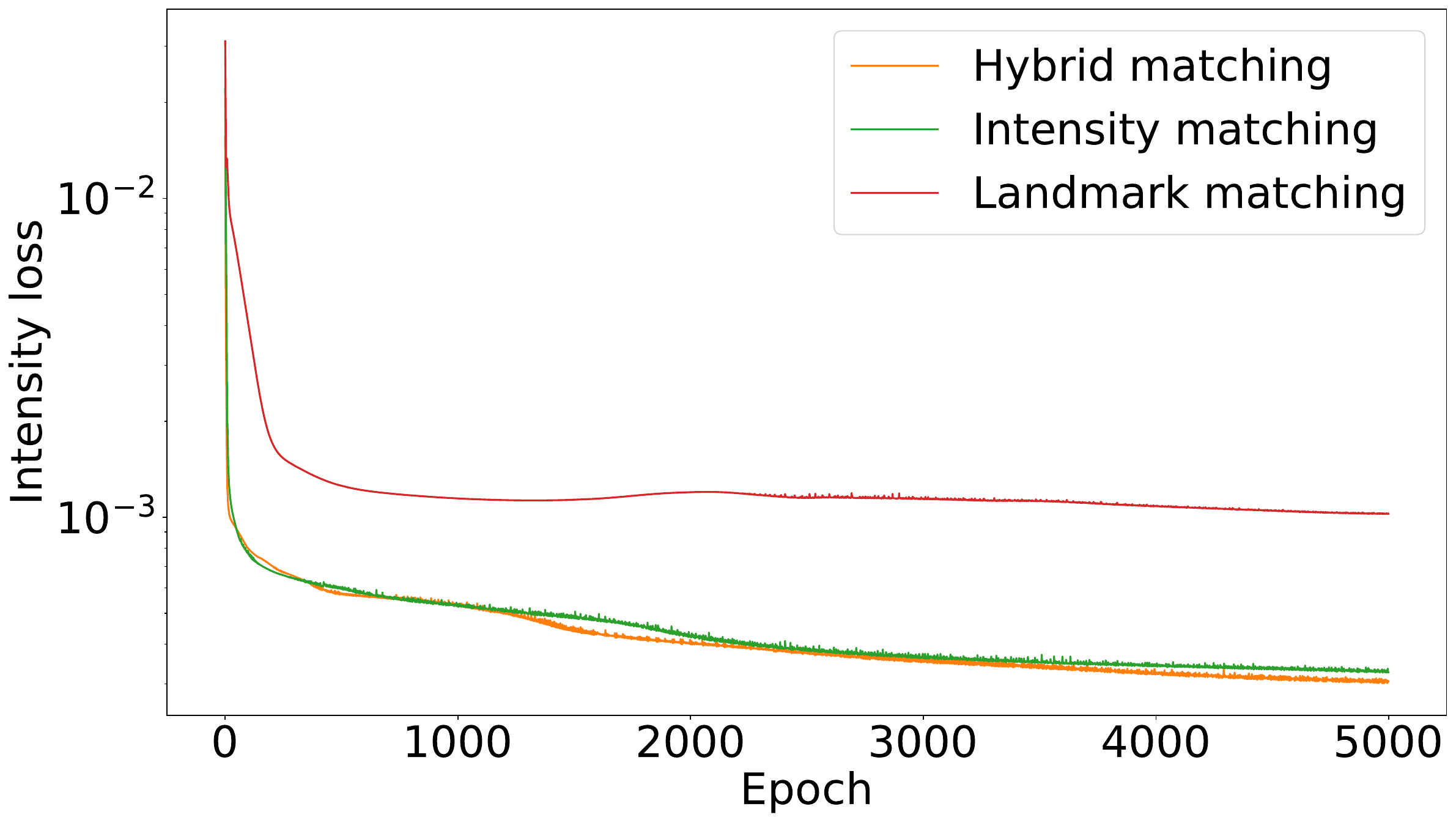}}\hspace{.1em}
    \subfloat[Landmark loss \label{fig:landmarkLoss_4DCT}]{\includegraphics[width=.32\textwidth]{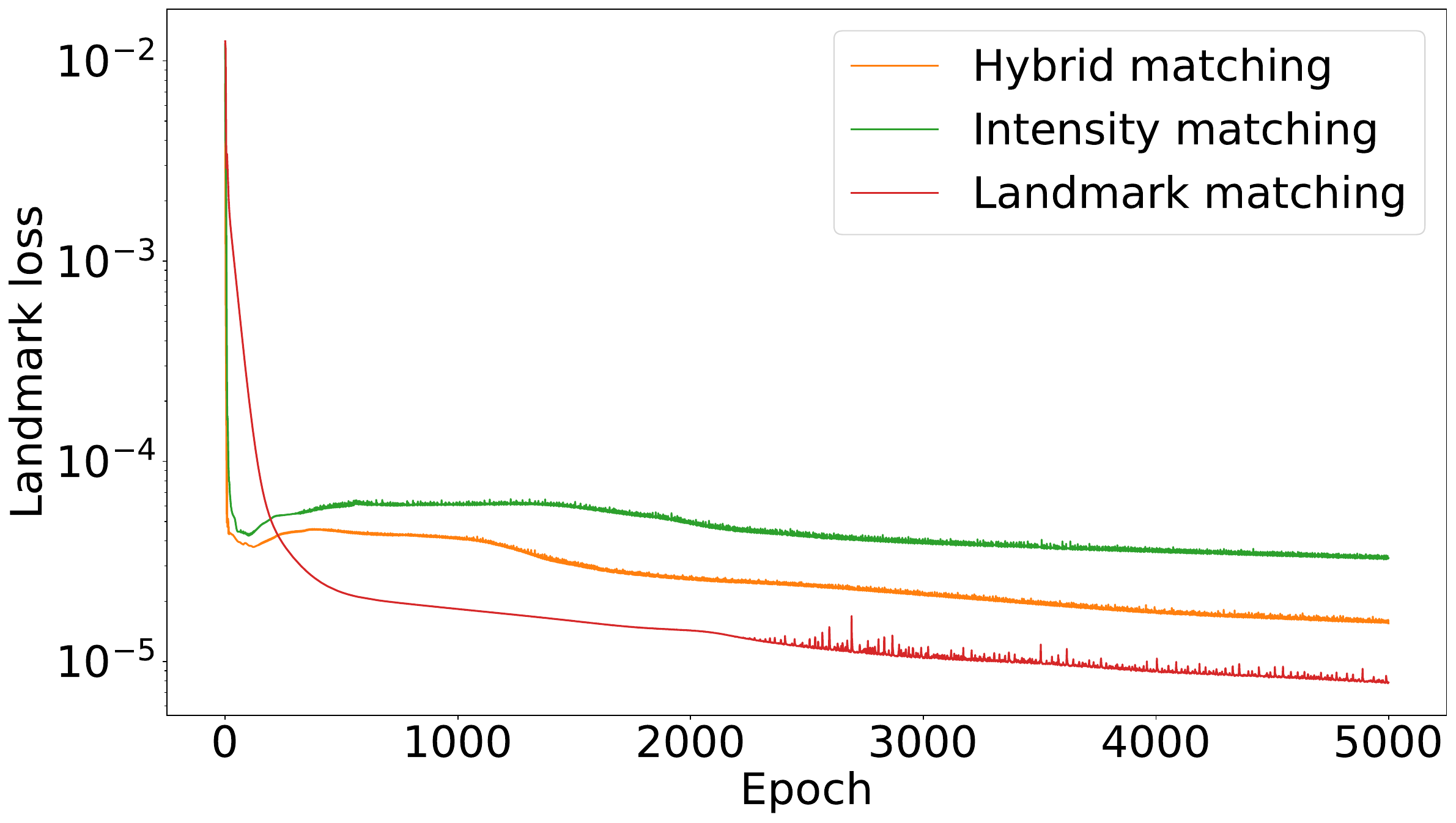}}\hspace{.1em}
    
    \caption{Results of the 4DCT Lung CT example. (a)-(c) visualize the mappings $f_\theta$ obtained from our proposed framework based on the three formulations in Section~\ref{sec:three_formulations}. (d)-(f) show the histograms of $\det \nabla f_\theta$ derived from the three formulations. (g)-(i) display the conformality loss, intensity loss, and landmark loss during the training of the three formulations, respectively.} 
    
    \label{fig:4DCT_maps}
\end{figure}

Following \cite{ZhangDaoping2022AUFf}, we resized the images to $128^3$ and normalized the pixel values to the range $[0,1]$. Unless otherwise specified, we empirically selected $\alpha_1=0.01$, $\alpha_2=50$, $\alpha_3=1$, $\alpha_4=0$, $\alpha_5=500$, and $\alpha_6=500$ in \eqref{eqn:hybrid_formulation}. In Fig.~\ref{fig:4DCT_maps}, we present snapshots of the resulting 3D mappings, along with the histograms of $\det \nabla f_\theta$ and the training losses for various variational models. From Fig.~\ref{fig:4DCT_maps} and Table~\ref{tab:4DCT}, we make the following observations: 
\begin{enumerate}
    \item The three formulations achieve small landmark mismatch errors and exhibit stable convergence (cf. Fig.~\ref{fig:4DCT_maps}(g)-(i));  
    \item The conformality loss is similar across the three formulations, indicating comparable levels of minimal conformality distortion;  
    \item The hybrid matching formulation achieves intensity mismatch errors comparable to or even smaller than those of the intensity matching formulation, likely due to the consistency of the data.  
\end{enumerate}

\begin{figure}[!htbp]
    \centering
    
    \subfloat[Source image $S$]{\includegraphics[width=.49\textwidth]{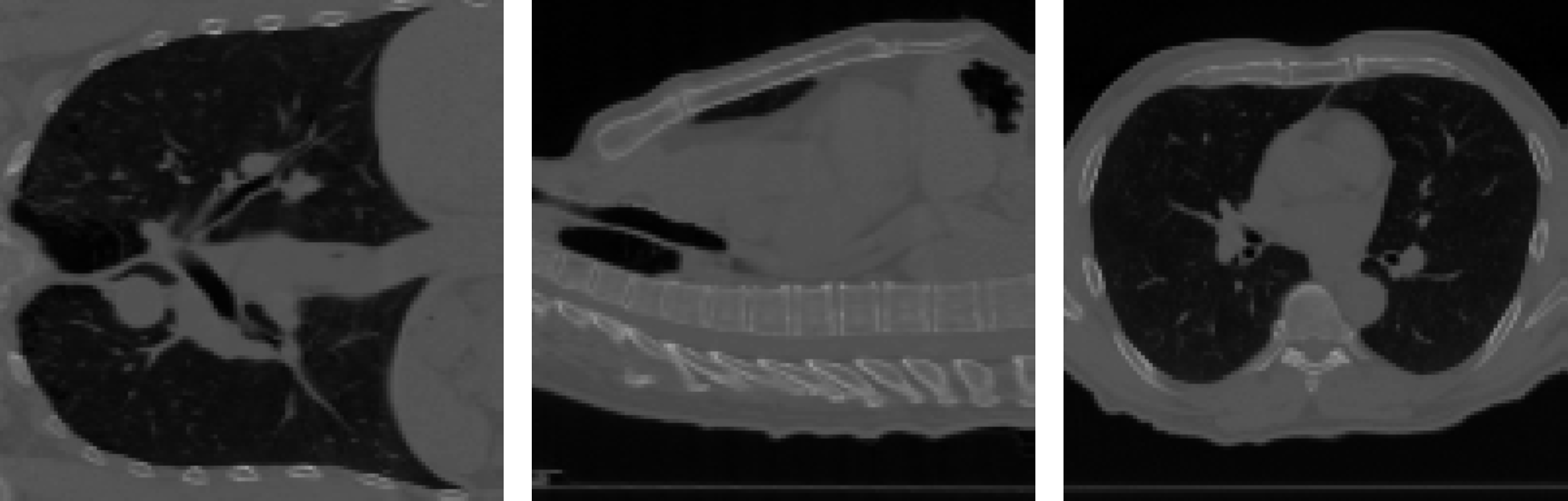}}\hspace{.1em}
    
    \subfloat[Target image $T$]{\includegraphics[width=.49\textwidth]{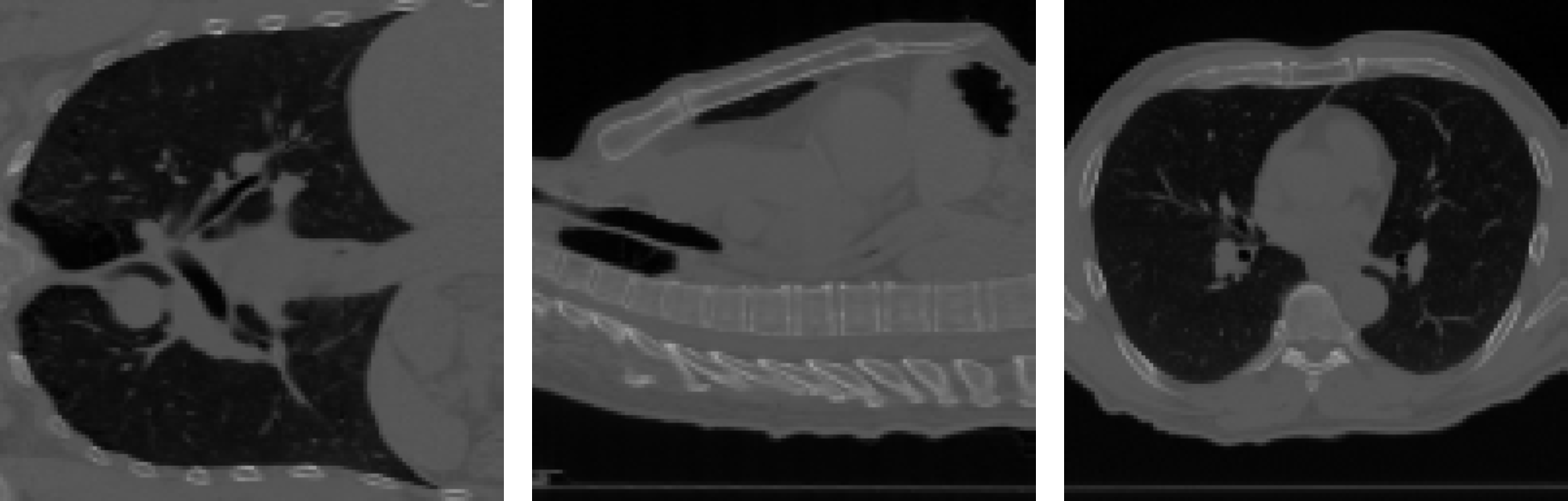}}\hspace{.1em}
    \subfloat[$|S-T|$]{\includegraphics[width=.49\textwidth]{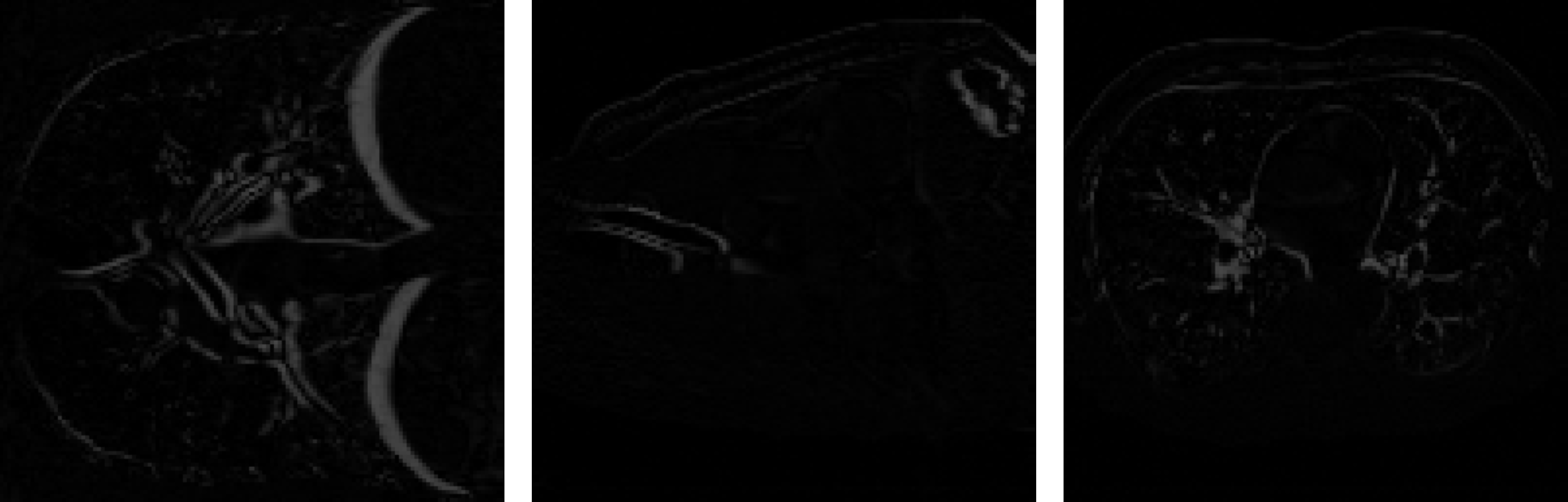}}\hspace{.1em}
    
    \subfloat[$S\circ f_\theta$: Landmark-matching ]{\includegraphics[width=.49\textwidth]{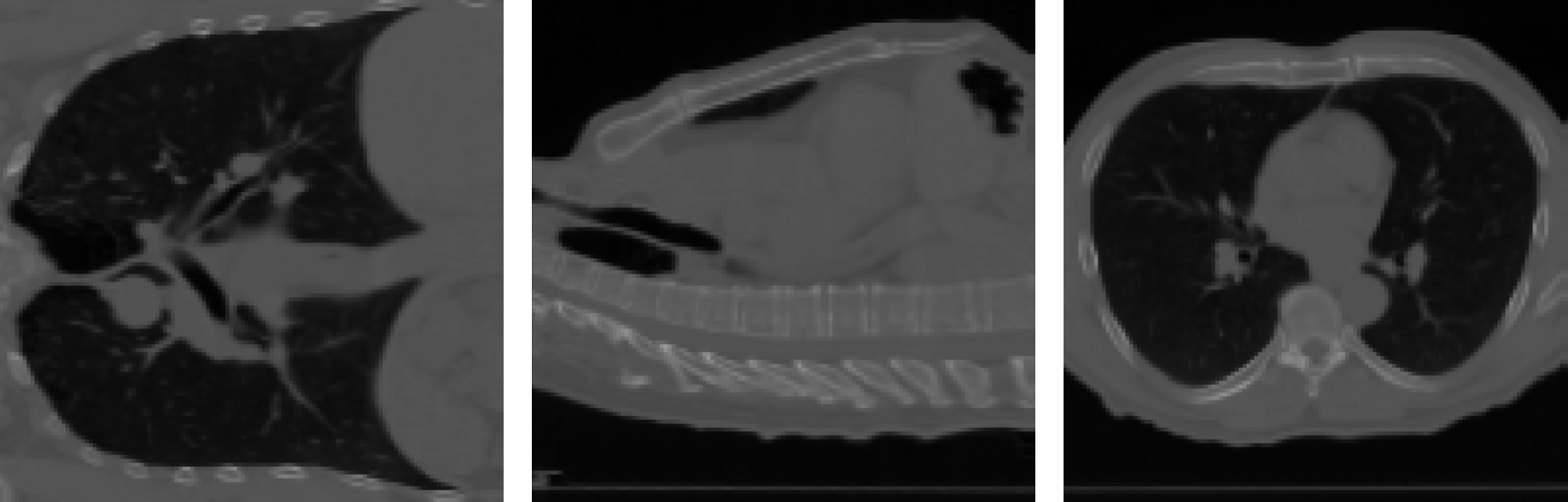}}\hspace{.1em}
    \subfloat[$|S\circ f_\theta - T|$ : Landmark-matching]{\includegraphics[width=.49\textwidth]{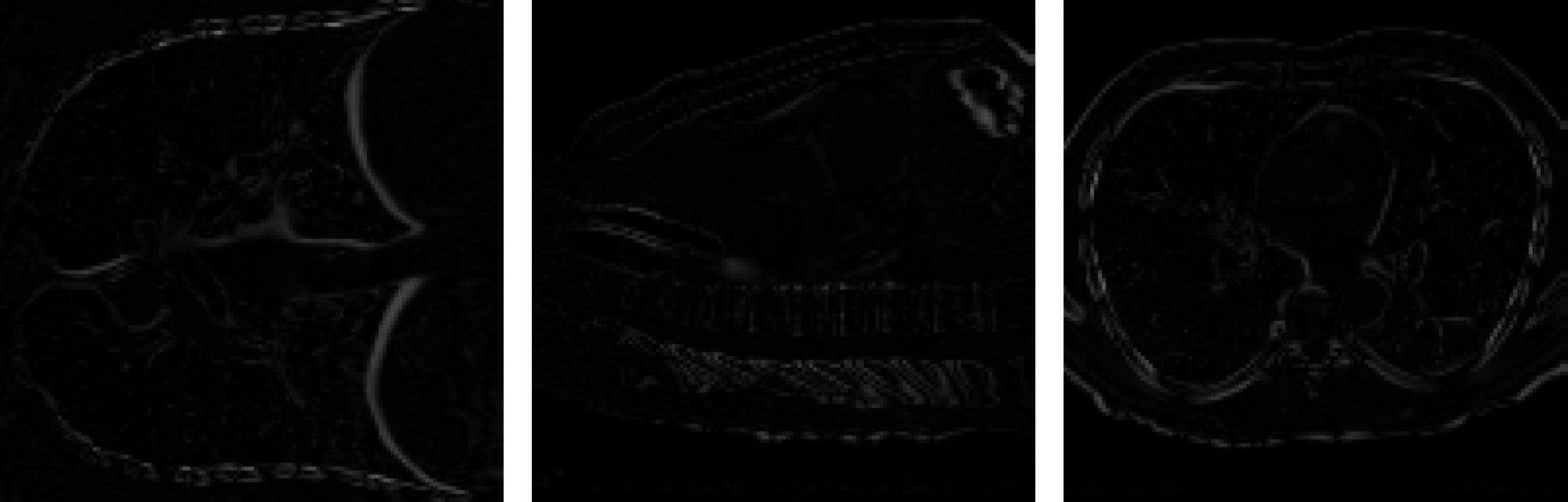}}\hspace{.1em}

    \subfloat[$S\circ f_\theta$: Intensity-matching ]{\includegraphics[width=.49\textwidth]{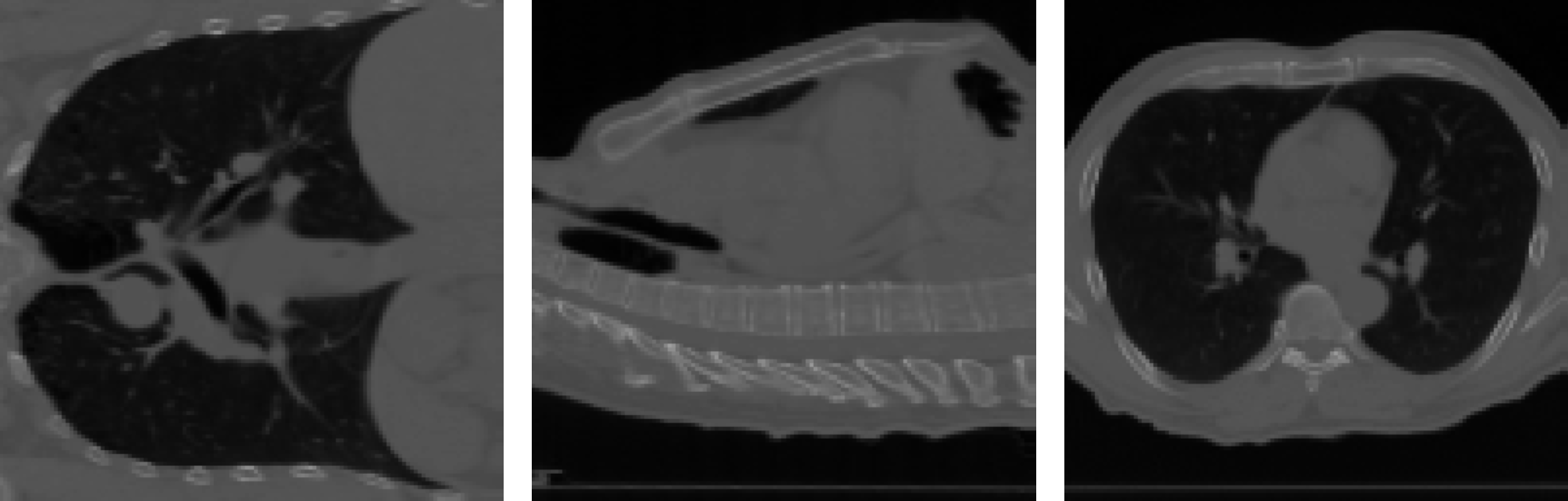}}\hspace{.1em}
    \subfloat[$|S\circ f_\theta - T|$: Intensity-matching]{\includegraphics[width=.49\textwidth]{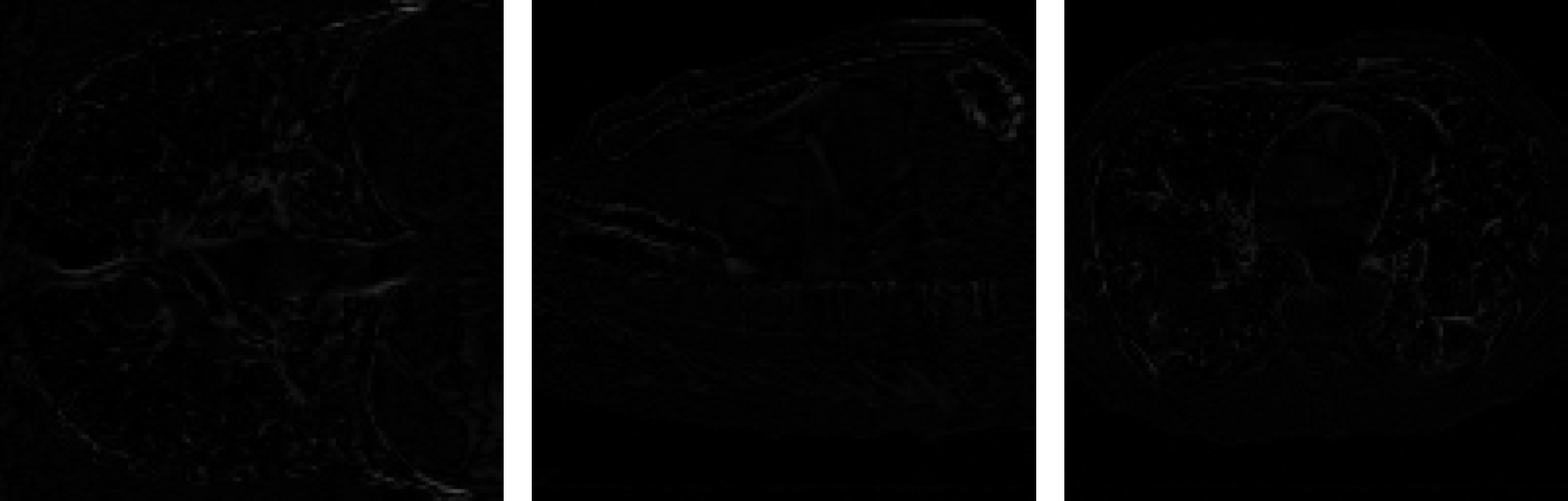}}\hspace{.1em}
    
    \subfloat[$S\circ f_\theta$: Hybrid-matching]{\includegraphics[width=.49\textwidth]{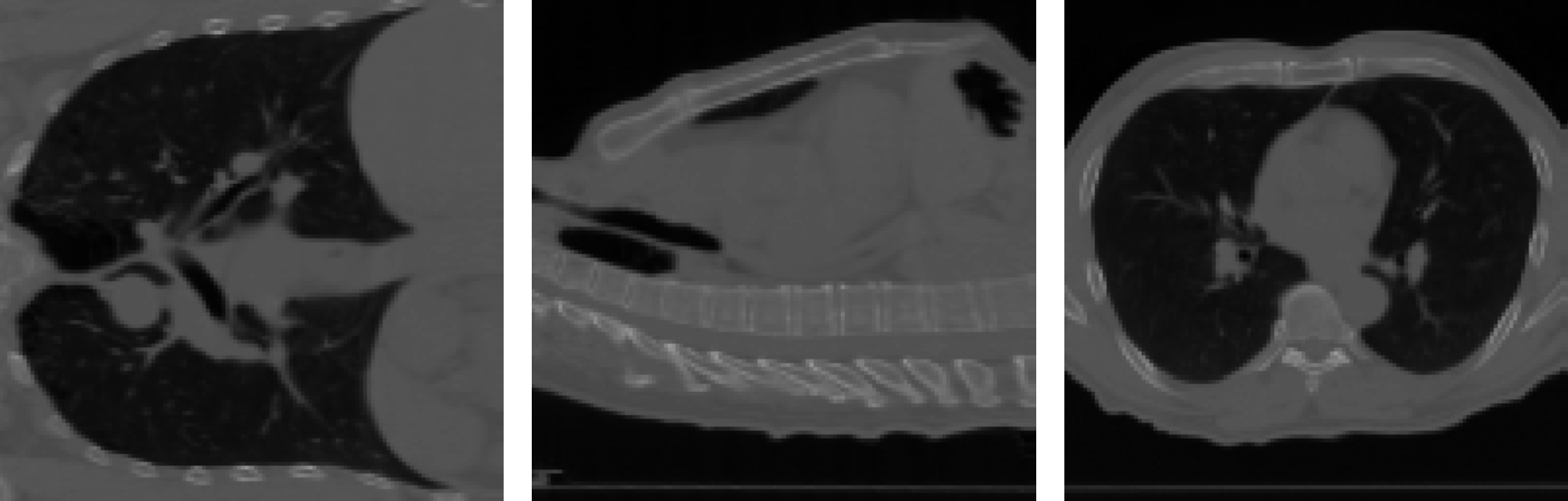}}\hspace{.1em}
    \subfloat[$|S\circ f_\theta - T|$: Hybrid-matching]{\includegraphics[width=.49\textwidth]{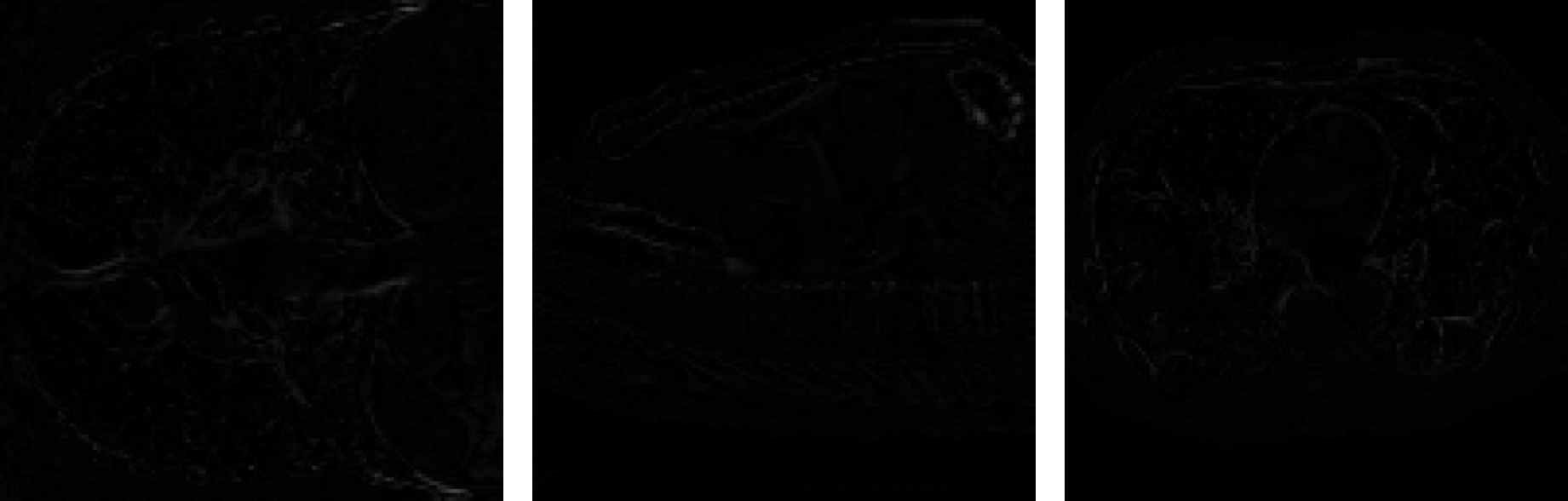}}\hspace{.1em}
    
    \caption{The 4DCT Lung CT example. Visualization of the registration results via three slice views, i.e., $x=0.5,\, y=0.5,\, z=0.5$. (a)-(b) Three views (slices) of the source image $S$ and target image $T$, respectively. (c) Three views of the absolute difference between $S$ and $T$. (d)-(e) The landmark matching registration results. (f)-(g) The intensity matching registration results. (h)-(i) The hybrid matching registration results.} 
    \label{fig:4DCT_slices}
\end{figure}

Additionally, Fig.~\ref{fig:4DCT_slices} displays various slice views of the source image $S$, target image $T$, warped image $S \circ f_\theta$, and their respective absolute differences. These results highlight the effectiveness of incorporating intensity information into our framework.

\begin{table}[!htbp]
    \centering
    \begin{tabular}{lccc}
        \hline
         & Landmark-matching & Intensity-matching & Hybrid-matching \\
         \hline
         Landmark loss & 6.1641e-6 & 2.0929e-5 & 1.1868e-5 \\ 
         Intensity loss & 1.0186e-3 & 2.8744e-4 & 2.6679e-4 \\ 
         Conformality loss & 1.0026e0 & 1.0098e0 & 1.0112e0 \\ 
         Smoothness loss & 1.4949e-1 & 1.5387e0 & 1.6597e0 \\
         \hline
    \end{tabular}
    \caption{The landmark loss, intensity loss, conformality loss and smoothness loss of three formulations for the 4DCT lung CT example. Losses are extracted from the last epoch of training.}
    \label{tab:4DCT}
\end{table}

\subsection{Ablation Study on Boundary Conditions: Soft v.s. Hard Constraint}
\label{sec:ablation_study_boundary}
In this section, we perform ablation study to compare the hard constraint approach and soft constraint approach for the boundary condition. This aims at highlighting the importance of our explicitly built in boundary constraint \eqref{eqn:hardboundary} for registration tasks.

We conducted a landmark matching experiment ($\alpha_6=0$) to compare hard constraint and soft constraint approaches using the Twisted Landmark Pairs example (cf. Section~\ref{sec:lm_8}). Three models were trained for comparison:  
\begin{enumerate}
    \item Soft constraint \eqref{eqn:hybird_formulation_softbdy_expectation} with $\alpha_7=50$,  
    \item Soft constraint \eqref{eqn:hybird_formulation_softbdy_expectation} with $\alpha_7=500$, and  
    \item Hard constraint \eqref{eqn:hybird_formulation_expectation}.  
\end{enumerate}

\begin{figure}[!htbp]
    \centering
    
    \subfloat[$f_\theta$: Soft. $\alpha_7=50$.]{\includegraphics[width=.32\textwidth]{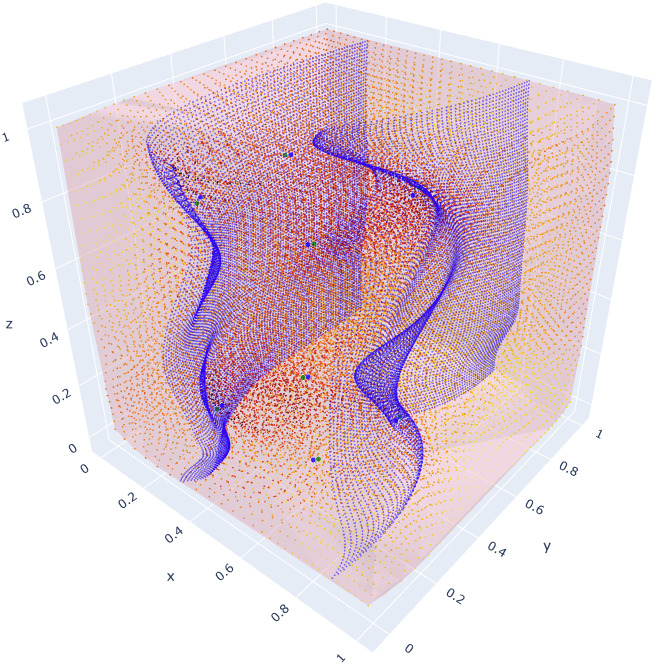}}\hspace{.1em}
    \subfloat[$f_\theta$: Soft. $\alpha_7=500$.]{\includegraphics[width=.32\textwidth]{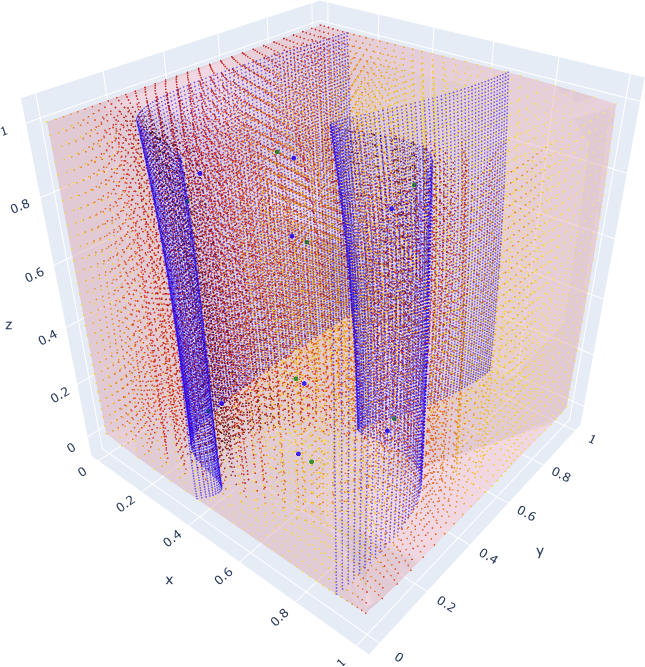}}\hspace{.1em}
    \subfloat[$f_\theta$: Hard.]{\includegraphics[width=.32\textwidth]{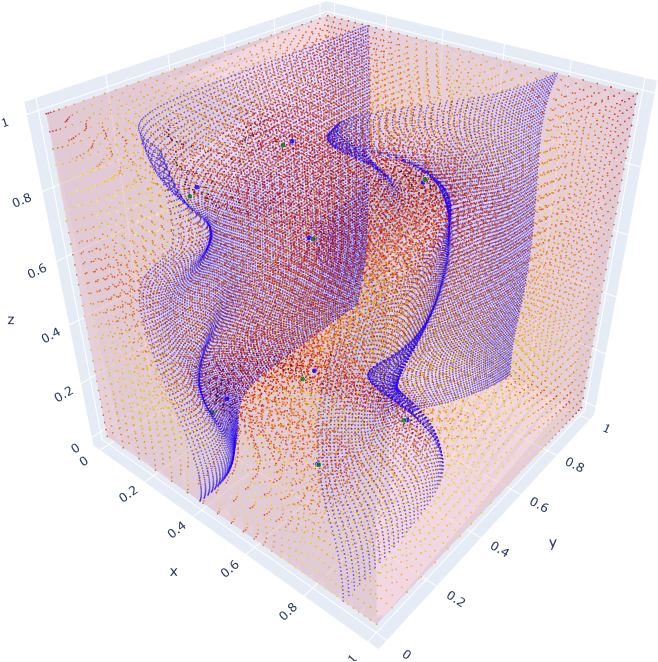}}\hspace{.1em}
    
    \subfloat[$f_\theta$: Soft. $\alpha_7=50$.]{\includegraphics[width=.32\textwidth]{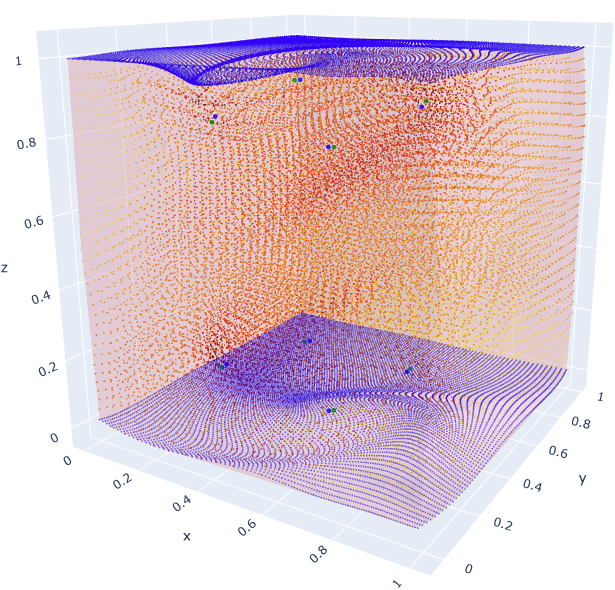}}\hspace{.1em}
    \subfloat[$f_\theta$: Soft. $\alpha_7=500$.]{\includegraphics[width=.32\textwidth]{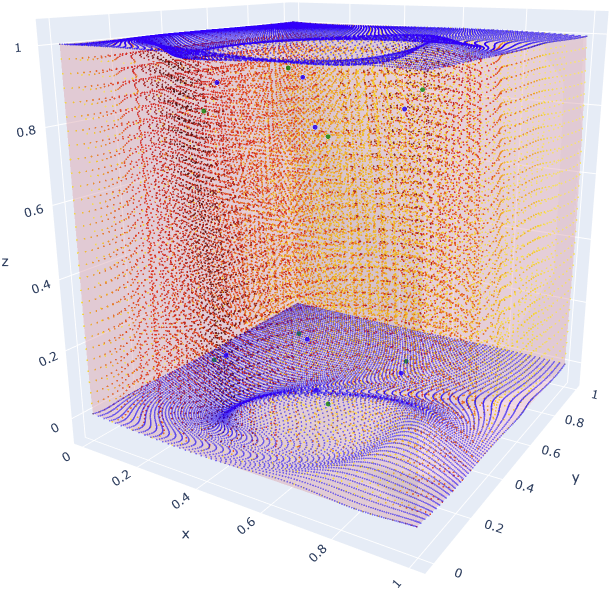}}\hspace{.1em}
    \subfloat[$f_\theta$: Hard.]{\includegraphics[width=.32\textwidth]{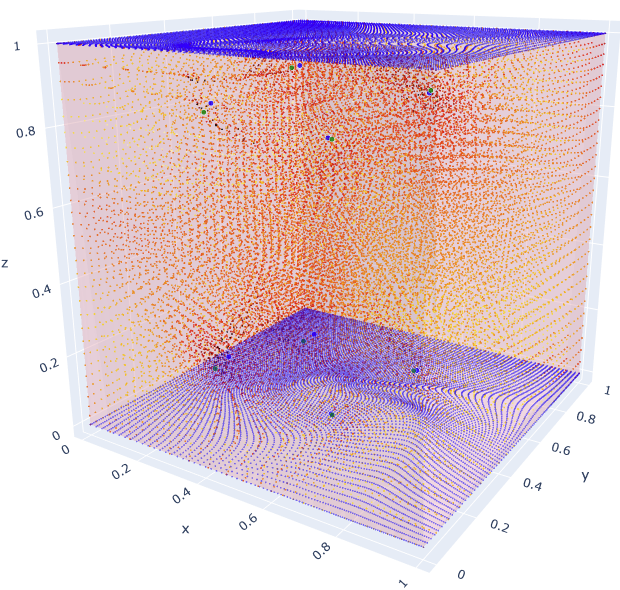}}\hspace{.1em}
    
    \subfloat[$\det\nabla f_\theta$: Soft. $\alpha_7=50$.]{\includegraphics[width=.32\textwidth]{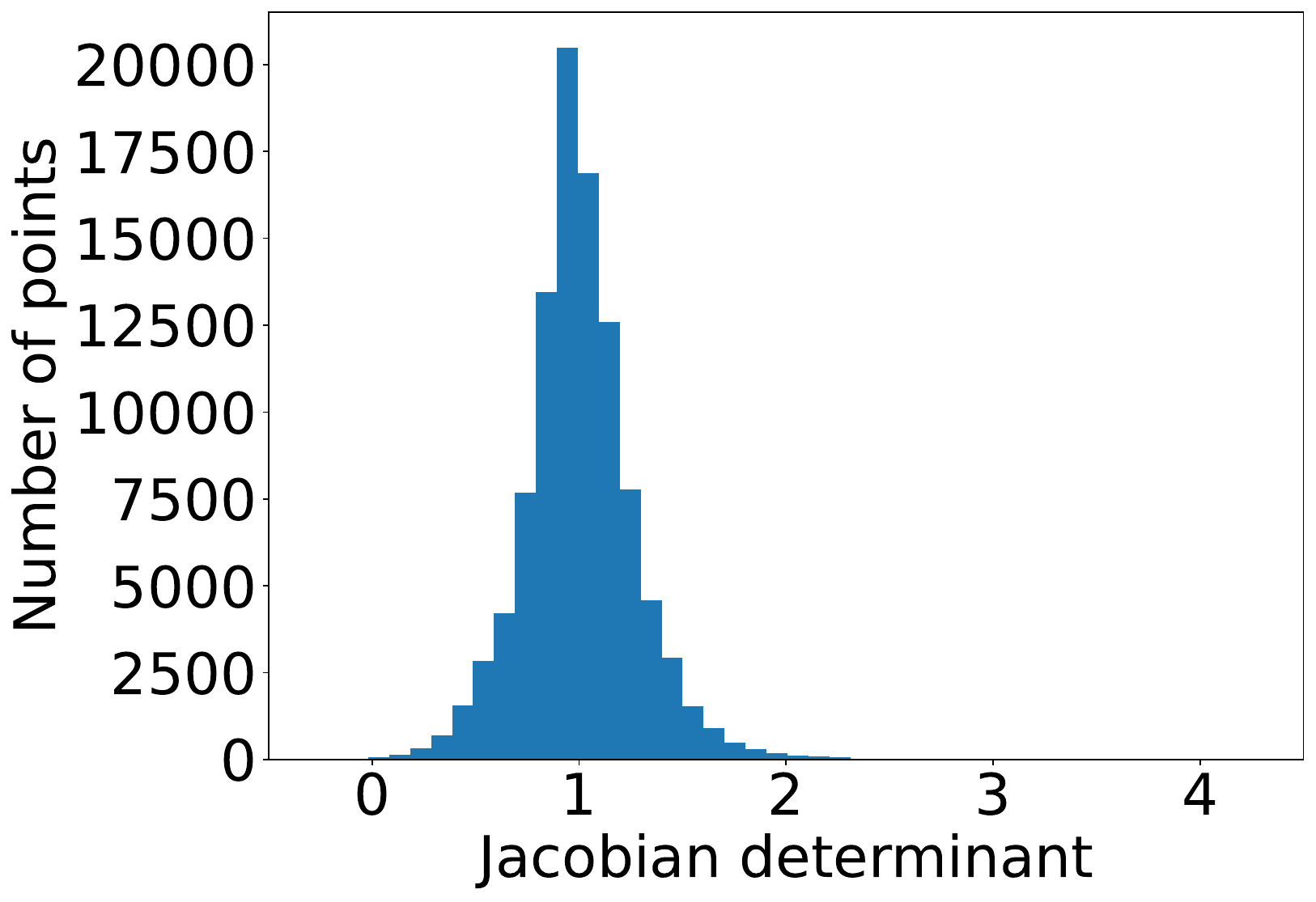}}\hspace{.1em}
    \subfloat[$\det\nabla f_\theta$: Soft. $\alpha_7=500$.]{\includegraphics[width=.32\textwidth]{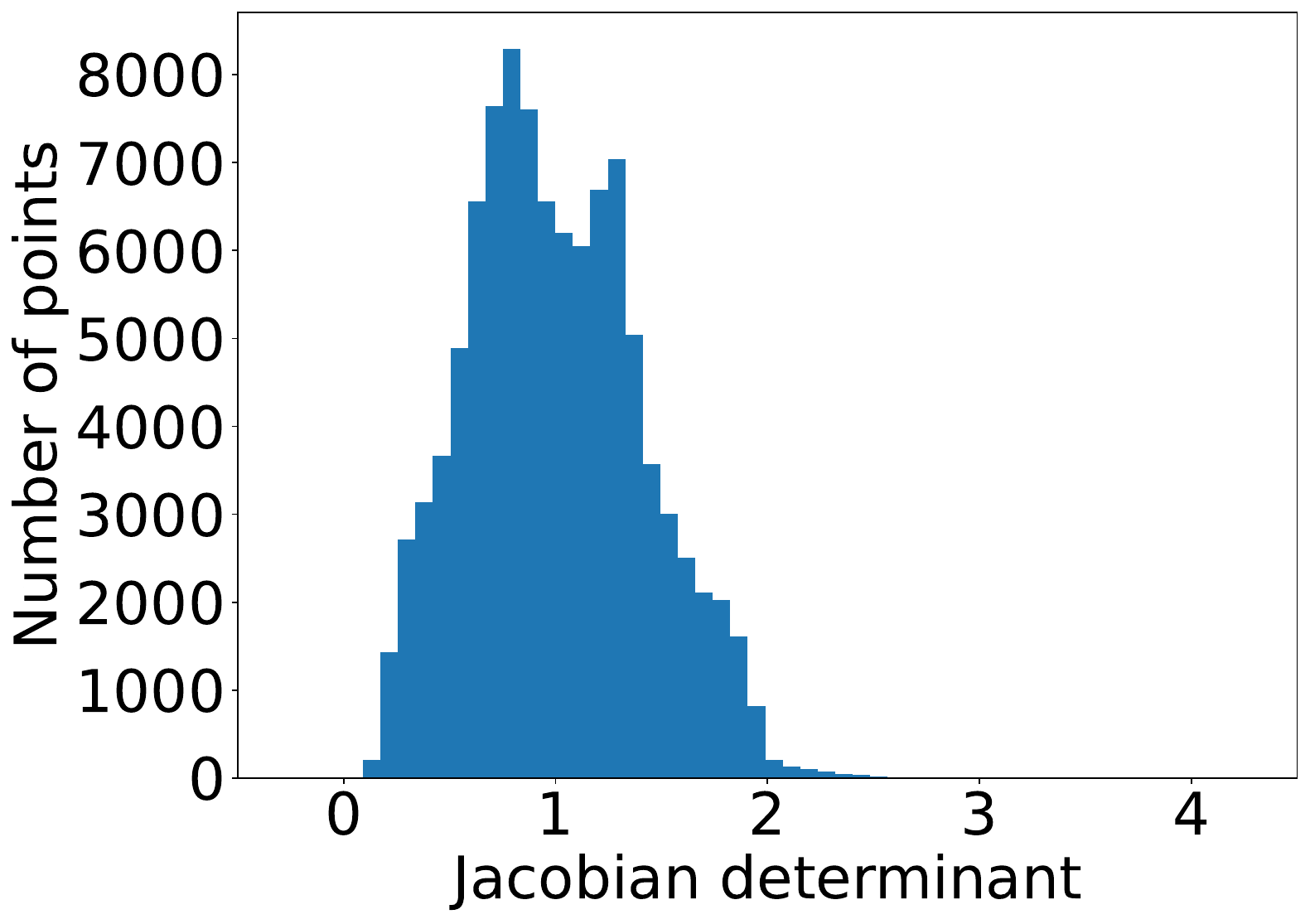}}\hspace{.1em}
    \subfloat[$\det\nabla f_\theta$: Hard.]{\includegraphics[width=.32\textwidth]{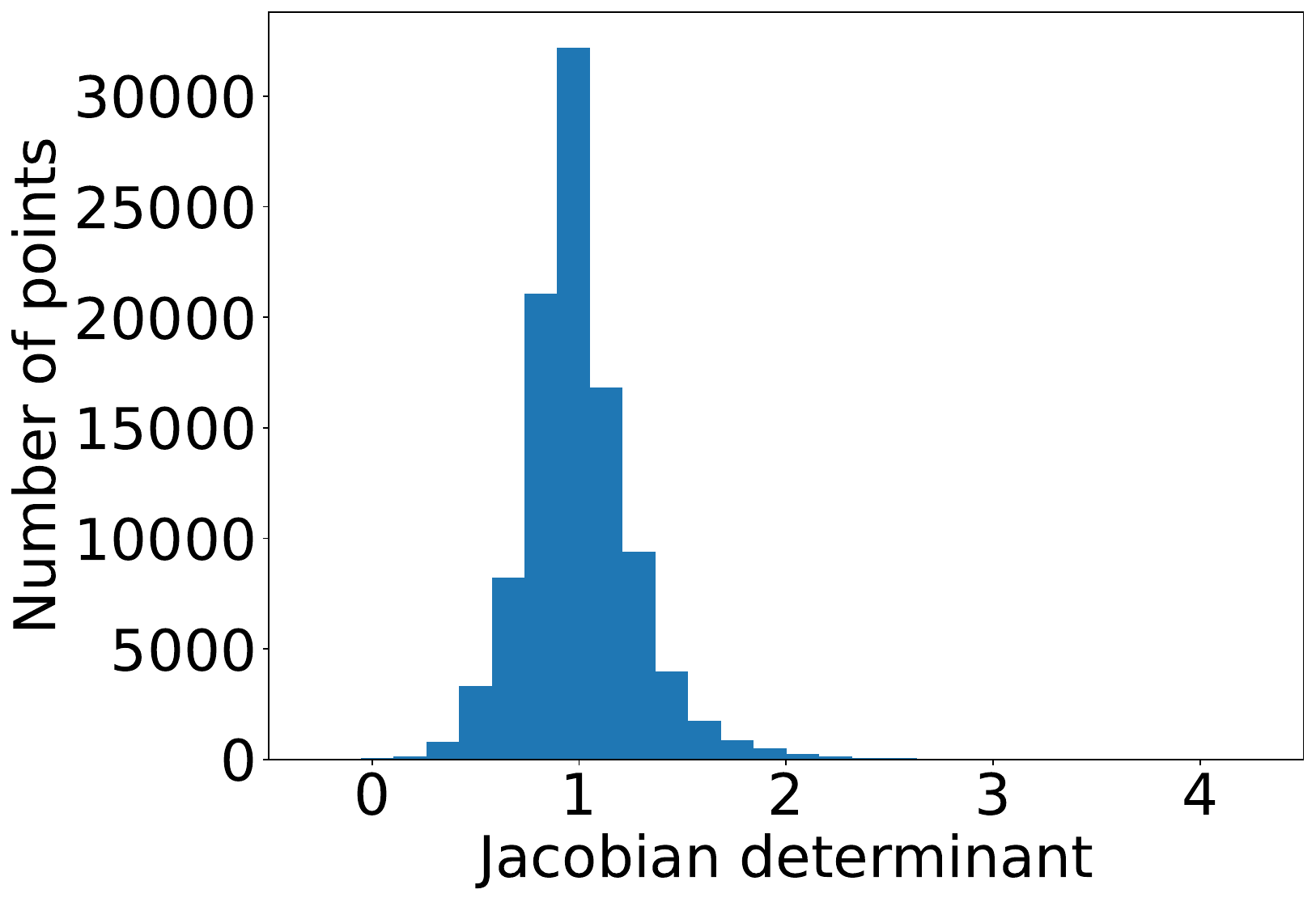}}\hspace{.1em}
    
    \subfloat[Boundary loss.]{\includegraphics[width=.32\textwidth]{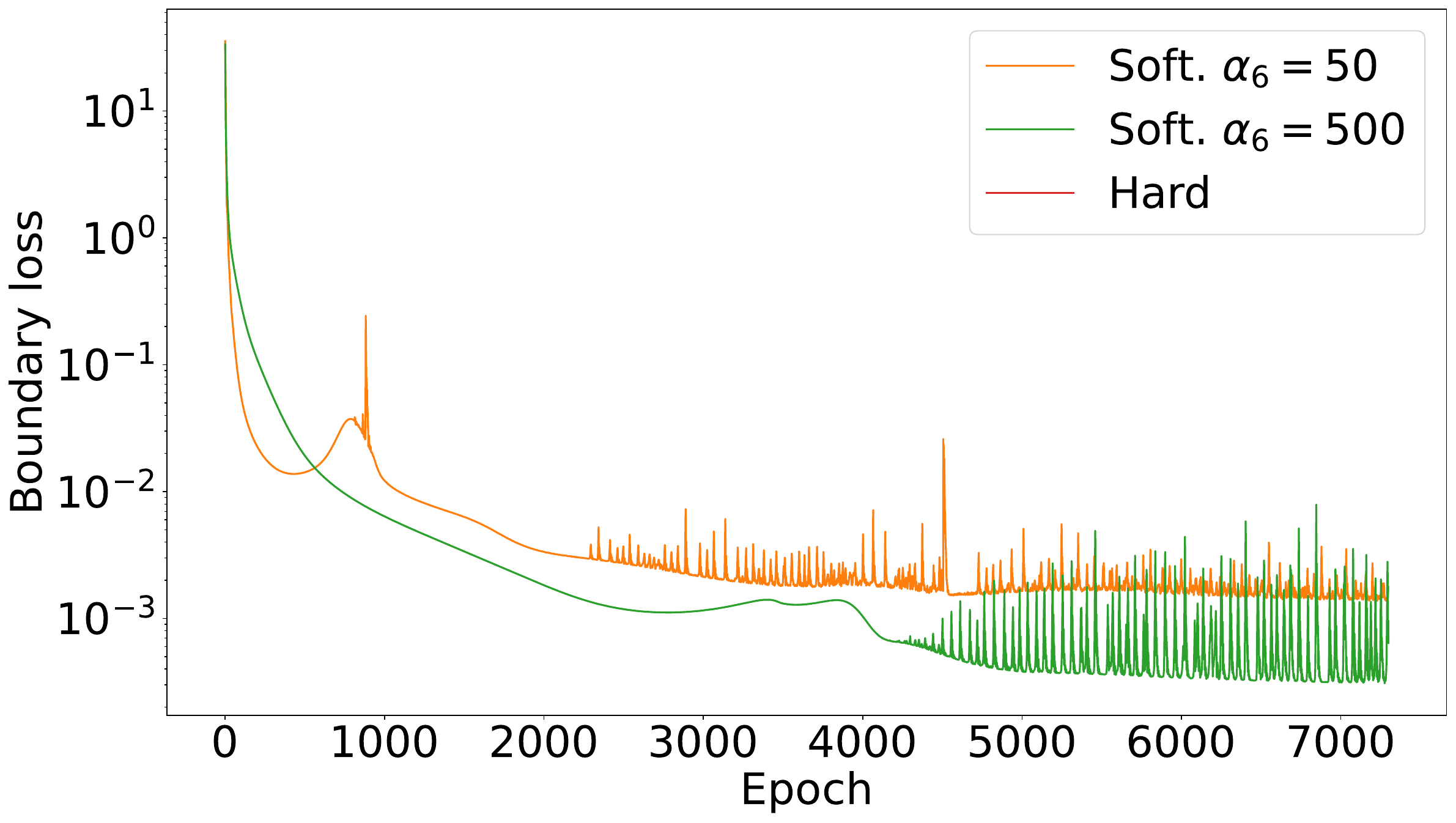}}\hspace{.1em}
    \subfloat[Landmark loss.]{\includegraphics[width=.32\textwidth]{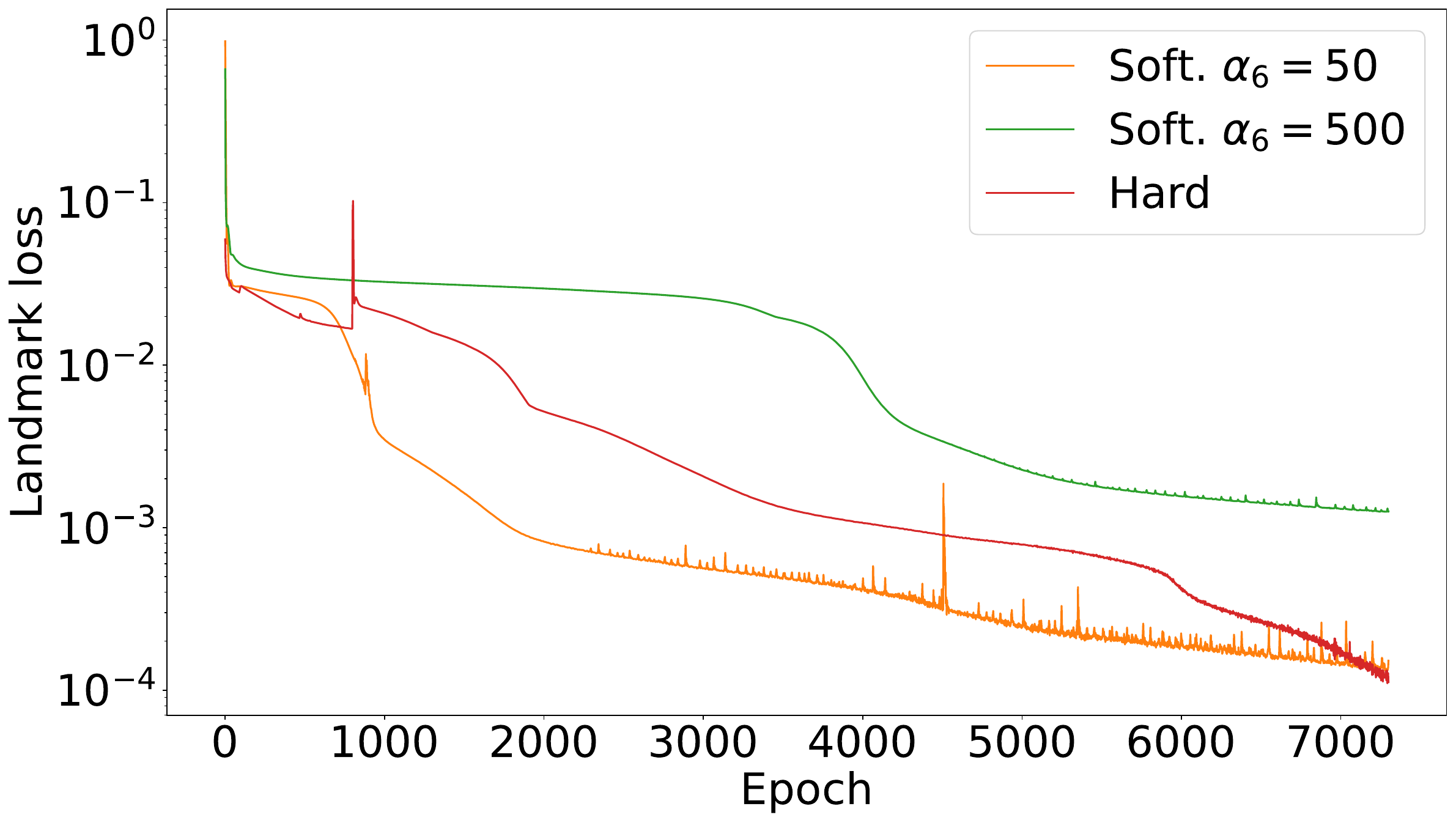}}\hspace{.1em}
    \subfloat[Conformality loss.]{\includegraphics[width=.32\textwidth]{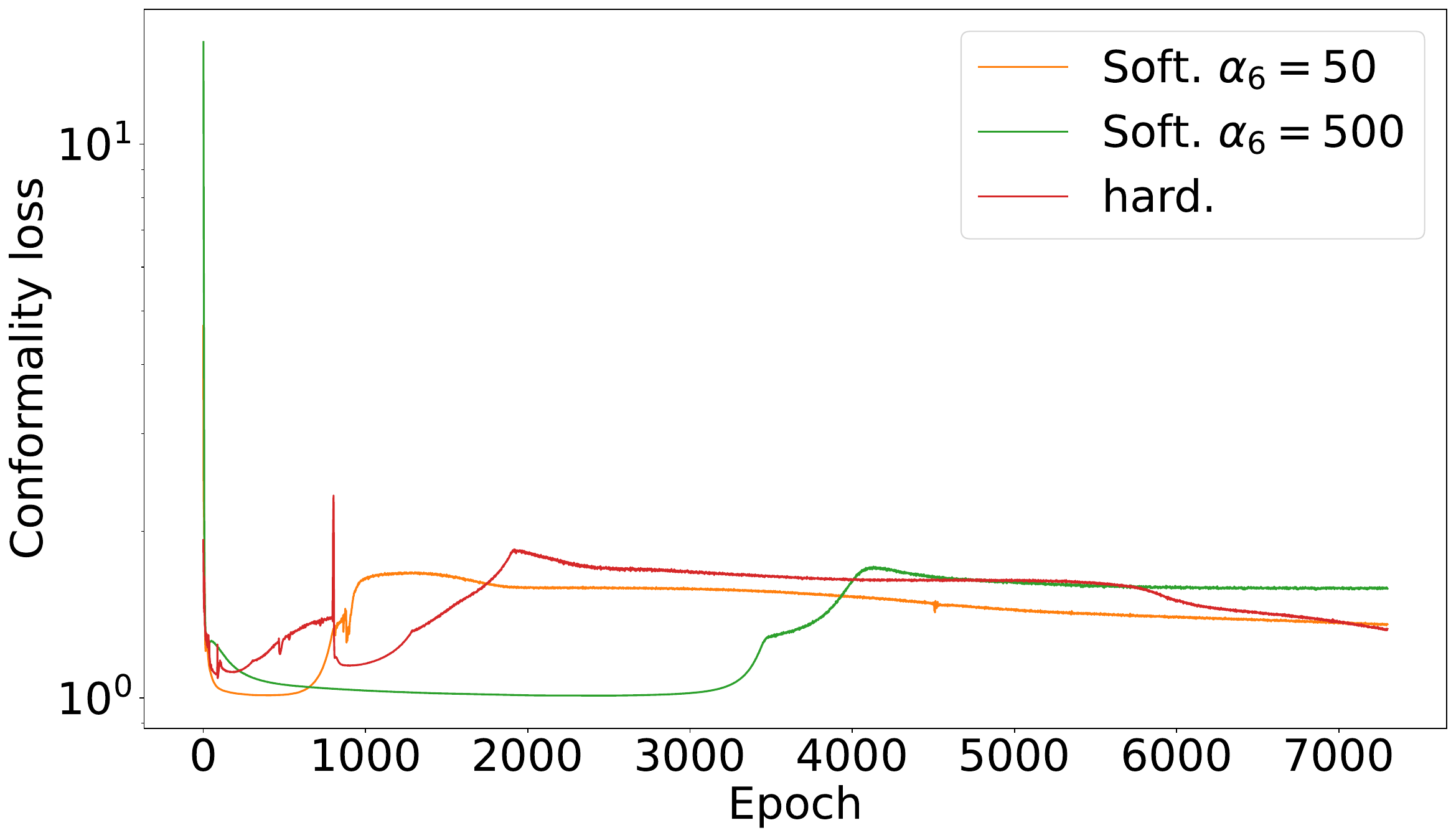}}\hspace{.1em}
    
    \caption{Results of the ablation study on boundary conditions with three settings: (i) Soft constraint with $\alpha_7=50$, (ii) Soft constraint with $\alpha_7=500$, and (iii) Hard constraint. (a)-(c) show the mappings $f_\theta$, with two cross-sectional views $f_\theta(x=0.2),\, f_\theta(x=0.8)$ colored in blue, under the three settings (i)-(iii) respectively. (d)-(f) show the mapping $f_\theta$ with two boundary planes $f_\theta(z=0),\,f_\theta(z=1)$ colored in blue under (i)-(iii) respectively. (g)-(i) are the histograms of $\det \nabla f_\theta$ under (i)-(iii) respectively. (j)-(l) are the conformality loss, intensity loss and landmark loss under (i)-(iii) during training respectively.} 
    \label{fig:ablation_lm8}
\end{figure}

From the results in Fig.~\ref{fig:ablation_lm8}, we made the following key observations:
\begin{enumerate}
    \item Boundary Condition Satisfaction:
    As shown in Fig.~\ref{fig:ablation_lm8}, the hard constraint approach ensures that the resulting mapping satisfies the boundary conditions \eqref{eqn:boundary_cond} with zero boundary error (cf. Fig.~\ref{fig:ablation_lm8}(f)). In contrast, the soft constraint approach incrementally increases $\alpha_7$ but still exhibits visually inaccurate boundaries (cf. Fig.~\ref{fig:ablation_lm8}(d)(e)).
    \item Impact of Smoothness and Conformality Terms:
    If prescribed landmarks are located near a boundary plane and are expected to undergo significant distortion, the boundary plane is prone to high boundary error due to the diffeomorphic term enforcing smooth properties. This limitation is evident in the soft constraint approach.
    \item Local Minima in Soft Constraint Optimization:
    The model trained with a soft boundary constraint ($\alpha_7 = 500$) is more likely to converge to a local minimum (cf. Fig.~\ref{fig:ablation_lm8}(a)(b)(c)) compared to $\alpha_7 = 50$ or the hard constraint approach. This suggests that the hard boundary constraint improves the parameter landscape by performing optimization on a further constrained subspace with exact boundary conditions.
    \item Superiority of Hard Constraint Approach:
    Despite the landmarks being situated close to the plane $f_\theta(z=0)$, the hard constraint model not only achieves zero boundary error but also exhibits the smallest landmark loss compared to the other two soft constraint models (cf. Fig.~\ref{fig:ablation_lm8}(k)). This result demonstrates the superiority of the hard constraint approach over the soft constraint approach in solving the diffeomorphism optimization problem.
\end{enumerate}
The above observations, directly support the use of hard constraint in our setting. 

\section{Conclusion}
In this work, we introduced a mesh-free learning framework for high-dimensional diffeomorphic mapping problems, with a particular focus on $n$-dimensional registration tasks. By bridging variational formulations and modern machine learning techniques, our method addresses the curse of dimensionality that commonly arises from domain discretization. Central to our approach is the smooth parameterization of the transformation via neural networks, which not only guarantees differentiability and flexibility but also naturally incorporates Dirichlet boundary conditions—crucial in many imaging applications.

Through a principled integration of Quasi-conformal theory, volume prior and a novel bijectivity loss, we demonstrated that our method scales effectively to high-dimensional problems and performs reliably across synthetic and real-world datasets. Empirical results confirm that our framework enables stable optimization and accurate diffeomorphic mappings without the need for mesh-based discretization or specialized solvers.

Overall, our findings highlight the potential of using smooth neural parameterizations to reformulate classical PDE-constrained mapping problems in a scalable and efficient manner. This opens new possibilities for solving a wide range of high-dimensional geometric registration and transformation problems in computational science and medical imaging.




\newpage

\appendix
\section*{Appendix}

\section{Construction of the Synthetic $3$D Mapping $g$ and Source Image $S$}
\label{appendix:synthetic_map}

The synthetic $3$D mapping $g(x,y,z)=(u(x,y,z), v(x,y,z), w(x,y,z))$ used in the large distortion example (cf. Section \ref{sec:low_freq}) is formulated as
%
\begin{align*}
    u(x,y,z) &= x + D^u(x)\, \tilde{u}(x,y,z) \\
    v(x,y,z) &= y + D^v(y)\, \tilde{v}(x,y,z) \\
    w(x,y,z) &= z + D^w(z)\, \tilde{w}(x,y,z)
\end{align*}
where $D^u(x)$, $D^v(y)$ and $D^w(z)$ are smooth functions that are nonzero within $(0,1)$ and vanish at $0$ and $1$, while $\tilde{u}(x,y,z)$, $\tilde{v}(x,y,z)$ and $\tilde{w}(x,y,z)$ are smooth functions that control the interior distortion of $g$.

\begin{figure}[!htbp] 
    \centering

    \begin{subfigure}[b]{0.45\textwidth}
        \includegraphics[width=\textwidth]{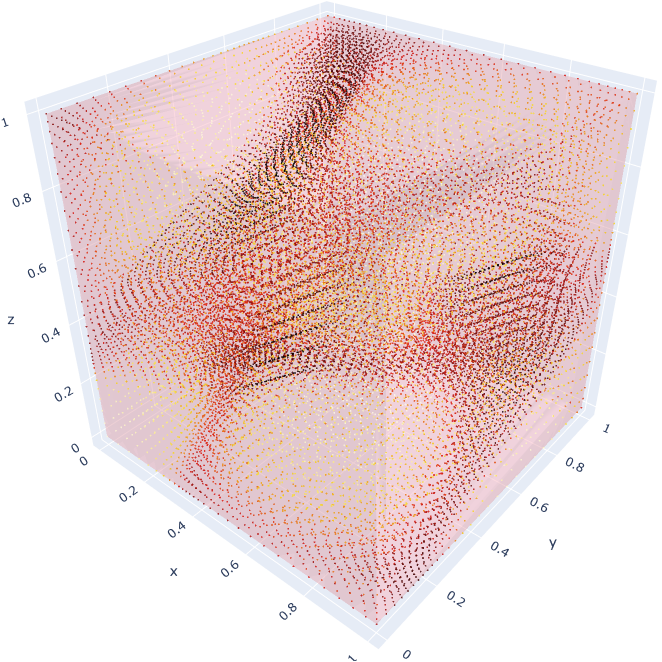}
        \caption{Mapping $g$ (large distortion)}
    \end{subfigure}\hfill
    \begin{subfigure}[b]{0.45\textwidth}
        \includegraphics[width=\textwidth]{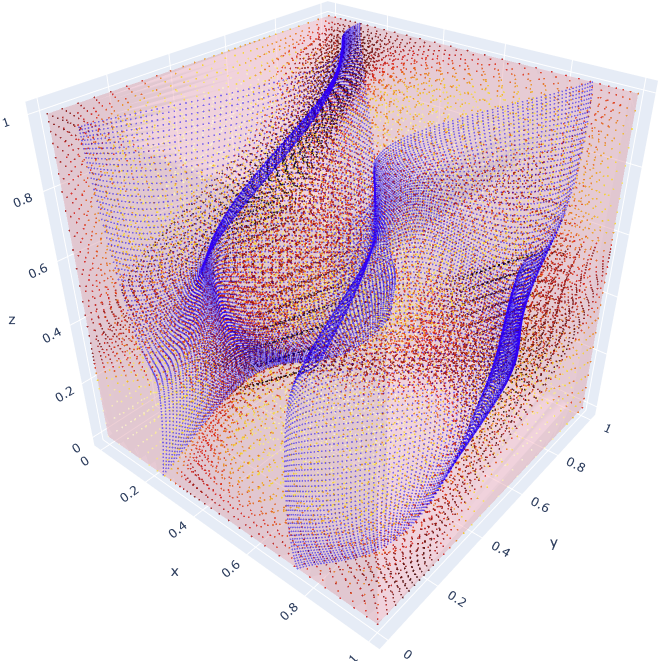}
        \caption{Mapping $g$ with cross-sectional views}
    \end{subfigure}

    \vspace{1em}

    \begin{subfigure}[b]{0.45\textwidth}
        \includegraphics[width=\textwidth]{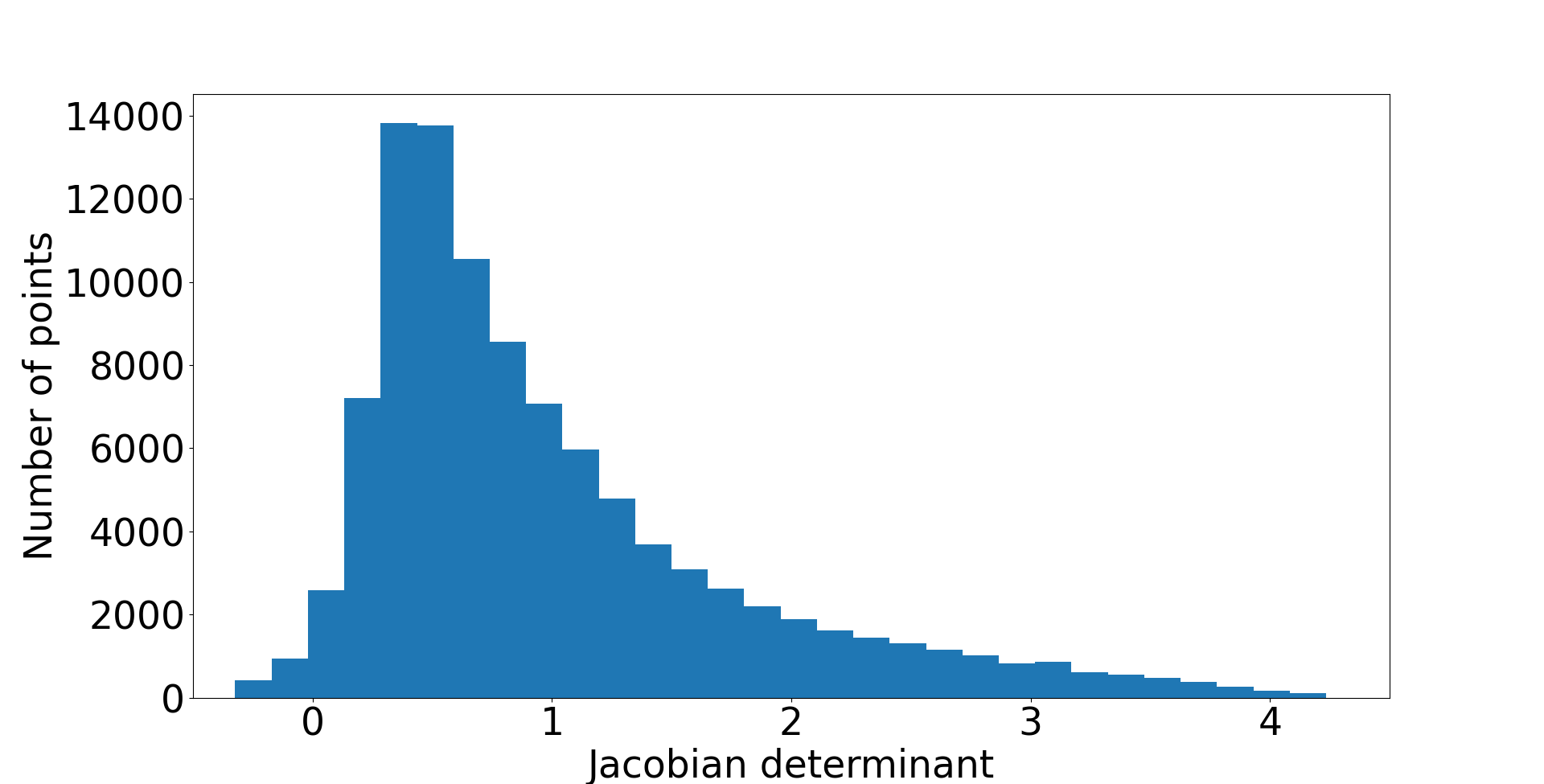}
        \caption{Histogram of $\det\nabla g$}
    \end{subfigure}

    \caption{Synthetic 3D mapping $g$ and corresponding histogram of $\det\nabla g$ for the large distortion example.}
    \label{fig:synthetic_maps}
\end{figure}

\textbf{Large Distortion Mapping} (cf. Section \ref{sec:low_freq}). We define $g:[0,1]^3\rightarrow [0,1]^3$ (cf. Fig. \ref{fig:synthetic_maps}) with
\begin{align*}
    D^u(x) = x(1-x) & \text{ and } \tilde{u}(x,y,z) = \cos(5x+6y-4z) / 2\\ 
    D^v(y) = y(1-y) & \text{ and } \tilde{v}(x,y,z) = \cos(-5x+4y+5z) / 2\\ 
    D^w(z) = z(1-z) & \text{ and } \tilde{w}(x,y,z) = \cos(3x+5y-6z) / 2
\end{align*}
and the source image $S:[0,1]^3\rightarrow \mathbb{R}$ is given by
\begin{equation} \label{eqn:src_img_low_freq}
    S(x,y,z) := \frac{1}{2} \cos(2 \cdot 2\pi \cdot (x^2 + y^2 + z^2)) + \frac{1}{2}.
\end{equation}
%

%

\section{Proof of Theorem 
\ref{thm:existence}}
\label{append:thm_existence}

\begin{proof}
    For notational simplicity we write $\mathcal{C}^p = \mathcal{C}^p(\Omega)$, where $p \in \mathbb{N} \cup \{\infty\}$. We first show that $\mathcal{A}$ is non-empty. By Homogeneity Lemma, there exists a $f^* \in \mathcal{C}^{\infty}$ satisfying the boundary constraint \cite{JoshiS.C.2000Lmvl}. By boundedness of $\Omega$, $c_1 := \| f \|_{\infty}$, $c_2 := \| \nabla f \|_{\infty}$, $c_3 := \| \nabla^2 f \|_{\infty}$ are all bounded. This shows that $\mathcal{A}$ is non-empty.

    To show that $\mathcal{A}$ is compact, it suffices to show that $\mathcal{A}$ is complete and totally bounded. For any $f \in \mathcal{A}$, define:
    \begin{equation}
        \| f \| := \|f\|_{\infty} + \| \nabla f\|_{\infty} + \|\nabla^2 f\|_{\infty}.
    \end{equation}
    We first show that $\mathcal{A}$ is complete with respect to $\| \cdot \|$. Let $\{ f^k \}^{\infty}_{k = 1}$ be a Cauchy sequence in $\mathcal{A}$. Now, it is easy to see that
    \begin{align*}
        \mathcal{A}_{f} &=  \{f \in \mathcal{C}^2 \,|\, \| f \|_{\infty} \leq c_1 \}\\
        \mathcal{A}_{\nabla f} &=  \{f \in \mathcal{C}^2 \,|\, \| \nabla f \|_{\infty} \leq c_2\}\\
        \mathcal{A}_{\nabla^2 f} &= \{f \in \mathcal{C}^2 \,|\, \| \nabla^2 f \|_{\infty} \leq c_3 \}
    \end{align*}
    are all complete with respect to their corresponding sup-norm. Therefore, there exist $\bar{f}, \mathbf{u}$ and $\mathbf{v}$ such that $f^k \rightarrow \bar{f}$, $\nabla f^k \rightarrow \mathbf{u}$ and $\nabla^2 f^k \rightarrow \mathbf{v}$. Furthermore, since $\bar{f}$ is $\mathcal{C}^2$, we have $\mathbf{u} = \nabla \bar{f}$ and $\mathbf{v} = \nabla^2 \bar{f}$. Lastly, by continuity, we have that $\bar{f}$ also satisfy the boundary constraints. This shows that $\mathcal{A}$ is complete with respect to $\| \cdot \|.$

    To show that $\mathcal{A}$ is totally bounded, we make use of a result in \cite{ZhangDaoping2022AUFf} that the set:
    \begin{equation}
        \tilde{\mathcal{A}} := \{f \in \mathcal{C}^2 \,|\, \|f\|_{\infty} \leq c_1, \| \nabla f \| \leq c_2, \| \nabla^2 f \|_{\infty} \leq c_3\}
    \end{equation}
    is totally bounded. Since $\mathcal{A} \subset \tilde{\mathcal{A}}$, it follows that $\mathcal{A}$ is also totally bounded.

    Therefore, $\mathcal{A}$ is compact and \eqref{eqn:hybrid_formulation} admits a minimizer in $\mathcal{A}$. This finishes the proof.
\end{proof}
%


%
%



\newpage
\bibliographystyle{spmpsci}
\bibliography{mylib}

\end{document}